 \documentclass[accepted]{uai2021} 

\usepackage[american]{babel}

\usepackage{natbib} 
    \renewcommand{\bibsection}{\subsubsection*{References}}
\usepackage{mathtools} 
\usepackage{booktabs} 
\usepackage{tikz} 
\usepackage{amsthm}
\usepackage{hyperref}

\newcommand{\swap}[3][-]{#3#1#2} 

\usepackage{amsmath,amsfonts,bm}

\def\eqref#1{equation~\ref{#1}}

\def\1{\bm{1}}

\DeclareMathAlphabet{\mathsfit}{\encodingdefault}{\sfdefault}{m}{sl}
\SetMathAlphabet{\mathsfit}{bold}{\encodingdefault}{\sfdefault}{bx}{n}

\newcommand{\E}{\mathbb{E}}

\newcommand{\bd}{\mathbf{d}}
\newcommand{\be}{\mathbf{e}}

\newcommand{\bu}{\mathbf{u}}

\newcommand{\bx}{\mathbf{x}}

\newcommand{\bz}{\mathbf{z}}

\newcommand{\ie}{\emph{i.e.}}
\newcommand{\eg}{\emph{e.g.}}
\usepackage{textcomp}
\usepackage{hyperref}
\usepackage{url}
\usepackage{booktabs}
\usepackage{subfigure}
\usepackage{graphicx}
\usepackage{psfrag}
\usepackage{epsfig}
\usepackage{enumitem}
\usepackage[noend,ruled,vlined]{algorithm2e}
\usepackage{grffile}
\usepackage{amsmath}
\usepackage{amssymb}
\usepackage{multirow}
\newcommand{\blue}[1]{\textcolor[rgb]{0.00,0.00,1.00}{#1}}
\newcommand{\red}[1]{\textcolor[rgb]{1.00,0.00,0.00}{#1}}
\newtheorem{theorem}{Theorem}
\newcommand{\define}{\triangleq}
\newcommand{\ci}{\perp\!\!\!\perp}

\title{Symmetric Wasserstein Autoencoders}

\author[1]{Sun Sun}
\author[2]{Hongyu Guo}
\affil[1,2]{%
    National Research Council Canada\\ 
    
    Ottawa, ON., K1A 0R6, Canada\\
    
    \{sun.sun,hongyu.guo\}@nrc-cnrc.gc.ca
}

\begin{document}
\maketitle

\begin{abstract}

Leveraging the framework of Optimal Transport, we introduce a new family of generative autoencoders with a learnable prior, called Symmetric Wasserstein Autoencoders (SWAEs). 
We propose to symmetrically match the joint distributions of the observed data and the latent representation induced by the encoder and the decoder. The  
resulting
algorithm jointly optimizes
the modelling losses in both the data and the latent spaces with  the loss in the data space leading to the denoising effect. With the symmetric treatment of the data and the latent representation, the algorithm implicitly preserves 
the local structure of the data in the latent space.
To further improve the quality of the latent representation, we incorporate a reconstruction loss into the objective, which significantly benefits both the generation and reconstruction.
We empirically show the superior performance of SWAEs over 
the
state-of-the-art generative autoencoders in terms of classification, reconstruction, and 
generation. 
\end{abstract}

\section{Introduction}
Deep generative models have emerged as powerful frameworks for modelling complex data. 
Widely used  families of such
models include Generative Adversarial Networks (GANs)~\citep{goodfellow2014generative},  Variational Autoencoders (VAEs)~\citep{rezende2014stochastic,kingma2013auto}, and autoregressive models \citep{uria2013rnade,van2016pixel}. The VAE-based framework has been popular as it yields a bidirectional mapping, \ie, it consists of  both an inference model (from data to latent space) and a generative model (from latent to data space). With an inference mechanism VAEs can provide a useful latent representation  that  captures salient information about the observed data. Such latent representation  can in turn benefit  downstream tasks such as clustering, classification, and data generation.  
In particular, the VAE-based approaches  have
achieved impressive performance results on 
challenging real-world applications, including  image synthesizing~\citep{RazaviOV19}, natural text generation~\citep{HuYLSX17}, and neural machine translation~\citep{SutskeverVL14}.

VAEs aim to maximize a tractable variational lower bound on the log-likelihood of the observed data, commonly called the ELBO.
Since VAEs focus on modelling the marginal likelihood of the data instead of the joint likelihood of the data and the latent representation, the quality of the latent  is not well assessed \citep{alemi2017fixing,zhao2019infovae}, which is undesirable for learning useful representation. 
Besides the perspective of  maximum-likelihood learning of the data, the objective of VAEs is equivalent to minimizing the  KL divergence between the encoding and the decoding distributions, with the former modelling the joint distribution of the observed data and the latent representation induced by the encoder and the latter modelling the corresponding joint distribution induced by the decoder. Such connection has been revealed in several recent work \citep{livne2019mim,esmaeili2018structured,pu2017adversarial,chen2018symmetric}. 
Due to the asymmetry of the KL divergence, it is highly likely that the generated samples are of a low probability in the data distribution, which often leads to unrealistic generated samples \citep{li2017towards,alemi2017fixing}.

A lot of work has proposed to improve VAEs from different perspectives. For example, 
to enhance the latent expressive power   
 VampPrior \citep{tomczak2018vae}, normalizing flow \citep{rezende2015variational}, and Stein VAEs \citep{pu2017vae} replace the Gaussian distribution imposed on the latent variables with a more sophisticated and flexible distribution.
However, these methods are all based on the objective of VAEs, which therefore are unable to alleviate the limitation of VAEs induced by the objective.  To improve the latent representation \citet{zhao2019infovae} explicitly includes the mutual information between the data and the latent into the objective. Moreover,
to address the asymmetry of the KL divergence in VAEs ~\citet{livne2019mim,chen2018symmetric,pu2017adversarial} leverage a  symmetric divergence measure between the encoding and the decoding distributions. 
Nevertheless, these methods typically involve a sophisticated objective function that either depends on unstable adversarial training or challenging  approximation of the mutual information.

In this paper, we leverage Optimal Transport (OT)  \citep{villani2008optimal,peyre2019computational} to  symmetrically match the encoding and the decoding distributions. The OT optimization is generally challenging particularly in high dimension, and we address this difficulty by transforming the OT cost into a simpler form amenable to efficient numerical implementation. Owing to the symmetric treatment of the observed data and the latent representation,
the local structure of the  data can be implicitly preserved  in the latent space. However, we found that with the symmetric treatment only the performance of the generative model may not be satisfactory.
To  improve the generative model we additionally include a reconstruction loss into the objective, which is shown to significantly benefit the quality of the generation and reconstruction.    

Our contributions can be summarized as follows. Firstly, we propose a new family of generative autoencoders, called Symmetric Wasserstein Autoencoders (SWAEs). 
Secondly, we adopt a learnable latent prior, parameterized as a mixture of the conditional priors given the learnable pseudo-inputs, which prevents
SWAEs from over-regularizing the latent variables.  
Thirdly, we empirically perform an ablation study of SWAEs in terms of the
KNN classification, denoising, reconstruction, and  sample generation. 
Finally, we empirically verify, using  benchmark tasks,  the superior performance of SWAEs over several state-of-the-art generative autoencoders.

\section{Symmetric Wasserstein Autoencoders}
In this section we introduce a new family of generative autoencoders, called Symmetric Wasserstein Autoencoders (SWAEs).

\subsection{OT Formulation}\label{subsec:ot}

Denote the random vector at the encoder as $\be \define (\bx_e, \bz_e)\in \mathcal{X}\times \mathcal{Z}$, which contains both the observed data $\bx_e\in \mathcal{X}$ and the latent representation $\bz_e \in \mathcal{Z}$. We call the distribution $p(\be) = p(\bx_e)p(\bz_e|\bx_e)$  \textit{the encoding distribution}, where $p(\bx_e)$ represents the data distribution and $p(\bz_e|\bx_e)$ characterizes an inference model.  Similarly, denote the random vector at the decoder as $\bd \define (\bx_d, \bz_d)\in \mathcal{X} \times \mathcal{Z}$, which consists of both the latent prior $\bz_d\in \mathcal{Z}$ and the generated data $\bx_d\in \mathcal{X}$. We call the distribution  $p(\bd) = p(\bz_d)p(\bx_d|\bz_d)$ \textit{the decoding distribution}, where $p(\bz_d)$ represents the prior distribution and $p(\bx_d|\bz_d)$ characterizes a generative model. 

Particularly, the objective of VAEs is to maximize a tractable variational lower bound on the data log-likelihood, called the Evidence Lower Bound (ELBO):
\begin{align}\label{eq:vaeelbo}
    \E_{p(\bx_e)}\left[ \E_{ p(\bz_e|\bx_e)}[\log p(\bx_d|\bz)] - \mathcal{D}_{\textrm{KL}}(p(\bz_e|\bx_e)||p(\bz_d))\right].
\end{align}
It can also  be shown that the objective of VAEs is equivalent to minimizing the KL divergence (or maximizing the negative KL divergence) between the encoding and the decoding distributions \citep{livne2019mim,esmaeili2018structured,pu2017adversarial,chen2018symmetric}:
\begin{align}
\nonumber
    &\hspace{0.3cm} -\mathcal{D}_{\textrm{KL}}(p(\bx_e, \bz_e)||p(\bx_d, \bz_d)) \\
    \label{eq:vaekl}
    &= \E_{p(\bx_e, \bz_e)}\left[\log \frac{p(\bx_d, \bz_d)}{p(\bz_e|\bx_e)}\right] -\E_{ p(\bx_e)}[\log p(\bx_e)].
\end{align}
The 
right hand side
of  \eqref{eq:vaekl} is only different from  \eqref{eq:vaeelbo} in terms of a constant, which is the entropy of the observed data.

To address the limitation in VAEs,
first  we propose to treat the  data and the latent representation symmetrically instead of asymmetrically
by minimizing the $p$-th Wasserstein distance between $p(\be)$ and $p(\bd)$ leveraging Optimal Transport (OT) \citep{villani2008optimal,peyre2019computational}.


OT provides a framework for comparing two distributions in a Lagrangian framework, which seeks the minimum cost for transporting one distribution to another. We focus on the primal problem of OT, and
Kantorovich's formulation  \citep{peyre2019computational} is given by:
\begin{align}
\label{kant}
    W_c(p(\be), p(\bd)) & \define \inf_{\Gamma\in \mathcal{P}(\be \sim p\left(\be), \bd \sim p(\bd)\right)} \E_{(\be, \bd)\sim \Gamma} \quad c(\be, \bd), 
\end{align}
where $\mathcal{P}(\be \sim p\left(\be), \bd \sim p(\bd)\right)$, called the coupling between $\be$ and $\bd$, denotes the set of the joint distributions of $\be$ and $\bd$ with the marginals $p(\be)$ and $p(\bd)$, respectively, and $c(\be, \bd): (\mathcal{X},\mathcal{Z}) \times (\mathcal{X},\mathcal{Z}) \to [0, +\infty]$ denotes the cost function. When $((\mathcal{X},\mathcal{Z}) \times (\mathcal{X},\mathcal{Z}), d)$ is a metric space and the cost function $c(\be, \bd) = d^p(\be,\bd)$ for $p\ge 1$, $W_p$, the $p$-th root of $W_c$ is defined as the $p$-th Wasserstein distance. In particular, it can be proved that the $p$-th Wasserstein distance is a metric hence \textit{symmetric}, and metrizes the weak convergence
(see, \eg, \citep{santambrogio2015optimal}).
      
Optimization of \eqref{kant} is computationally prohibitive especially in high dimension \citep{peyre2019computational}.
To provide an efficient solution, we restrict to the deterministic encoder and decoder. Specifically,  at the encoder we have the latent representation $\bz_e = E(\bx_e)$ with the function $E: \mathcal{X}\to \mathcal{Z}$, and at the decoder we have the generated data $\bx_d = D(\bz_d)$ with the function $D: \mathcal{Z}\to \mathcal{X}$. It turns out that with the deterministic condition instead of searching for an optimal coupling in high dimension, we can find a proper conditional distribution $p(\bz_d|\bx_e)$ with the marginal $p(\bz_d)$.  

\begin{theorem}
\label{th:wae}
Given the deterministic encoder $E: \mathcal{X}\to \mathcal{Z}$ and the deterministic decoder $D: \mathcal{Z}\to \mathcal{X}$, the OT problem in \eqref{kant} can be transformed to the following:
\begin{align}
\label{th:wed}
    W_c(p(\be), p(\bd)) = \inf_{p(\bz_d|\bx_e)} \E_{p(\bx_e)}\E_{p(\bz_d|\bx_e)}  \quad c(\be, \bd),
\end{align}
where the observed data follows the distribution $p(\bx_e)$ and the prior follows the distribution $p(\bz_d)$.
\end{theorem}


\begin{proof}
The proof extends that of Theorem 1 in \citep{tolstikhin2018wasserstein}. In particular, \citep{tolstikhin2018wasserstein} aims to minimize the OT cost of the marginal distributions $p(\bx_e)$ and $p(\bx_d)$, and the proof there is based on the joint probability of three random variables: the observed data, the generated data, and the latent representation. In contrast, we propose to minimize the OT cost of the joint distributions of the observed data and the latent representation induced by the encoder and the decoder. As a result our proof is based on the joint distribution of four random variables $(\bx_e, \bz_e, \bx_d, \bz_d)\in \mathcal{X} \times \mathcal{Z} \times \mathcal{X} \times \mathcal{Z} $. We assume that the joint distribution $p(\bx_e, \bz_e, \bx_d, \bz_d)$ satisfies the following three conditions:
\begin{enumerate}
    \item $\be \define (\bx_e,\bz_e) \sim p(\bx_e)p(\bz_e|\bx_e)$;
    \item $\bd \define (\bx_d, \bz_d) \sim p(\bz_d)p(\bx_d|\bz_d)$; and
    \item $\bx_d \ci \bx_e |\bz_d$ (conditional independence).
\end{enumerate}
The first two conditions specify the encoder and the decoder respectively, and the last condition indicates that given the latent prior the generated data and the observed data are independent.

Denote the set of the above joint distributions as $\mathcal{P}(\bx_e, \bz_e, \bx_d, \bz_d)$.
Obviously, we have $\mathcal{P}(\bx_e, \bz_e, \bx_d, \bz_d)\subseteq \mathcal{P}(\be \sim p\left(\be), \bd \sim p(\bd)\right)$ due to the third condition. If the decoder is deterministic, $p(\bx_d|\bz_d)$ is a Dirac distribution thus $\mathcal{P}(\bx_e, \bz_e, \bx_d, \bz_d) = \mathcal{P}(\be \sim p\left(\be), \bd \sim p(\bd)\right)$. With this result, we can rewrite the objective of the underlying OT problem as follows:
\begin{align}
\nonumber
    &W_c(p(\be), p(\bd)) \\
    \nonumber
    & = \inf_{\Gamma\in \mathcal{P}(\bx_e, \bz_e, \bx_d, \bz_d)} \E_{(\be, \bd)\sim \Gamma} \quad c(\be, \bd)\\
    \label{eq:nx}
    & = \inf_{\Gamma\in \mathcal{P}(\bx_e, \bz_e, \bz_d)} \E_{(\bx_e, \bz_e, \bz_d)\sim \Gamma} \quad c(\be, \bd)\\
    & = \inf_{p(\bz_e|\bx_e),\; p(\bz_d|\bx_e, \bz_e)} \E_{p(\bx_e)}\E_{p(\bz_e|\bx_e)}\E_{p(\bz_d|\bx_e, \bz_e)}  \quad c(\be, \bd)\\
    \label{eq:den}
    & = \inf_{p(\bz_d|\bx_e)} \E_{p(\bx_e)}\E_{p(\bz_d|\bx_e)}  \quad c(\be, \bd),
\end{align}
where in \eqref{eq:nx} $\mathcal{P}(\bx_e, \bz_e, \bz_d)$ denotes the set of the joint distributions of $(\bx_e, \bz_e, \bz_d)$ induced by $\mathcal{P}(\bx_e, \bz_e, \bx_d, \bz_d)$ and it holds due to the deterministic decoder, and \eqref{eq:den} holds due to the deterministic encoder.
\end{proof}

If $\mathcal{X}\times \mathcal{Z}$ is the Euclidean space endowed with the $L_p$ norm, then the expression in \eqref{th:wed} equals the following:
\begin{align}
\nonumber
    &W_c(p(\be), p(\bd)) \\
    &= \inf_{p(\bz_d|\bx_e)} \E_{p(\bx_e)}\E_{p(\bz_d|\bx_e)}  \quad \|\bx_e-D(\bz_d)\|_p^p + \|E(\bx_e)-\bz_d\|_p^p,
    \label{obj2}
\end{align}
where in the objective we call the first  term  the $\bx$-loss and the second term  the $\bz$-loss. 
With the above transformation, we decompose the loss in the joint space into the losses in both the data and the latent spaces. Such decomposition is crucial and allows us to treat the data and the latent representation symmetrically.

The $\bx$-loss, \ie, $\|\bx_e-D(\bz_d)\|_p^p$,  represents the discrepancy in the data space, and can be interpreted from two different  perspectives. Firstly,  since $D(\bz_d)$ represents the generated data, the $\bx$-loss  essentially  minimizes the dissimilarity between the observed data and the generated data.
Secondly, the $\bx$-loss is closely related to the objective of Denoising Autoencoders (DAs) \citep{vincent2008extracting,vincent2010stacked}. In particular,  DAs aim to minimize the discrepancy between the observed data and a partially destroyed version of the observed data. The corrupted data can be obtained by means of a stochastic mapping from the original data (\eg, via adding noises).  
By contrast, the $\bx$-loss  can be explained in the same way with the generated data  being interpreted as the corrupted data. This is because the prior $\bz_d$ in $D(\bz_d)$ is sampled from the conditional distribution $p(\bz_d|\bx_e)$, which depends on the observed data $\bx_e$. Consequently, the generated data $D(\bz_d)$, obtained by feeding $\bz_d$ to the decoder, is  stochastically related to  the observed data $\bx_e$.  With this insight, the same as the objective of DAs, the  $\bx$-loss can lead to the denoising effect.

\begin{figure}[t]
\centering
\subfigure[SWAE]
{\includegraphics[width=0.4\textwidth]{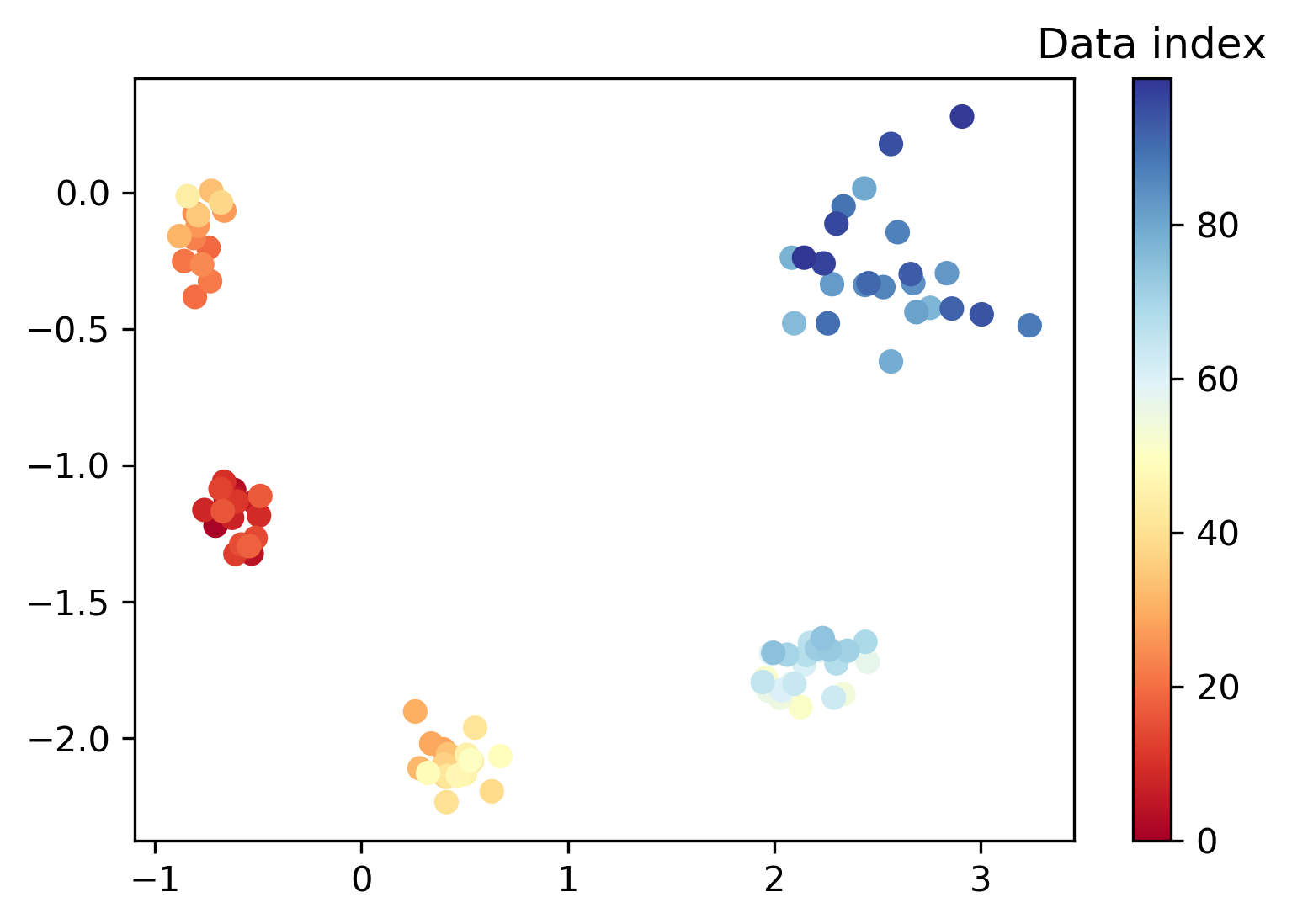}}
\subfigure[VAE]
{\includegraphics[width=0.4\textwidth]{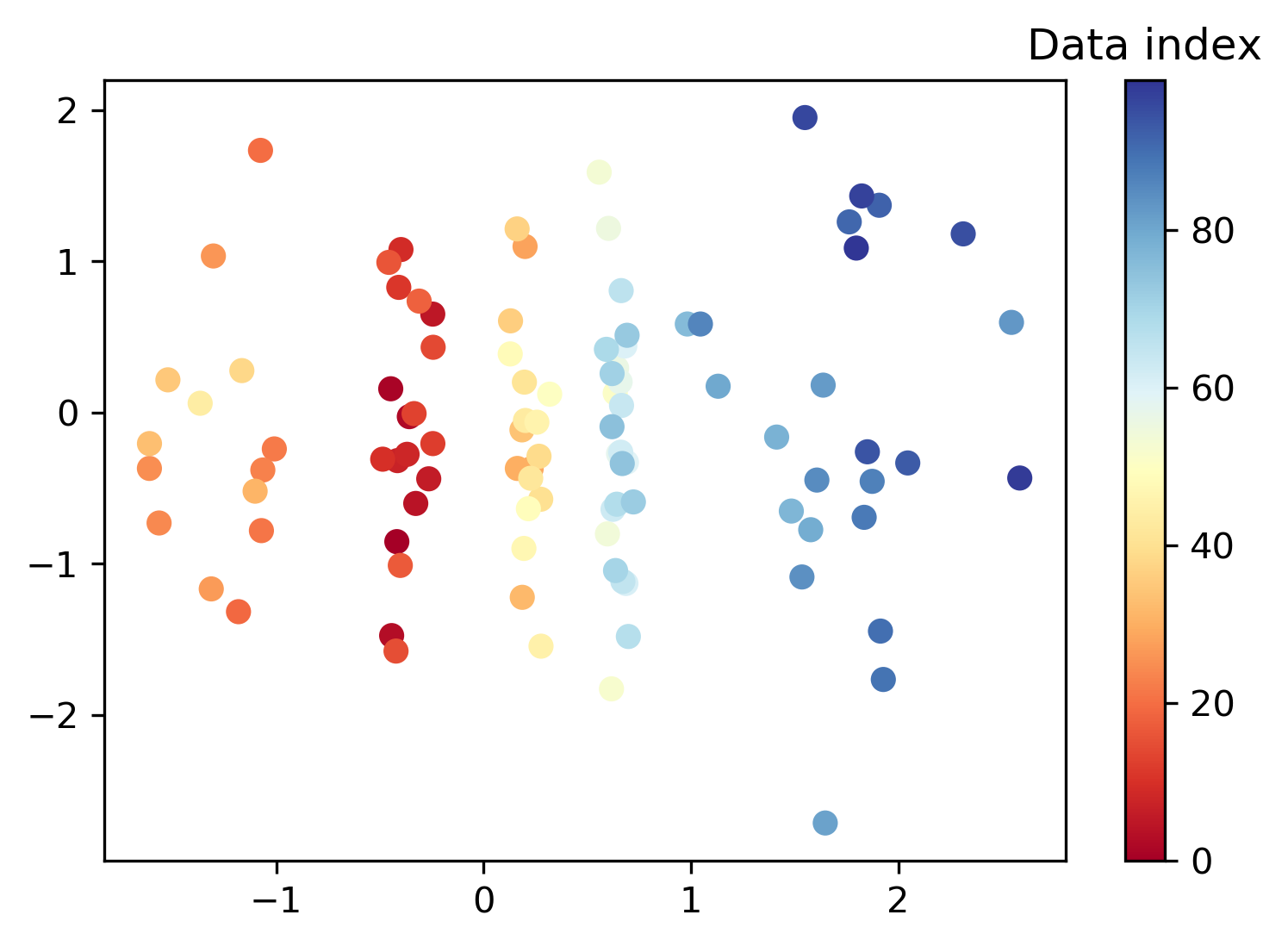}}	
\caption{Latent representations of  $100$ GMM samples (mode $5$ and dimension $10$) with dim-$\bz = 2$.  The indexes of these latent representations are sorted based on the distance to a target sample in the data space, \ie,  Index $0$ is associated with  the target sample and  Index $100$ is associated with the furthest sample to the target in the data space. With SWAE (top) data samples that are close in the data space are also close in the latent space, while VAE (bottom) cannot preserve such correspondence. }
\label{fig:gmm}
\end{figure}

The $\bz$-loss, \ie, $\|E(\bx_e)-\bz_d\|_p^p$, represents the discrepancy in the latent space. The whole objective in \eqref{obj2} hence simultaneously minimizes the discrepancy in the data and the latent spaces. 
Observe that in \eqref{obj2} $E(\bx_e)$ is the latent representation of $\bx_e$ at  the encoder, while $\bz_d$ can be thought of as the latent representation of $D(\bz_d)$ at the decoder. 
With such connection, the optimization of \eqref{obj2} can  preserve  \textit{the local data structure} in the latent space. More specifically, since $\bx_e$ and $D(\bz_d)$ are stochastically dependent, roughly speaking, if two data samples are close to each other in the data space, their corresponding latent representations are also expected to be close. 
This is due to the symmetric treatment of the data and the latent representation. In Figure \ref{fig:gmm} we illustrate this effect and compare  SWAE with VAE.

\textbf{Comparison with WAEs} \citep{tolstikhin2018wasserstein}
The objective in \eqref{obj2} minimizes the OT cost between the joint distributions of the data and the latent, \ie, $W_c(p(\be), p(\bd))$, while
the objective of
WAEs \citep{tolstikhin2018wasserstein} minimizes the OT cost between the marginal distributions of the data, \ie,  $W_c(p(\bx_e), p(\bx_d))$, where $p(\bx_d)$ is the marginal data distribution induced by the decoding distribution $p(\bd)$. 
The problem of WAEs is first formulated as an optimization with the constraint $p(\bz_e) = p(\bz_d)$, where $p(\bz_e)$ is the marginal distribution induced by the encoding distribution $p(\be)$, and then relaxed by adding a regularizer. With the deterministic decoder, the final optimization problem of WAEs is as follows:
\begin{align}
\label{waeobj}
    \inf_{p(\bz_e|\bx_e)} \E_{p(\bx_e)}\E_{p(\bz_e|\bx_e)}\quad c(\bx_e, D(\bz_e)) + \lambda \mathcal{D}(p(\bz_e), p(\bz_d)),
\end{align}
where $\mathcal{D}(,)$ denotes some divergence measure.
Comparing \eqref{waeobj} to \eqref{obj2}, we can see that both methods decompose the loss into the losses in the data and the latent spaces. Differently, in \eqref{waeobj} the first term  reflects the reconstruction loss in the data space and the second term  
represents the distribution-based dissimilarity in the latent space; while in \eqref{obj2}  the $\bx$-loss is closely related to the denoising and the generation quality, and  the $\bz$-loss  measures the sample-based dissimilarity. Moreover, \eqref{waeobj} is optimized over the posterior $p(\bz_e|\bx_e)$ with a fixed prior $p(\bz_d)$, while \eqref{obj2} is optimized over the conditional prior $p(\bz_d|\bx_e)$ with a potentially learnable prior.

\begin{figure*}[t]
	\centering
	\includegraphics[width=0.9\textwidth]{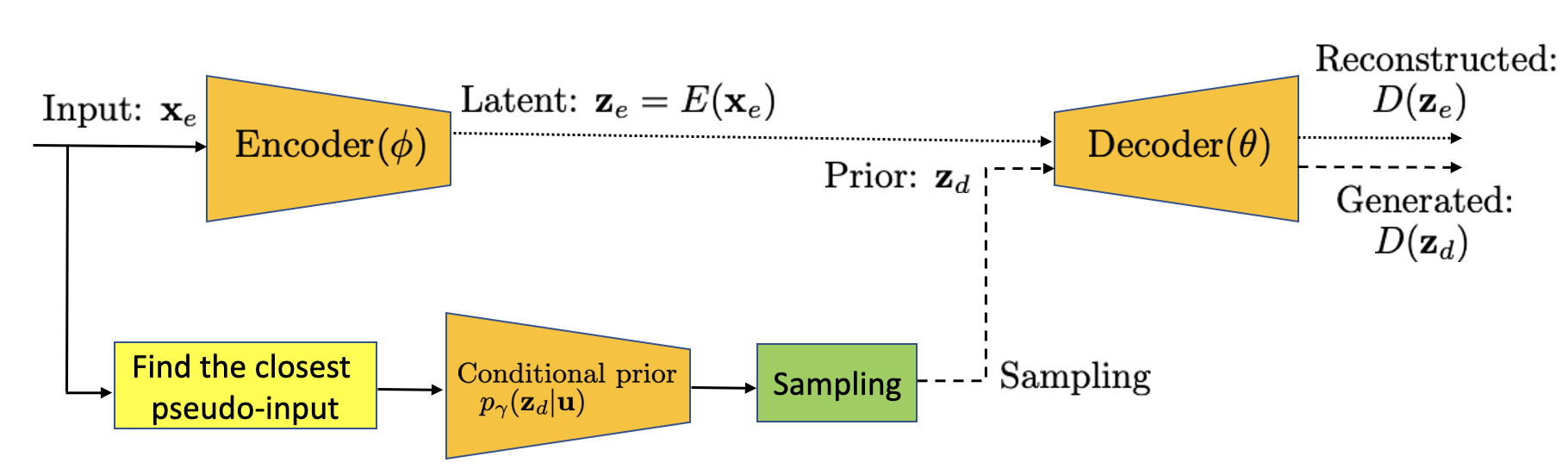}
	\caption{Network architecture of SWAEs. To generate new data latent samples are first drawn from the marginal prior  $p(\bz_d)$ based on the conditional priors $p(\bz_d|\bu_k)$, and are then fed to the decoder.}
	\label{fig:sys}
\end{figure*}

\subsection{Improvement of latent representation}
The objective in \eqref{obj2} only seeks to match the encoding and the decoding distributions.  Besides the encoder and the decoder structures, there is no explicit constraint on the correlation between the data and the latent representation within each joint distribution. Lacking of such constraint typically results in a low quality of reconstruction \citep{dumoulin2016adversarially,li2017alice}. 
Therefore, we incorporate a reconstruction-based loss into the objective  associated with a controllable coefficient. Additionally, since the dimension of the latent space is usually much smaller than that of the 
data space, we associate a weighting parameter to balance
these two types of losses.
Overall, the objective function can be represented as follows:
\begin{align}
\nonumber
    &\hspace{-1cm} \inf_{p(\bz_d|\bx_e)} \E_{p(\bx_e)}\E_{p(\bz_d|\bx_e)}  \quad \beta \|\bx_e-D(\bz_d)\|_p^p \\
    \label{obj3}
    &+ (1-\beta) \|\bx_e-D(\bz_e)\|_p^p + \alpha \|E(\bx_e)-\bz_d\|_p^p,
\end{align}
where $\|\bx_e-D(\bz_e)\|_p^p$ denotes the reconstruction loss, and $\beta (0 < \beta < 1)$ and $\alpha (\alpha > 0)$ are the weighting parameters. The weighting parameter $\beta$ controls the trade-off between the $\bx$-loss and the reconstruction loss, and a smaller value of $\beta$ generally leads to better reconstruction. To achieve a better trade-off between the generation and reconstruction $\beta$ needs to be carefully chosen.
We will perform an ablation study of SWAEs and show the importance of including the reconstruction loss into the objective for the generative model in Section \ref{sec:exp}.


\subsection{Algorithm}

Similar to many VAE-based generative models, we assume that the encoder, the decoder, and the conditional prior are parameterized by deep neural networks. Unlike the canonical VAEs, where the prior distribution is simple and given in advance, the proposed method adopts a learnable prior.
The benefits of a learnable prior, \eg, avoiding over-regularization and hence improving the quality of the latent representation, have been revealed in several recent works \citep{hoffman2016elbo, tomczak2018vae,atanov2019deep,klushyn2019learning}. Obviously, the conditional prior is related to the marginal prior via $\E_{\bx_e}p(\bz_d|\bx_e) = p(\bz_d)$. This indicates a way to design the prior as a mixture of the  conditional distributions, \ie,  $p^*(\bz_d) = \frac{1}{N}\sum_{n=1}^N p(\bz_d|\bx_{e,n})$, where   $\bx_{e,1}, \cdots, \bx_{e,N}$ are the training  samples. To avoid over-fitting, similar to   \citep{tomczak2018vae}, we replace the training samples with learnable pseudo-inputs and parameterize the prior distribution $p(\bz_d)$ as $p_{\gamma}(\bz_d) = \frac{1}{K} \sum_{k=1}^K p_{\gamma}(\bz_d| \bu_{k})$, where $\gamma$ denotes the  parameters of the conditional prior network, $\bu_k \in \mathcal{X}$ are the learnable pseudo-inputs, and $K$ is the number of the pseudo-inputs. 
We emphasize that the conditional prior $p(\bz_d|\bx_e)$ (or approximated $p(\bz_d|\bu_k)$) is used to obtain the marginal prior $p(\bz_d)$; while the posterior $p(\bz_e|\bx_e)$ is used for inference. In the experiment, we parameterize the conditional prior as a Gaussian distribution.

\begin{algorithm}[t]
\SetAlgoLined
\textbf{Require:} {The number of the pseudo-inputs $K$. The weighting parameters $\beta$ and $\alpha$. Initialize the parameters $\phi, \theta,$ and $\gamma$ of the encoder network,  the decoder network, and the conditional prior network, respectively. }

 \While{($\phi, \theta, \gamma, \{\bu_k\}$) not converged}{
 \begin{enumerate}[leftmargin=*,noitemsep]
     \item   Sample $\{\bx_{e,1}, \cdots, \bx_{e,N}\}$ from the training dataset.
     \item Find the closest pseudo-input $\bu^{(n)}$  of each training sample from the set $\{\bu_1,\cdots,\bu_K\}$.
     \item  Sample $\bz_{d,n}$ from the conditional prior $p_{\gamma}(\bz_d|\bu^{(n)})$ for $n=1, \cdots, N$.
     \item  Update ($\phi, \theta, \gamma, \{\bu_k\}$) by descending the cost function $\frac{1}{N}\sum_{n=1}^N \beta\|\bx_{e,n} - D(\bz_{d,n})\|^2_2 + (1-\beta)\|\bx_{e,n} - D(E(\bx_{e,n}))\|^2_2 + \alpha\|E(\bx_{e,n}) - \bz_{d,n}\|_2^2$.
 \end{enumerate}
 }
\caption{Symmetric Wasserstein Autoencoders (SWAEs)}
\label{alg}
\end{algorithm}

We call the proposed generative model Symmetric Wasserstein Autoencoders (SWAEs) as we treat the observed data and the latent representation symmetrically. We summarize the training algorithm in Algorithm \ref{alg} and show the network architecture in Figure \ref{fig:sys}. As an example, we define the cost function $c(,)$ as the squared $L2$ norm.

Since we use the pseudo-inputs instead of the training samples in the conditional prior, given each training sample we need to find the closest pseudo-input in Step 2. To measure the similarity, we can use, \eg, the $L2$ norm or the cosine similarity. Since the dimension of the latent  space is usually much smaller than that of the data space, to reduce the searching time we can alternatively perform Step 2 in the latent space as an approximation. Specifically, we can find the  closest latent representation of $E(\bx_{e,n})$ from the set $\{E(\bu_1), \cdots, E(\bu_K) \}$ so as to obtain the corresponding closest pseudo-input. From the experiment we found that such approximation  results in little performance degradation, 
and we attribute it to the preservation of the local structure as explained before.

\begin{table*}[t]
	\caption{Classification accuracy of $5$-NN (averaged over $5$ 
	trials). The standard deviation is generally less than $0.01$ and is  omitted in the table.}
	\label{tab:5nn}
	\centering
	\resizebox{0.983\textwidth}{!}{\begin{minipage}{\textwidth}
	\begin{tabular}{lcccccccccc }
		\toprule
		Dataset & dim-$\bz$ & SWAE  & SWAE & SWAE  &VAE & WAE-GAN & WAE-MMD & VampPrior & MIM    \\
		&&($\beta = 1$) & ($\beta = 0.5$) & ($\beta = 0$) &&&&&\\
		\hline
		\multirow{3}{4em}{MNIST} &8 & 0.96 & \textbf{0.97}&\textbf{0.97} &0.96&0.87&\textbf{0.97}&\textbf{0.97}&\textbf{0.97} \\
         & 40 &\textbf{0.97} &\textbf{0.97}&\textbf{0.97}&0.80&0.68&0.96&0.93&\textbf{0.97} \\
		&80 &\textbf{0.97}&\textbf{0.97}&\textbf{0.97}&0.60&0.90&0.94&0.86&0.96 \\
		\hline
		\multirow{3}{4em}{Fashion-MNIST} &8 &0.82 &0.81&\textbf{0.83}&0.80&0.71&0.80&0.81&0.82 \\
         & 40 &\textbf{0.84}&0.83&\textbf{0.84}&0.54&0.62 &0.82&0.81&0.82 \\
		&80 &\textbf{0.84}&0.83&0.83&0.37&0.54&0.76&0.74&0.81 \\
		\hline
		\multirow{3}{4em}{Coil20} &8 &0.95&0.97&0.97&0.95&\textbf{0.98}&0.78&0.91&0.89 \\
         & 40 & 0.97&0.98&0.97&0.90&\textbf{0.99}&\textbf{0.99}&0.96&0.96 \\
		&80 &0.97&\textbf{0.98}&\textbf{0.98}&0.97&\textbf{0.98}&\textbf{0.98}&0.96&\textbf{0.98} \\
		\hline
		\multirow{3}{4em}{CIFAR10-sub} &80 &\textbf{0.69} & 0.67&0.65 &0.67&0.61&0.68&0.68&0.66 \\
         & 256 &\textbf{0.70} & 0.66&0.61& 0.62&0.61&0.68&0.65&0.65 \\
		&512 & \textbf{0.70} & 0.66&0.62& 0.55&0.60&0.68&0.64& 0.66\\
		\bottomrule
	\end{tabular}
	\end{minipage}}
\end{table*}

\begin{figure*}[h]
	\centering
	\subfigure[SWAE ($\beta = 1$)]
	{\includegraphics[width=0.24\textwidth]{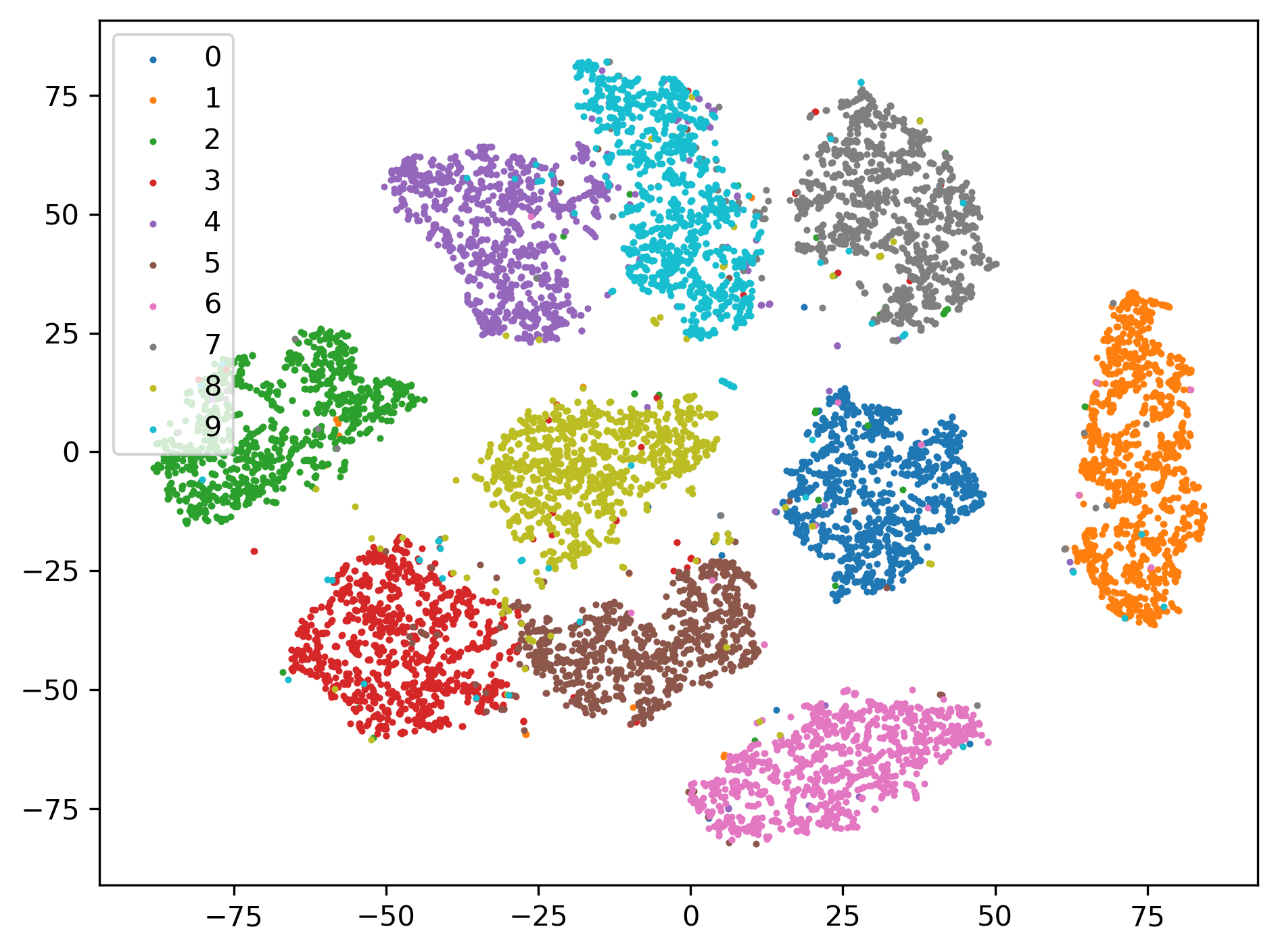}}	
		\vspace{-1mm}
	\subfigure[SWAE ($\beta = 0.5$)]
	{\includegraphics[width=0.24\textwidth]{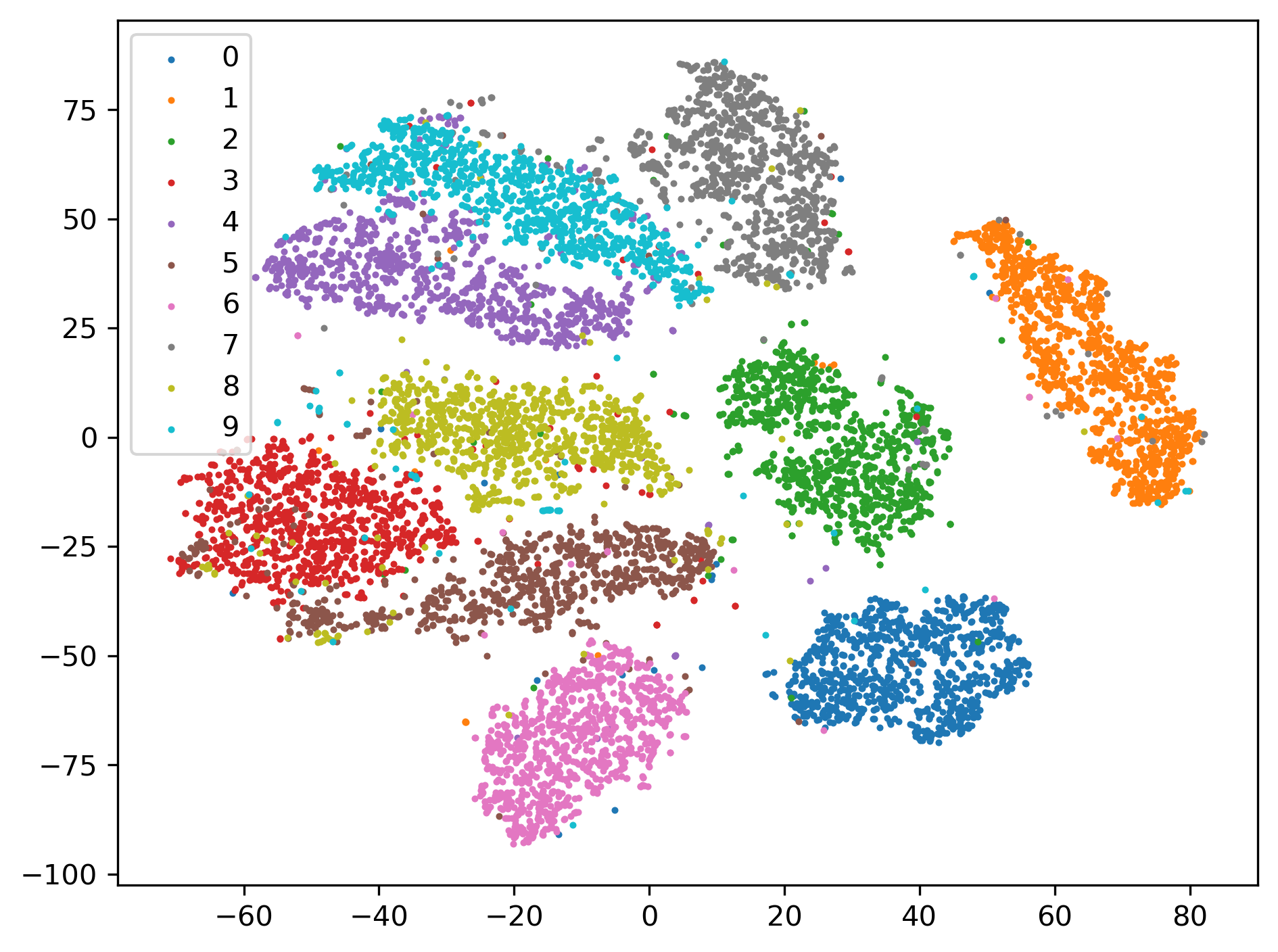}}	
		\vspace{-1mm}
	\subfigure[VAE]
	{\includegraphics[width=0.24\textwidth]{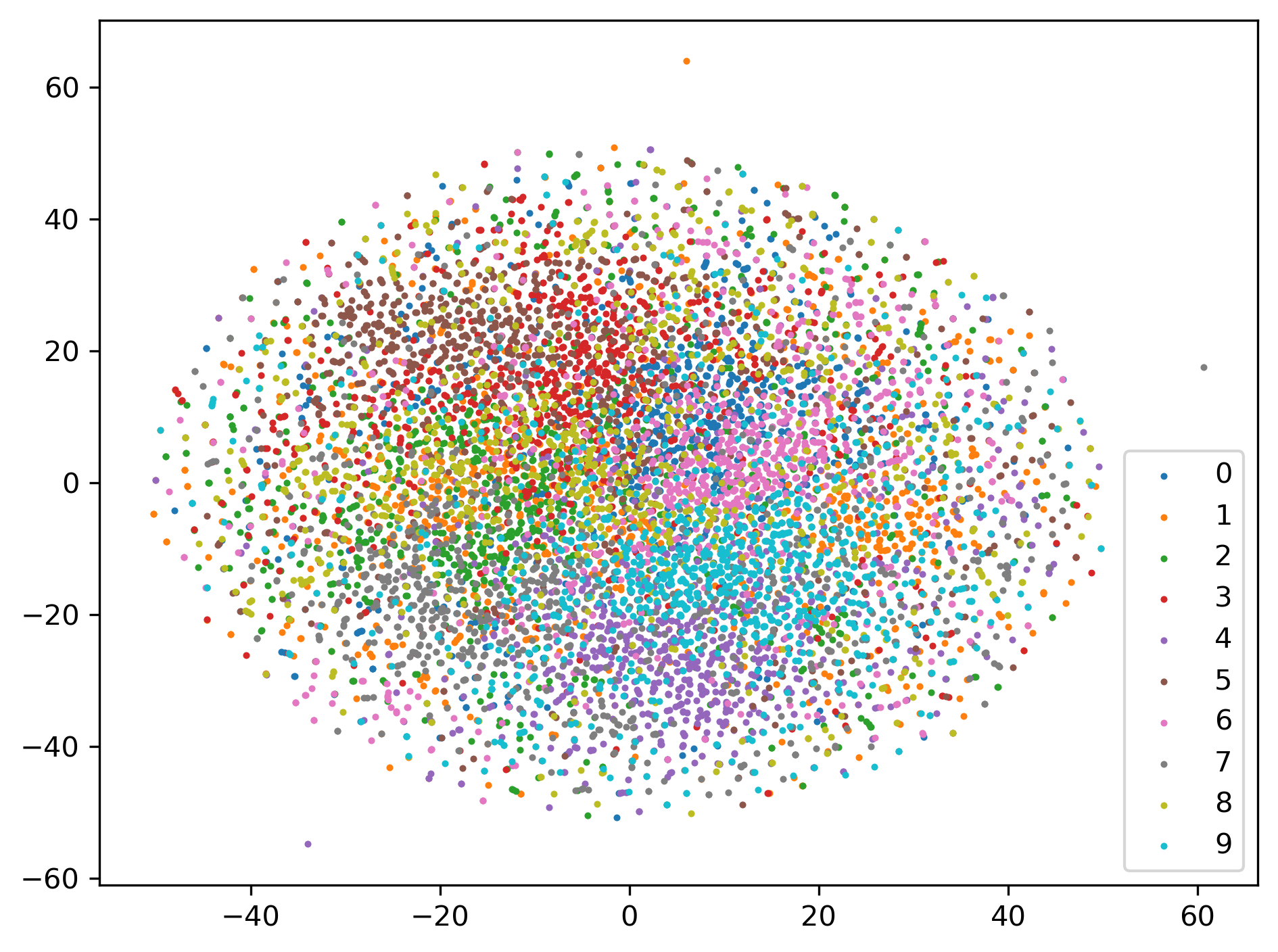}}	
		\vspace{-1mm}
	\subfigure[WAE-GAN]
	{\includegraphics[width=0.24\textwidth]{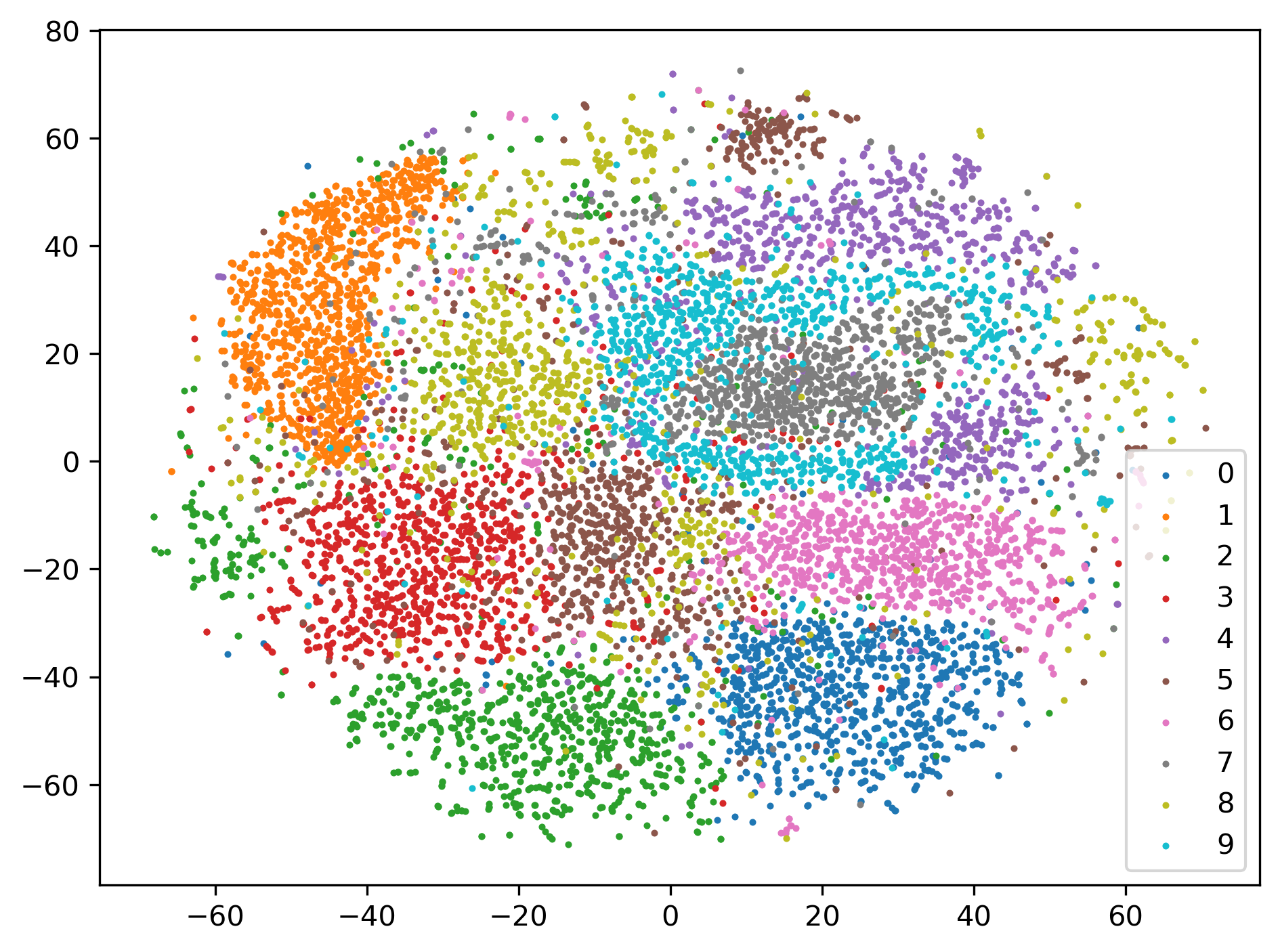}}	
		\vspace{-1mm}
	\subfigure[WAE-MMD]
	{\includegraphics[width=0.24\textwidth]{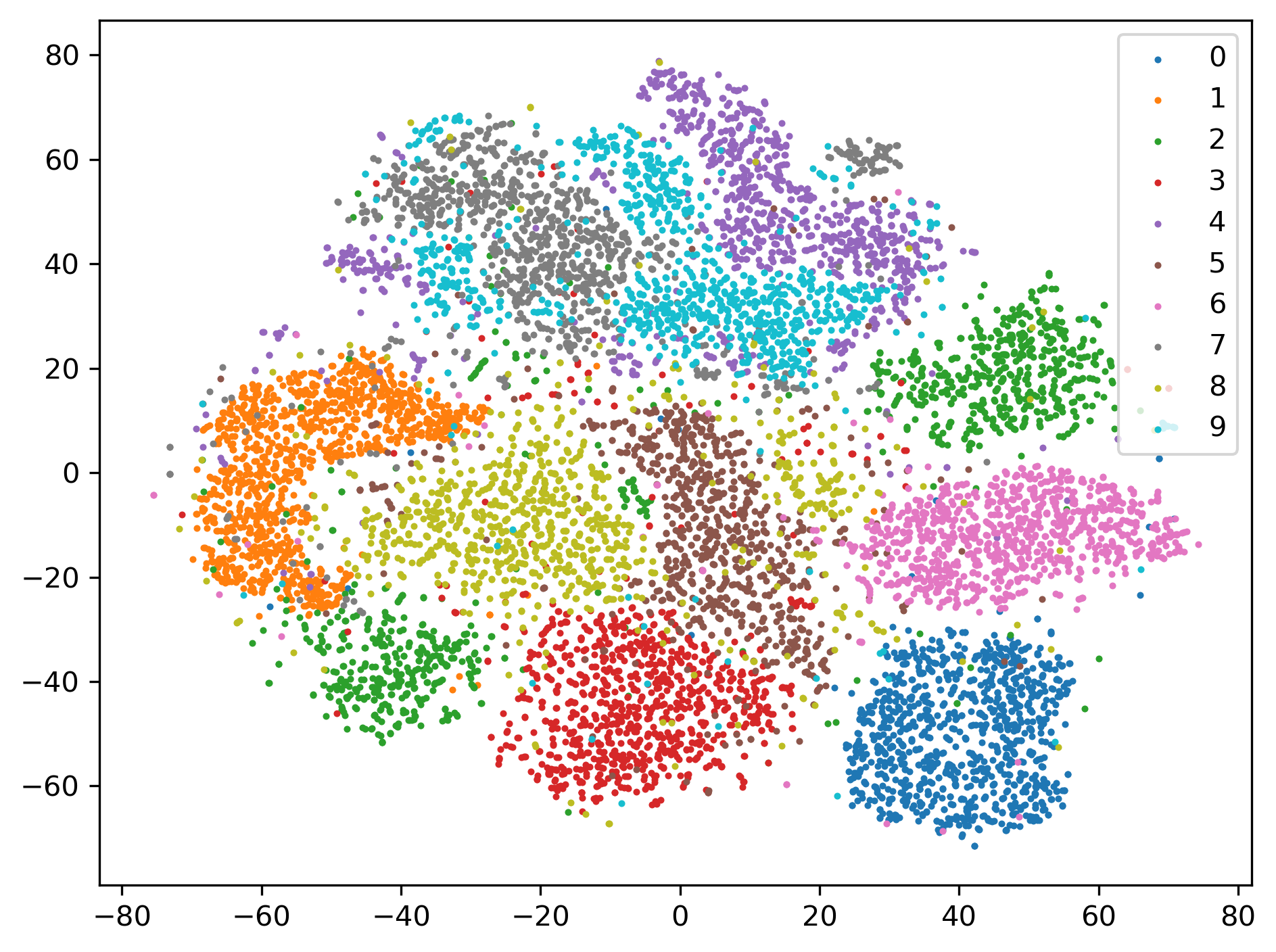}}	
	\subfigure[VampPrior]
	{\includegraphics[width=0.24\textwidth]{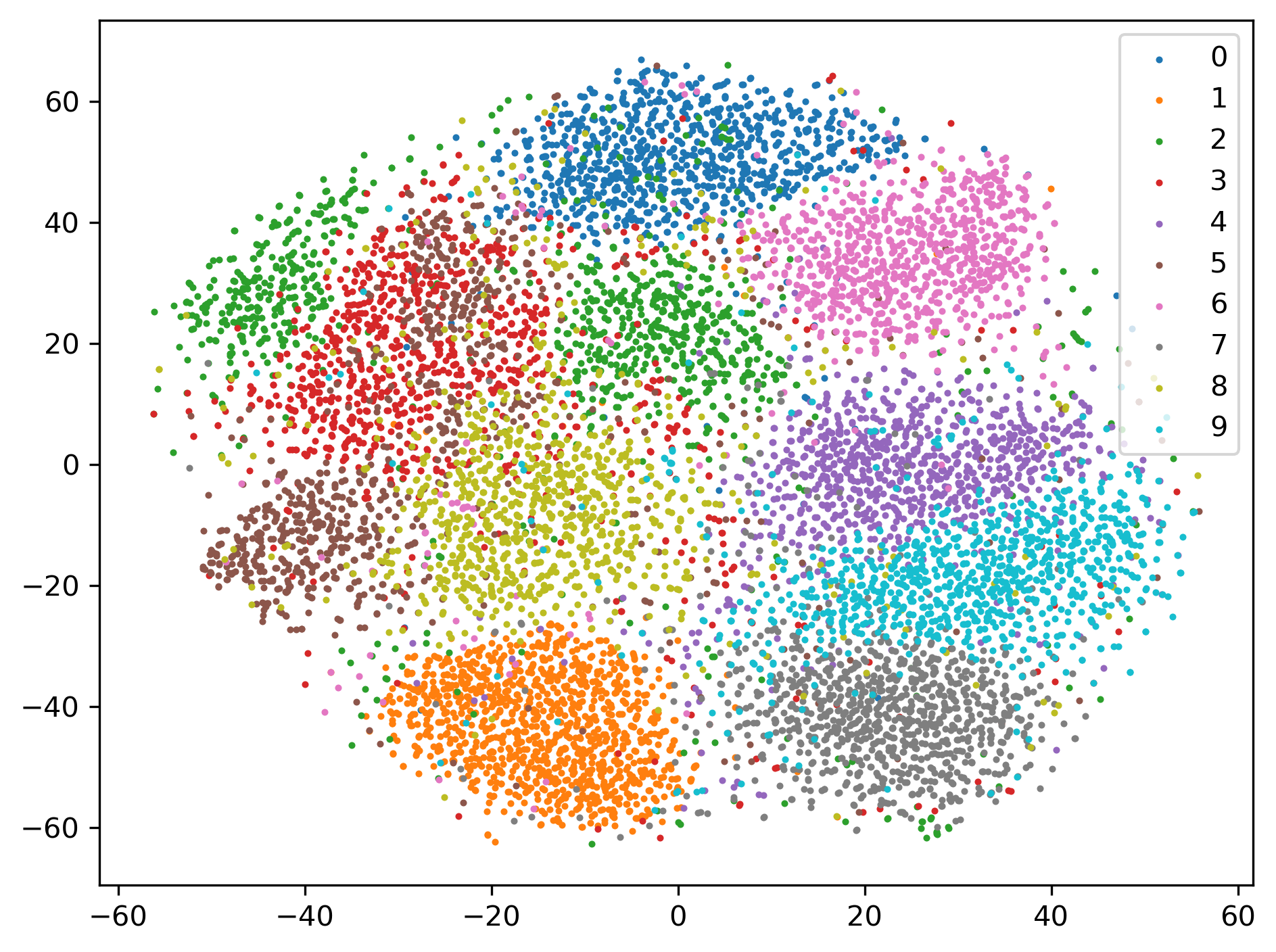}}
	\subfigure[MIM]
	{\includegraphics[width=0.24\textwidth]{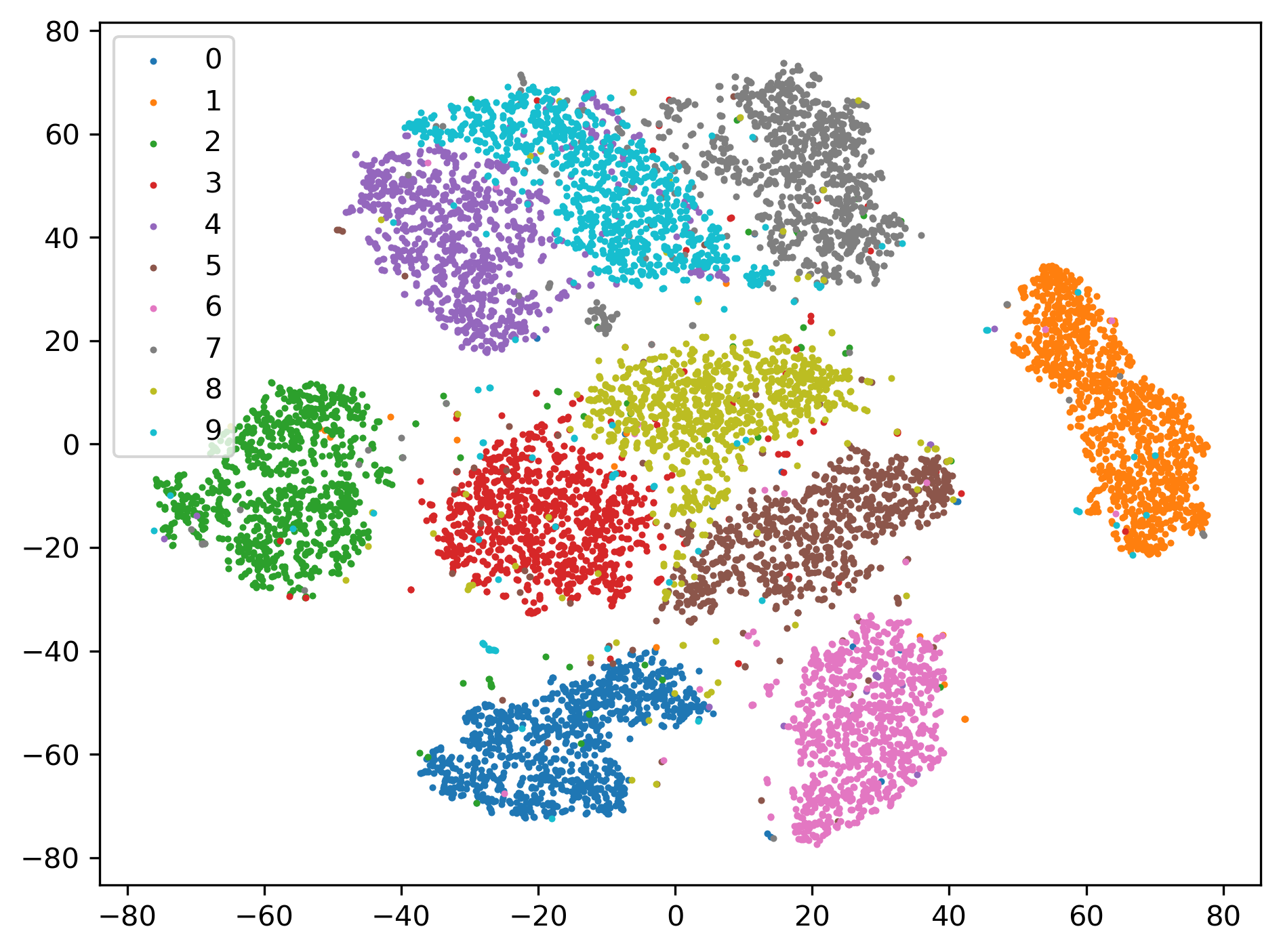}	}
	\caption{Projection of the latent representation to 2D via t-SNE on MNIST. dim-$\bz = 80$ for all methods.}
	\label{fig:mnist_latent}
\end{figure*}

\section{Experimental Results}\label{sec:exp}

In this section, we compare the performance of the proposed SWAE with several contemporary generative autoencoders, namely VAE \citep{kingma2013auto}, WAE-GAN \citep{tolstikhin2018wasserstein}, WAE-MMD \citep{tolstikhin2018wasserstein}, VampPrior  \citep{tomczak2018vae}, and MIM \citep{livne2019mim},  using four benchmark 
datasets: MNIST, Fashion-MNIST,  Coil20, and 
CIFAR10 with a subset of classes (denoted as  CIFAR10-sub). 

\subsection{Experimental Setup}
The design of neural network architectures is orthogonal to that of the algorithm objective, and can greatly affect the algorithm performance \citep{vahdat2020nvae}. Since MIM has the same network architecture as that of VampPrior, for fair comparison we also build SWAE  as well as VAE based on the VampPrior network architecture. In particular, VampPrior adopts the hierarchical latent structure with the convolutional layers (\ie, convHVAE ($L=2$)), where the gating mechanism is utilized as an element-wise non-linearity. The building block of the network structure of VAE and SWAE is the same as that of VampPior except that the latent structure is non-hierarchical. Different from SWAE, the prior of VampPrior and MIM is designed as a mixture of the posteriors (instead of a mixture of the conditional priors as in SWAE) conditioned on the learnable pseudo-inputs. The pseudo-inputs in SWAE, VampPrior, and MIM are 
initialized with the training samples. For VampPrior and MIM, the number of the pseudo-inputs $K$ is carefully chosen via the validation set. 
As suggested in
\citep{tomczak2018vae, livne2019mim} 
we set the value of $K$ in VampPrior and MIM on MNIST and Fashion-MNIST to $500$. We found that $K = 500$ is also suitable for
VampPrior and MIM on
Coil20 and CIFAR10-sub. Unlike VampPrior and MIM, for SWAE we found that increasing $K$ improves the performance and we
set $K$ to $4000$ on MNIST, Fashion-MNIST, and CIFAR10-sub. Coil20 is a relatively small dataset and we set $K$ to $500$ for SWAE, VampPrior, and MIM.

 WAE-GAN and WAE-MMD are the WAE-based models, where the divergence measure in the latent space is based on  GAN and the maximum mean discrepancy (MMD), respectively. The network structure of WAE-GAN and WAE-MMD is the same as that used
in \citep{tolstikhin2018wasserstein}. The prior of VAE, WAE-GAN, and WAE-MMD is set as an isotropic Gaussian. 

For SWAE, we set the weighting parameter $\alpha$ to $1$ in all cases; and in Step 2 we use the $L2$ norm as the similarity measure in the data space.
The algorithm is trained by Adam with the 
learning rate $=0.001$,
$\beta_1 = 0.9,$ and $ \beta_2 = 0.999$.
A detailed description of the datasets and the applied network architectures can be found in our supplementary file.

The code is available at 
\url{https://github.com/sunsunyyl/SWAE}.

\begin{table*}[h]
	\caption{Fr\'{e}chet Inception Distance (FID) on generated images (smaller is better).
	}
	\label{tab:fid}
	\centering
	\setlength{\tabcolsep}{4pt}
	\resizebox{0.988\textwidth}{!}{\begin{minipage}{\textwidth}
	\begin{tabular}{lcccccccccc }
		\toprule
		Dataset & dim-$\bz$ &  SWAE &  SWAE & SWAE & SWAE& VAE & WAE-GAN & WAE-MMD & VampPrior & MIM    \\
		 &  &  ($\beta = 1$) &  ($\beta = 0.5$) &($\beta = 0$)& $(\beta^*)$ & &  &  &  &     \\
		\hline
		MNIST &8 & 50 & 40 & 24 & 21 & 24 & \textbf{17} & 34 & 24 & 74 \\
		\hline
		Fashion-MNIST &8 & 65 & 57& 48 & 47 &60 & \textbf{41} & 100 & 51 & 83  \\
		\hline
		Coil20 &80 & 97 & \textbf{89}& 102& \textbf{89}& 278& 278& 320 &97& 113 \\
		\hline
        CIFAR10-sub &512 & 105& \textbf{44} & 183 & \textbf{44} & 242 & 114& 341 & 68 & 59 \\
 	\bottomrule
	\end{tabular}
	\end{minipage}}
\end{table*}

\begin{figure}[hbt!]
	\centering
	\subfigure[SWAE $(\beta^* = 0.5)$]
	{\includegraphics[width=0.22\textwidth]{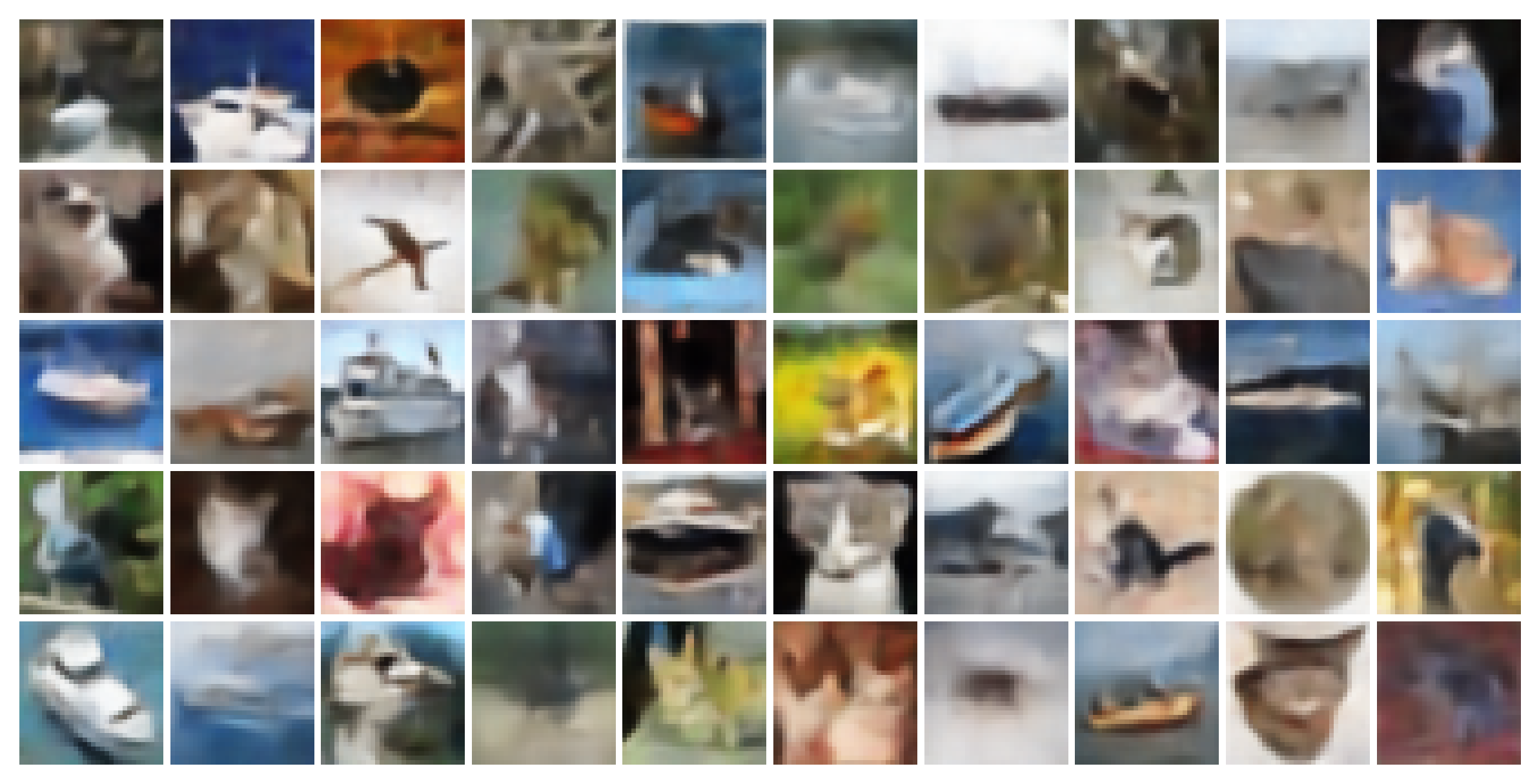}}
		\subfigure[SWAE $(\beta = 1)$]
	{\includegraphics[width=0.22\textwidth]{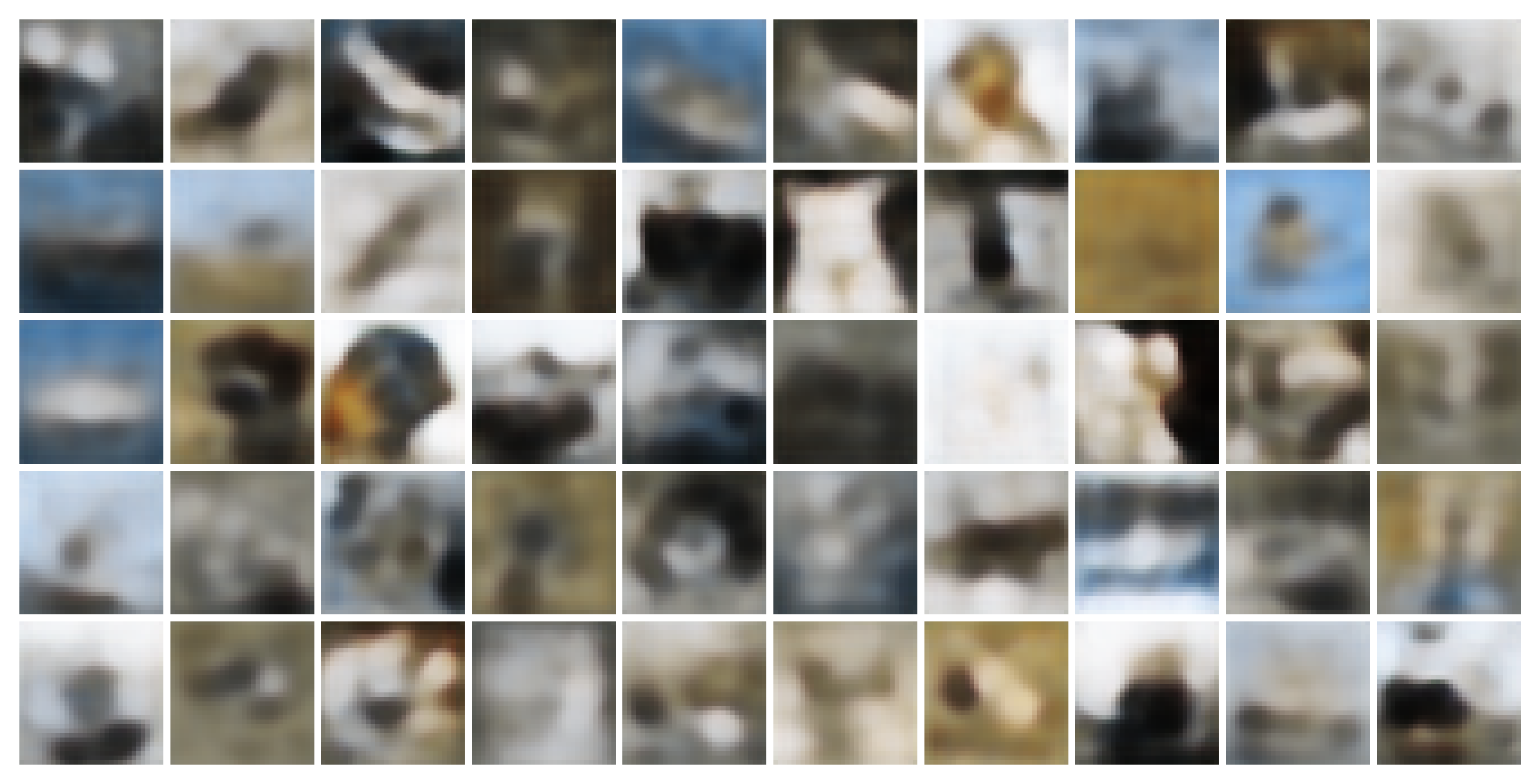}}
		\subfigure[SWAE $(\beta = 0)$]
	{\includegraphics[width=0.22\textwidth]{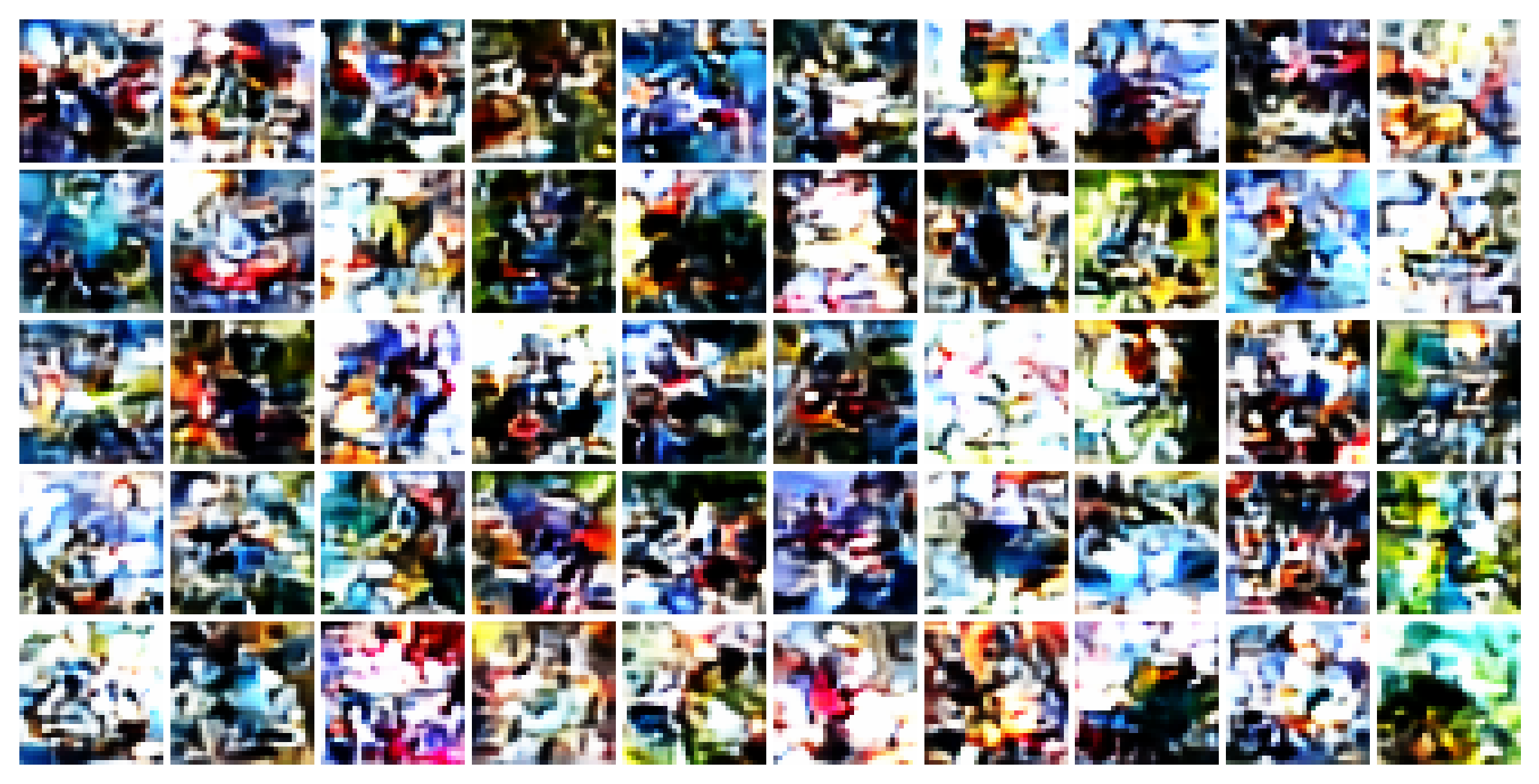}}
	\subfigure[VAE]
	{\includegraphics[width=0.22\textwidth]{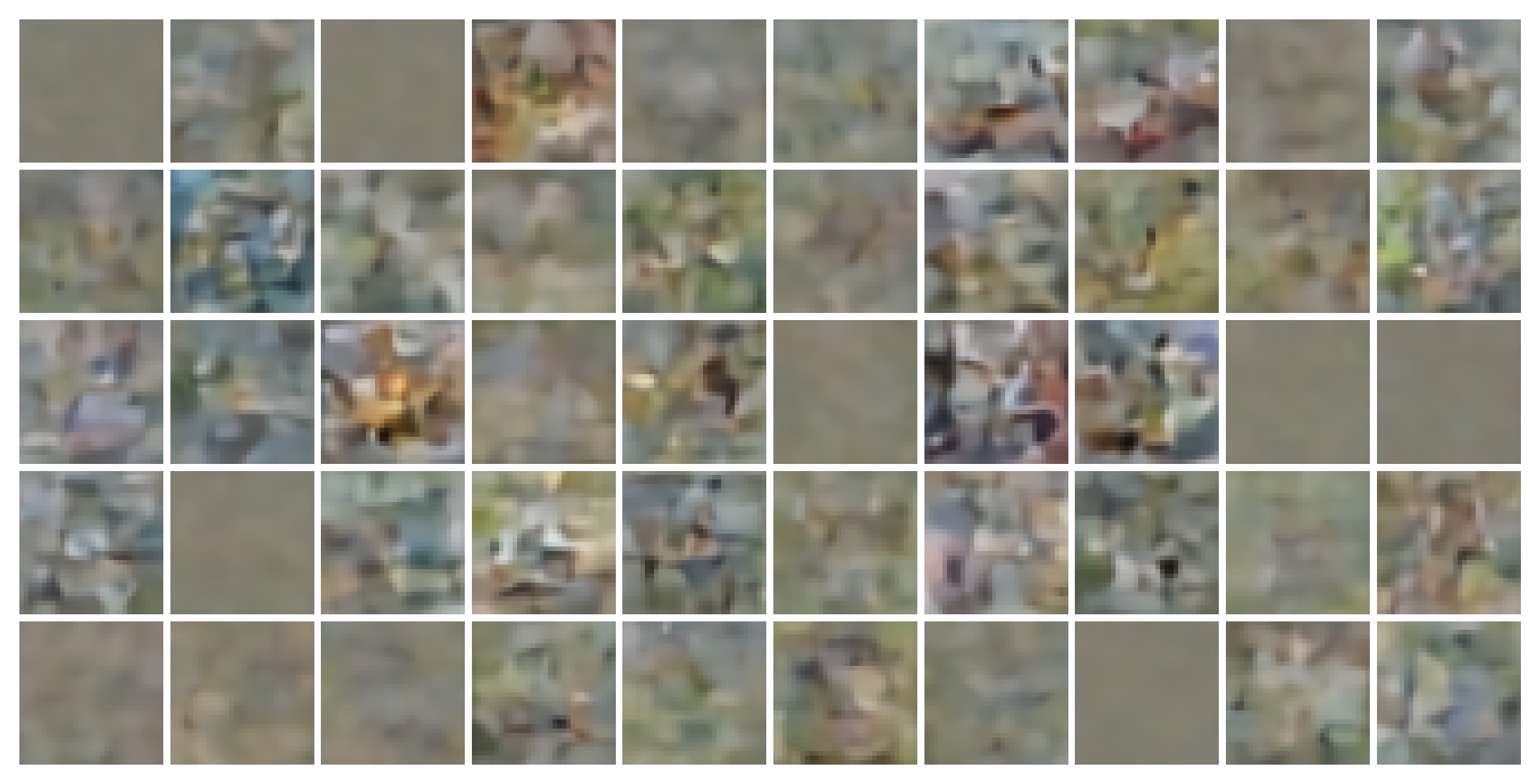}}	
	\subfigure[WAE-GAN]
	{\includegraphics[width=0.22\textwidth]{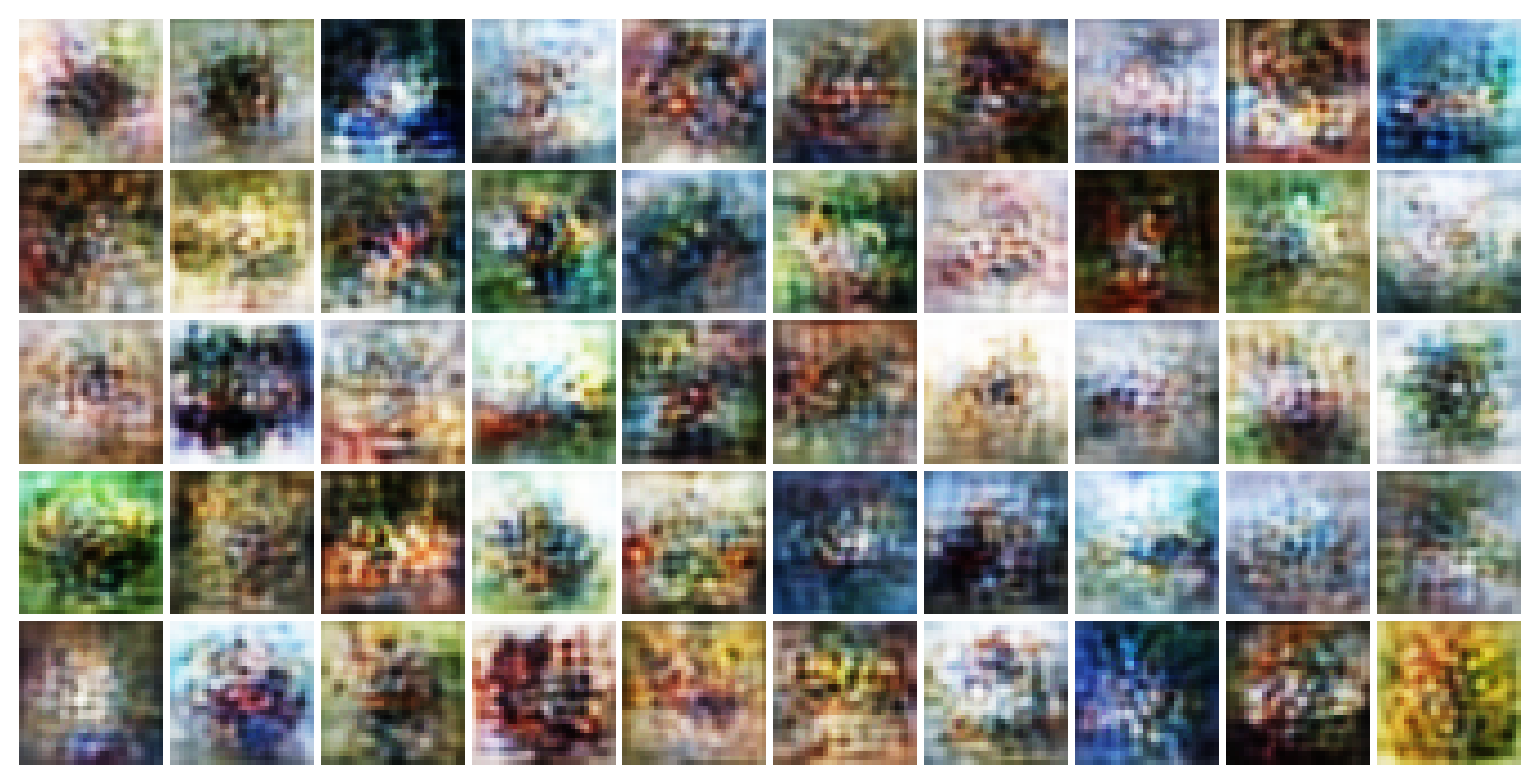}}	
	\subfigure[WAE-MMD]
	{\includegraphics[width=0.22\textwidth]{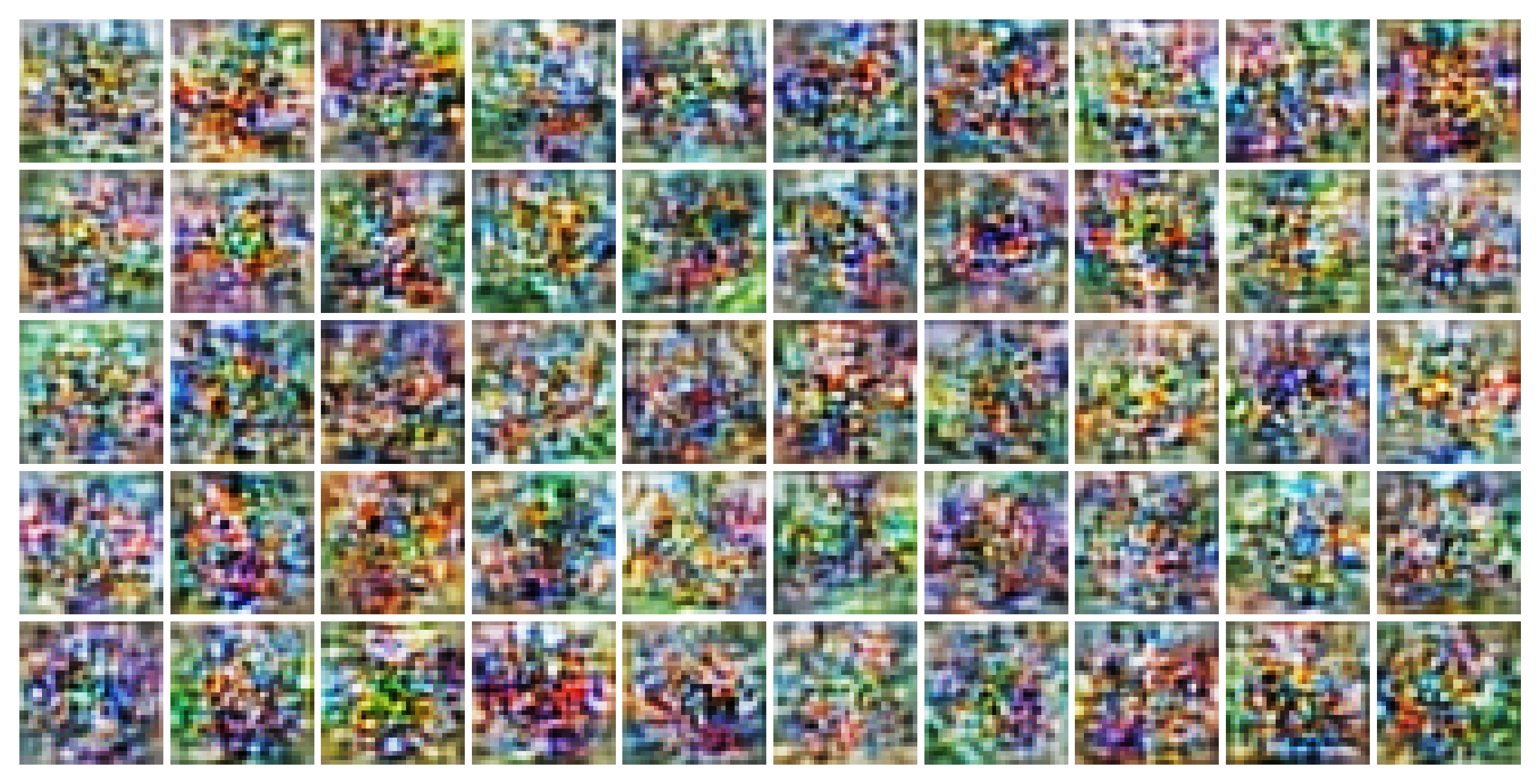}}	
	\subfigure[VampPrior]
	{\includegraphics[width=0.22\textwidth]{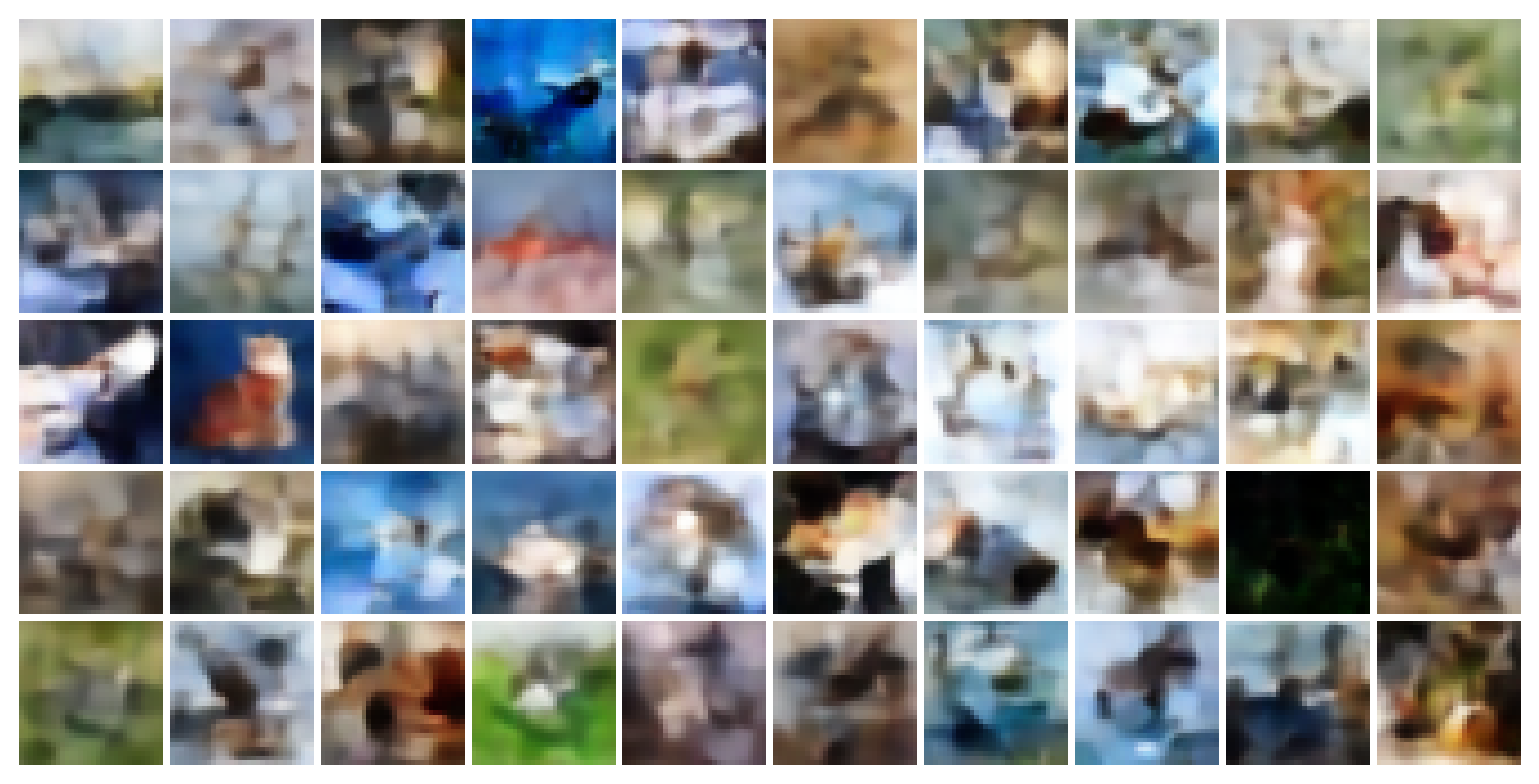}}
	\subfigure[MIM]
	{\includegraphics[width=0.22\textwidth]{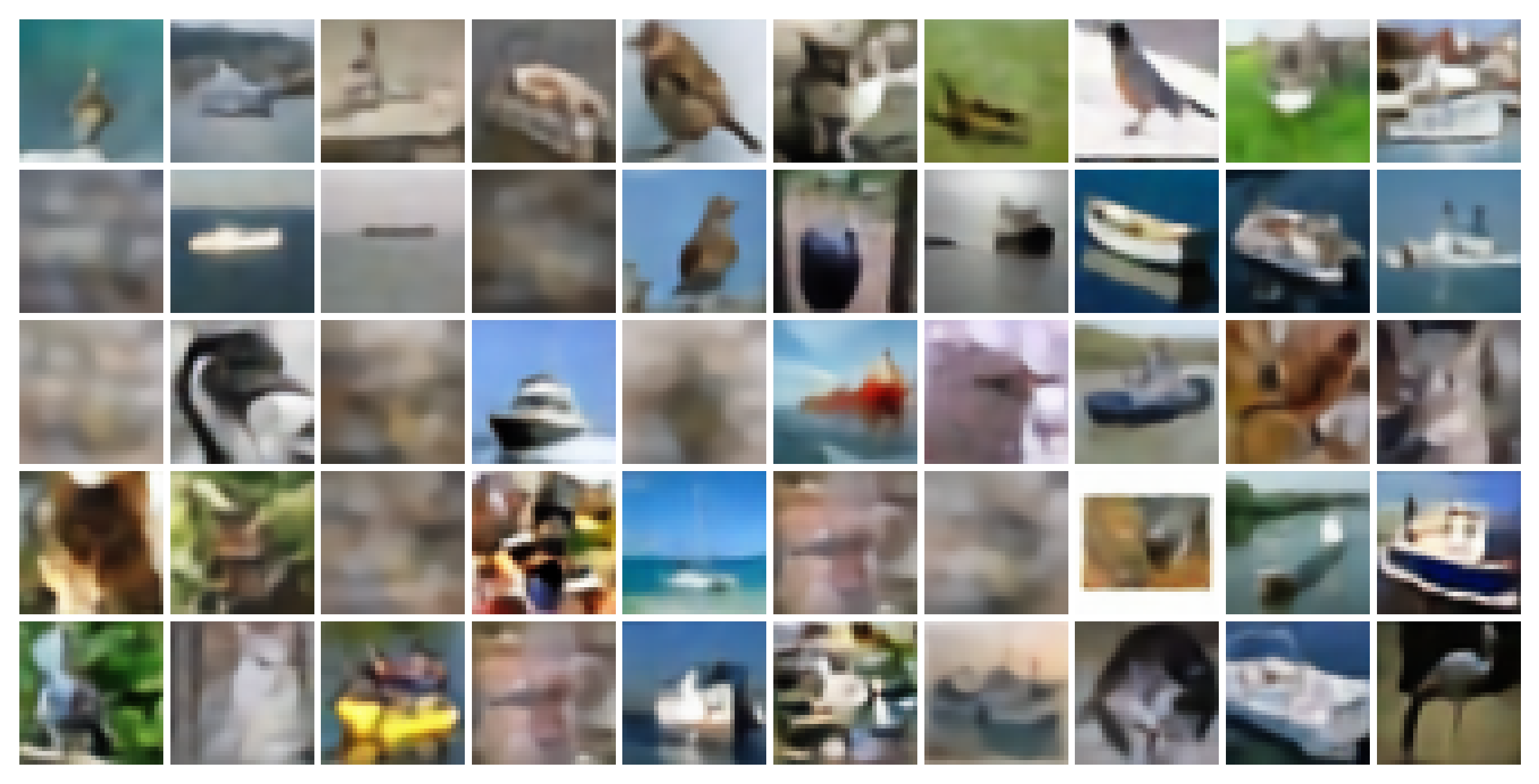}	}
	\caption{Generated new samples on CIFAR10-sub. dim-$\bz = 512$ for all methods.}
	\label{fig:cifarsub_gen}
\end{figure}

\begin{table*}[hbt!]
	\caption{Reconstruction loss (averaged over 5 trials). 
	}
	\label{tab:recon}
	\centering
	\setlength{\tabcolsep}{4pt}
	\resizebox{0.95\textwidth}
	{!}{\begin{minipage}{\textwidth}
	\begin{tabular}{lccccccccc }
		\toprule
		Dataset & dim-$\bz$ &  SWAE &  SWAE &VAE & WAE-GAN & WAE-MMD & VampPrior & MIM    \\
		 &  &  ($\beta = 1$) &  ($\beta = 0.5$) & &  &  &  &     \\
		\hline
		\multirow{3}{4em}{MNIST} &8 &30.11 \textpm \ 0.14  & \textbf{23.20 \textpm \ 0.04}& 24.34 \textpm \ 0.07& 26.86 \textpm \ 0.37& 24.76 \textpm \ 0.31& 24.05 \textpm \ 0.10& 24.04 \textpm \ 0.10\\
         & 40 & 26.29 \textpm \  0.13   & \textbf{6.93 \textpm \ 0.05}& 18.40 \textpm \  0.08& 16.06 \textpm \ 0.15 &13.78 \textpm \ 0.77 &17.32 \textpm \ 0.09 & 18.14 \textpm \ 0.33 \\
		&80 & 26.10 \textpm \  0.09& \textbf{1.25 \textpm \  0.02}& 18.50 \textpm \ 0.11& 10.78 \textpm \ 0.11& 9.63 \textpm \ 0.05& 17.42 \textpm \ 0.06 & 17.29 \textpm \ 0.20 \\
		\hline
		\multirow{3}{4em}{Fashion-MNIST} &8 &74.74 \textpm \ 0.04 &\textbf{71.03 \textpm \  0.06 }& 72.56 \textpm \ 0.02& 78.17 \textpm \ 1.41& 74.50 \textpm \ 0.60& 72.20 \textpm \ 0.04& 72.34 \textpm \ 0.03 \\
         & 40 & 73.39 \textpm \ 0.08& \textbf{57.90 \textpm \ 0.25}& 69.85 \textpm \ 0.04& 74.84 \textpm \ 0.23& 75.86 \textpm \ 0.41& 68.67 \textpm \ 0.07& 70.22 \textpm \ 0.87 \\
		&80 & 73.35 \textpm \ 0.08& \textbf{44.30 \textpm \ 0.71}& 69.90 \textpm \ 0.08& 70.74 \textpm \ 1.16& 71.28 \textpm \ 3.80& 68.54 \textpm \ 0.10& 69.10 \textpm \ 0.13 \\
		\hline
		\multirow{3}{4em}{Coil20} &8 & 7.07 \textpm \ 0.64& \textbf{5.69 \textpm \ 0.51}& 7.90 \textpm \ 0.36& 8.14 \textpm \ 0.34& 21.20 \textpm \ 15.30& 8.17 \textpm \ 1.02& 13.84 \textpm \ 3.82 \\
         & 40 & 5.52 \textpm \ 0.40 & \textbf{4.27 \textpm \ 0.66}& 5.67 \textpm \ 0.42& 5.82 \textpm \ 0.84& 8.07 \textpm \ 7.80& 6.31 \textpm \ 0.62& 5.75 \textpm \ 0.77 \\
		&80 & 5.56 \textpm \ 0.30& \textbf{4.33 \textpm \ 0.40}& 5.71 \textpm \ 0.67& 5.62 \textpm \ 1.26& 5.83 \textpm \ 1.92& 6.32 \textpm \ 0.37& 5.87 \textpm \ 0.69 \\
		\hline
		\multirow{1}{6em}{CIFAR10-sub} &512 & 50.82\textpm 3.78 & \textbf{6.50\textpm 0.08}  & 9.41\textpm 0.27& 13.37\textpm 1.62& 13.09\textpm 1.92& 12.06 \textpm 0.91& 10.02 \textpm 0.37
		\\
		\bottomrule
	\end{tabular}
	\end{minipage}}
\end{table*}

\subsection{Latent Representation}
The latent representation is expected to capture salient features 
of the observed data and be useful for the downstream applications. The considered datasets are all associated with the labels. 
In the experiment we use the latent representation for the K-Nearest Neighbor (KNN) classification and compare the classification accuracy of $5$-NN in Table \ref{tab:5nn}, where dim-$\bz$ denotes the dimension of the latent space. The results of $3$-NN and $10$-NN are similar  to those of 5-NN and thus are omitted. We found that the classification results of all algorithms on CIFAR10 are unsatisfactory based on the current networks 
(accuracy was around $0.3 - 0.4$; and this may due to the limited expressive power of the shallow network architectures used),  
so instead we create a subset of CIFAR10 (CIFAR10-sub) which contains $3$ classes: bird, cat, and ship. 

Since the prior of VAE, WAE-GAN, and WAE-MMD is an isotropic Gaussian, setting dim-$\bz$ greater than the intrinsic dimensionality of the observed data would force $p(\bz_e)$ to be in a manifold in the latent space \citep{tolstikhin2018wasserstein}. This  makes it impossible to match the marginal $p(\bz_e)$ with the prior $p(\bz_d)$ and thus leads to  unsatisfactory latent representation. Such concern can be verified particularly on Fashion-MNIST  where the classification accuracy of VAE and WAE-GAN drops dramatically 
when dim-$\bz$ is increased. 
For SWAE, we consider two cases: $\beta = 1$ (\ie, without the reconstruction loss) and $\beta = 0.5$.  The classification accuracy of SWAE $(\beta = 1)$ is comparable to SWAE $(\beta = 0.5)$ and is generally  superior for different values of dim-$\bz$ to the benchmarks.

To further show the structure of the latent representation, we project the latent representation to 2D using t-SNE \citep{maaten2008visualizing} as the visualization tool. As an example, we show the projection of the latent representation on MNIST in Figure \ref{fig:mnist_latent}.
We can see that SWAEs keep the local structure of the observed data in the latent space and lead to tight clusters, which is consistent to our expectation as explained in Section \ref{subsec:ot}.

\subsection{Generation and Reconstruction}
To generate new data, latent samples are first drawn from the marginal prior distribution $p(\bz_d)$ based on the conditional priors $p(\bz_d|\bu_k)$, and are then fed to the decoder. We display the generated new samples of all methods on CIFAR10-sub in Figure \ref{fig:cifarsub_gen}  and put  the generated images on the other datasets  in the supplementary file. We show the Fr\'{e}chet Inception Distance (FID) \citep{heusel2017gans}, which is commonly used for evaluating the quality of generated images,  in Table \ref{tab:fid}. For SWAEs, we observe that the reconstruction loss term is crucial for improving the generation quality as SWAE ($\beta = 1$) generally cannot lead to the lowest FID.  On MNIST and Fashion-MNIST,  the FID of the best SWAE (indicated as $\beta^*$) is 
slightly higher than that of WAE-GAN, but lower than all the other benchmarks. The visual difference between SWAE ($\beta^*$) and WAE-GAN on MNIST and Fashion-MNIST is however negligible. In Section \ref{subsec:ot}, we compare the formulation of SWAEs ($\beta = 1$) with WAE. In particular, the objective of WAE includes a distribution-based dissimilarity in the latent space while the $\bz$-loss in SWAEs measures the sample-based dissimilarity. 
On Coil20 and CIFAR10-sub, SWAE ($\beta^*$) achieves the lowest FID and generates new images that are visually much better than those generated by the benchmarks.


In Table \ref{tab:recon}, we compare the reconstruction loss, defined as $\|\bx_e - D(\bz_e)\|_2^2$, on the four datasets. As expected, increasing the value of dim-$\bz$ can reduce the reconstruction loss but the reduction becomes marginal when dim-$\bz$ is large enough. Additionally, since a smaller value of $\beta$ leads to more emphasis on the reconstruction-based loss the quality of reconstruction is generally better.
We observe that SWAE ($\beta = 0.5$) results in the lowest reconstruction loss in all cases. 
The reconstructed images of all methods on CIFAR10-sub are shown in Figure \ref{fig:cifarsub_rec}. Results on other datasets are provided in the supplementary file for reference. Without including the reconstruction loss into the objective, the reconstruction quality of SWAE ($\beta = 1$) can be unsatisfactory (\eg, on CIFAR10-sub).

\begin{figure}[h]
	\centering
	\subfigure[Real images]
	{\includegraphics[width=0.22\textwidth]{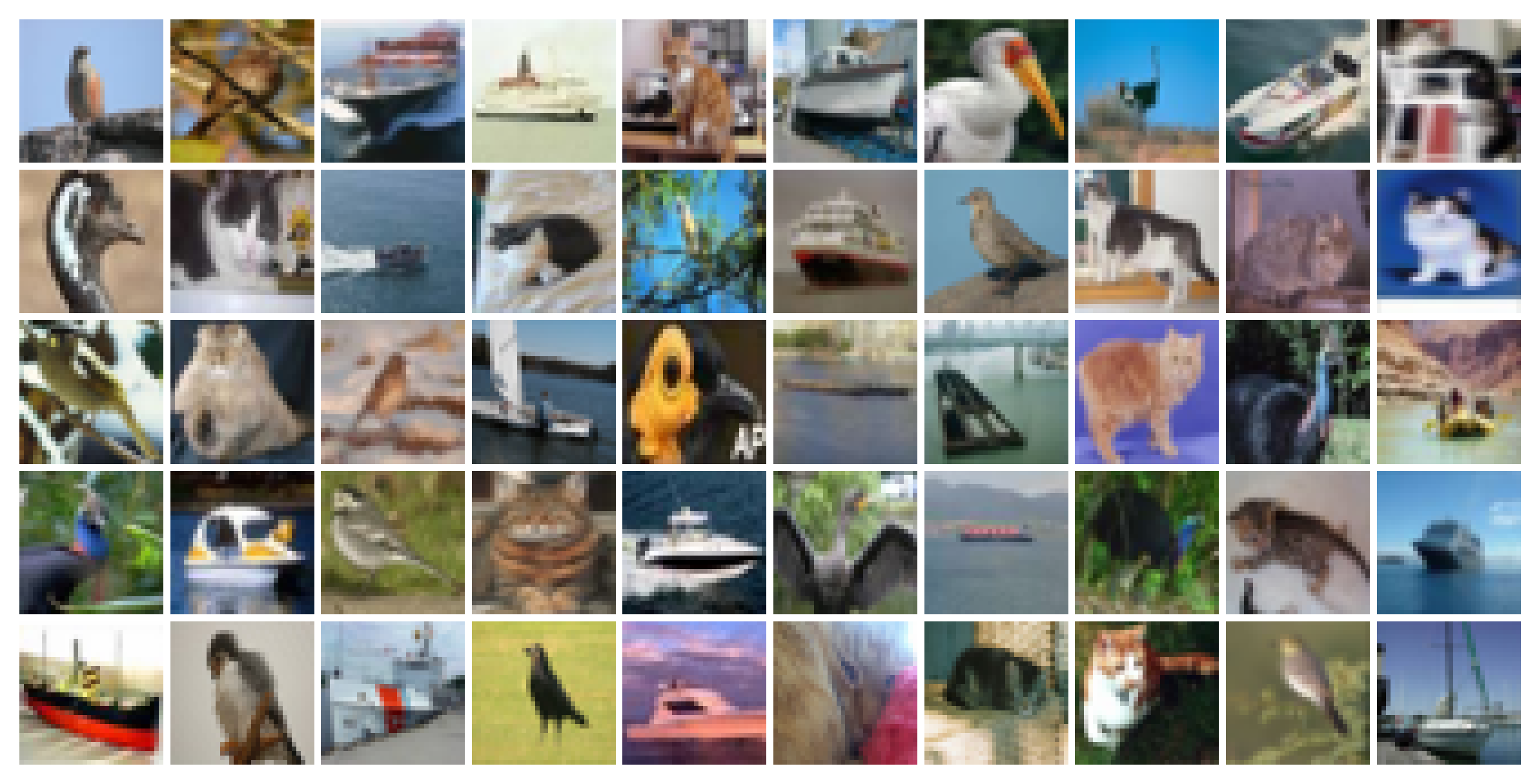}}	
	\subfigure[SWAE $(\beta = 1)$]
	{\includegraphics[width=0.22\textwidth]{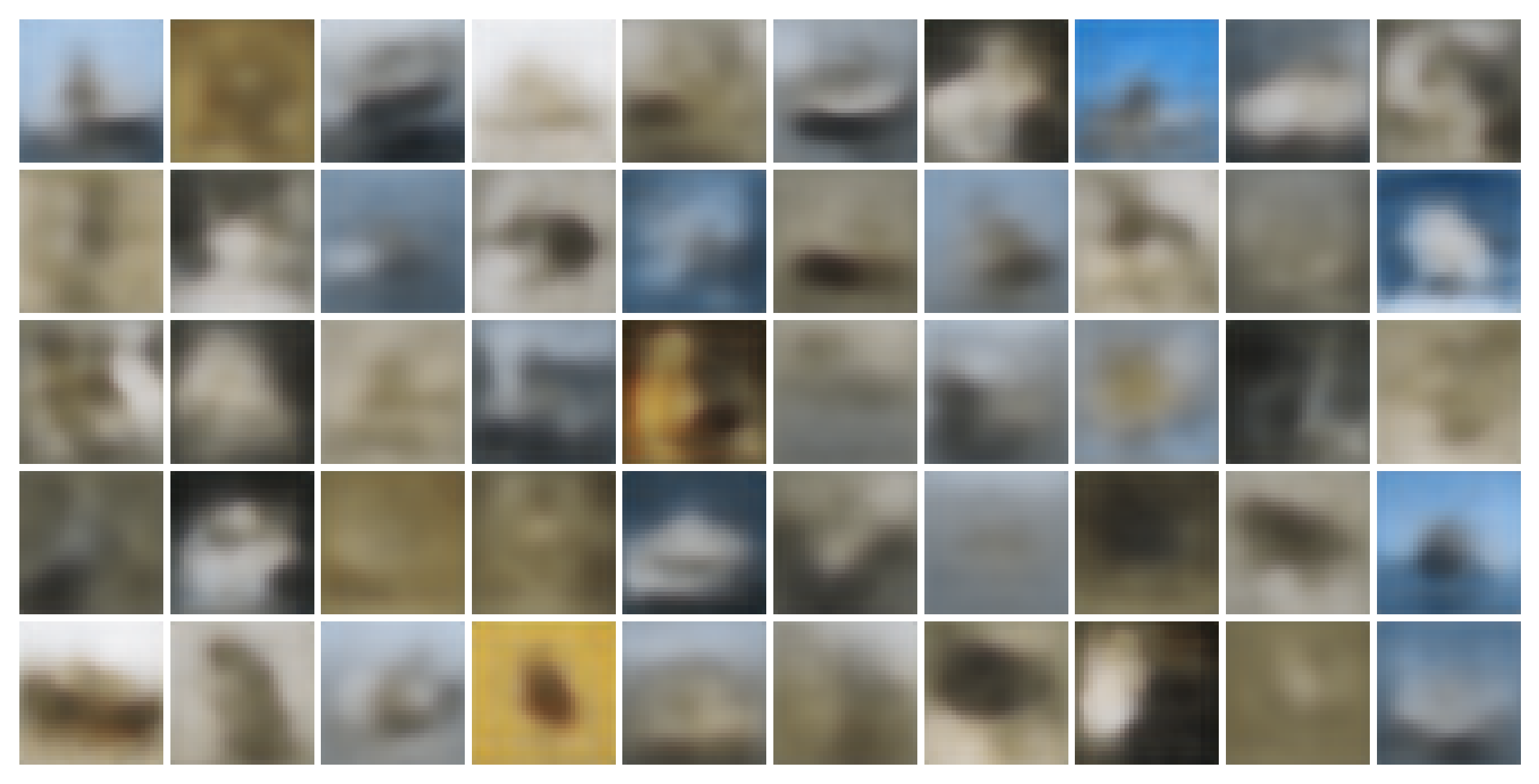}}	
	\subfigure[SWAE $(\beta = 0.5)$]
	{\includegraphics[width=0.22\textwidth]{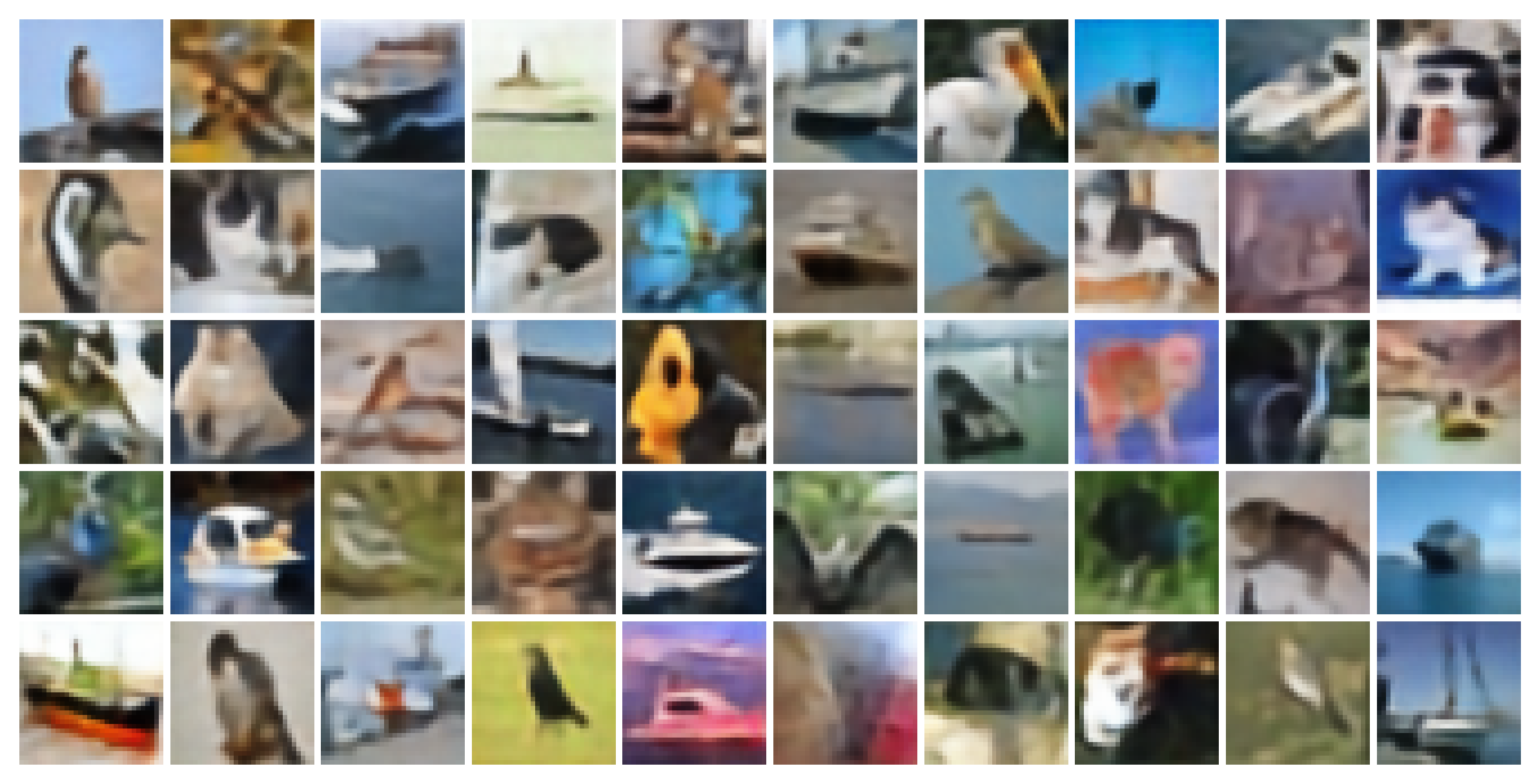}}	
	\subfigure[VAE]
	{\includegraphics[width=0.22\textwidth]{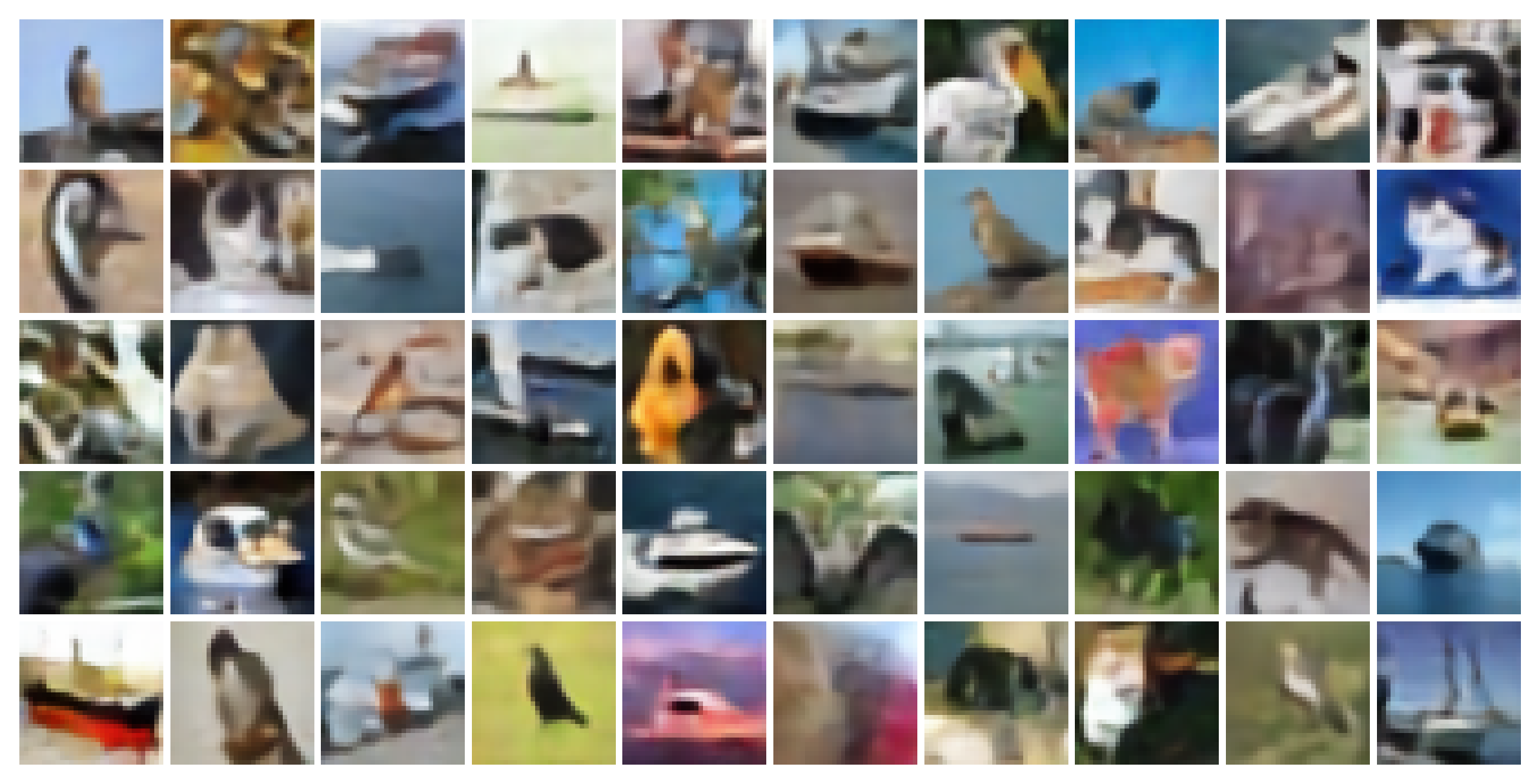}}
	\subfigure[WAE-GAN]
	{\includegraphics[width=0.22\textwidth]{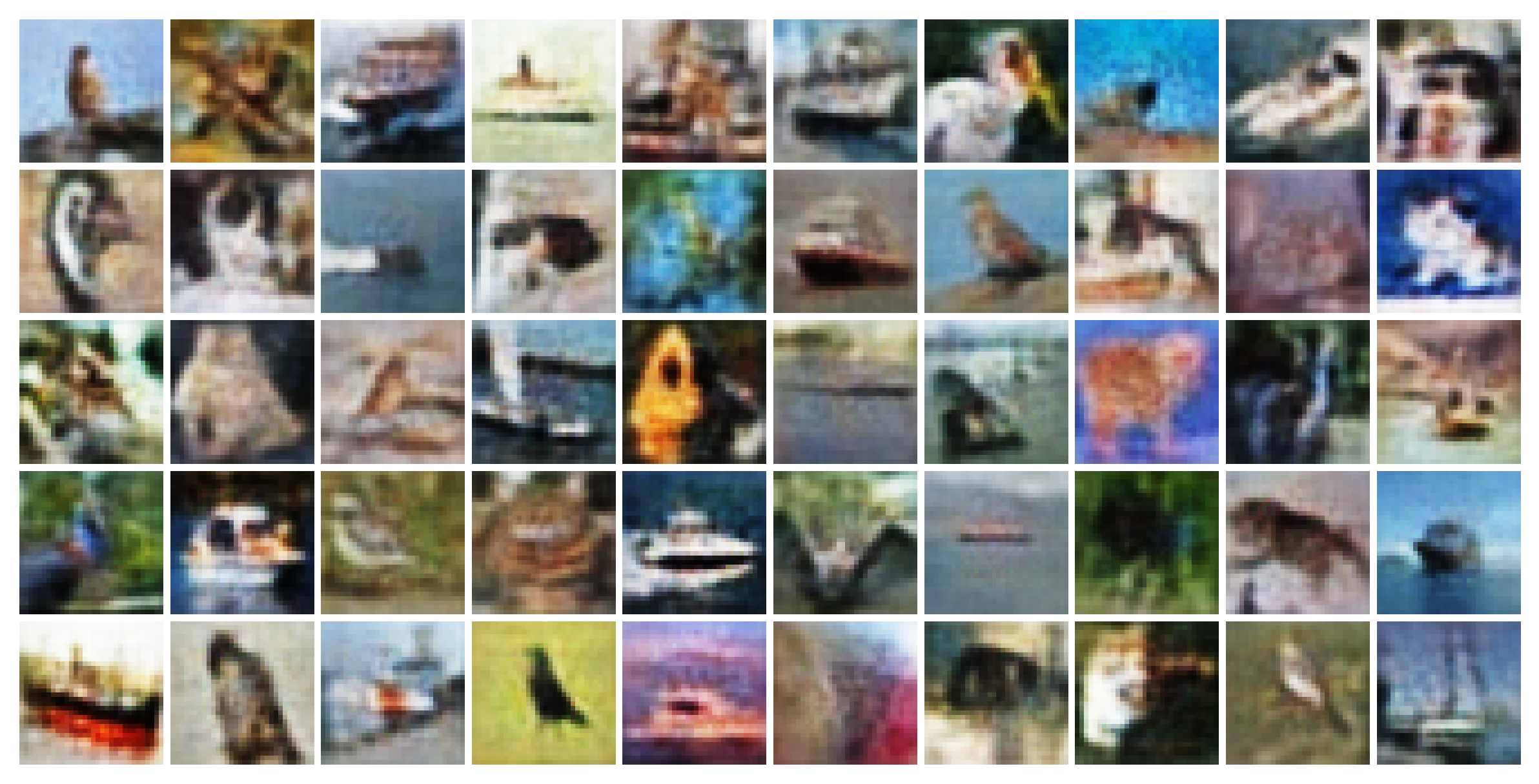}}
	\subfigure[WAE-MMD]
	{\includegraphics[width=0.22\textwidth]{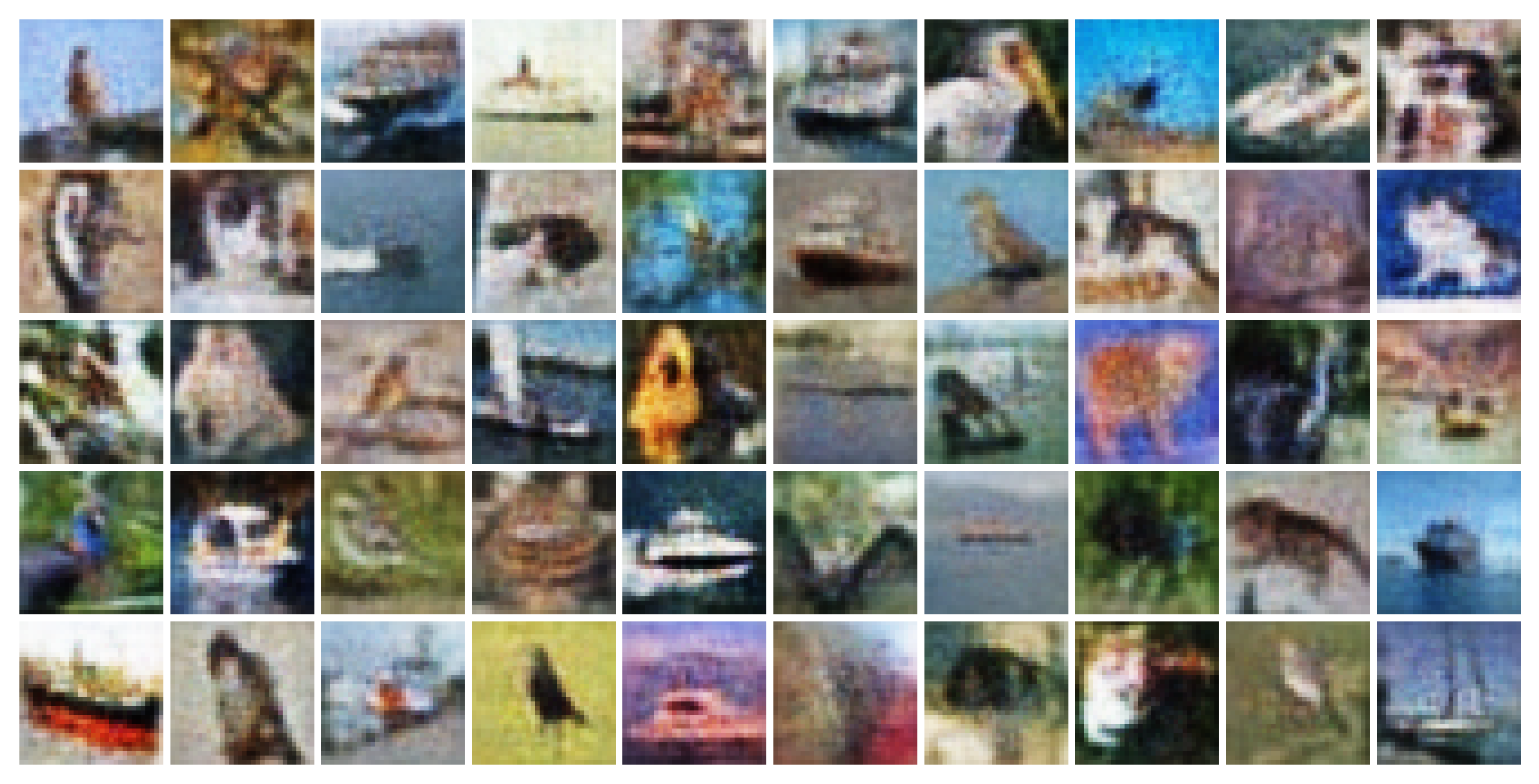}}
	\subfigure[VampPrior]
	{\includegraphics[width=0.22\textwidth]{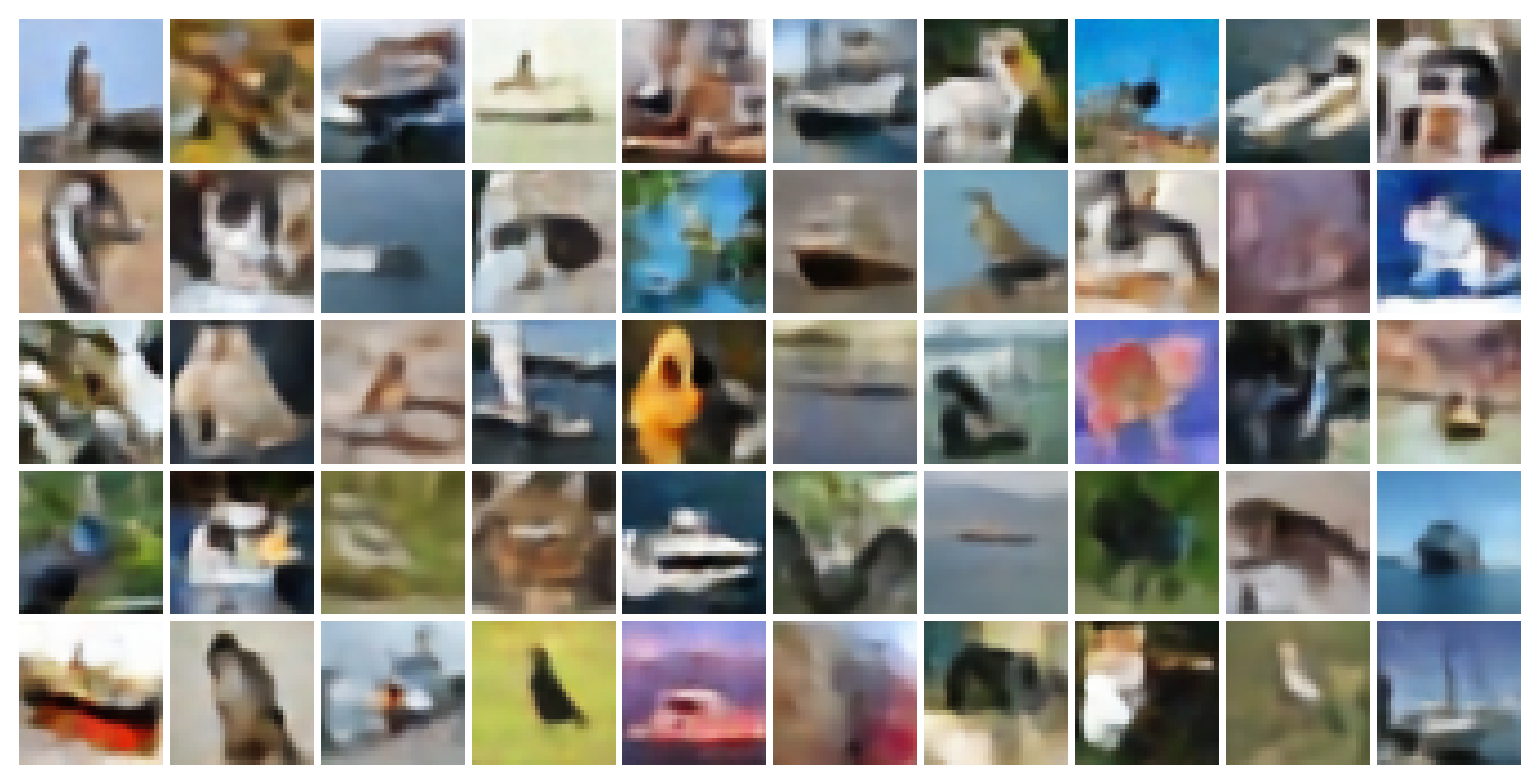}}
	\subfigure[MIM]
	{\includegraphics[width=0.22\textwidth]{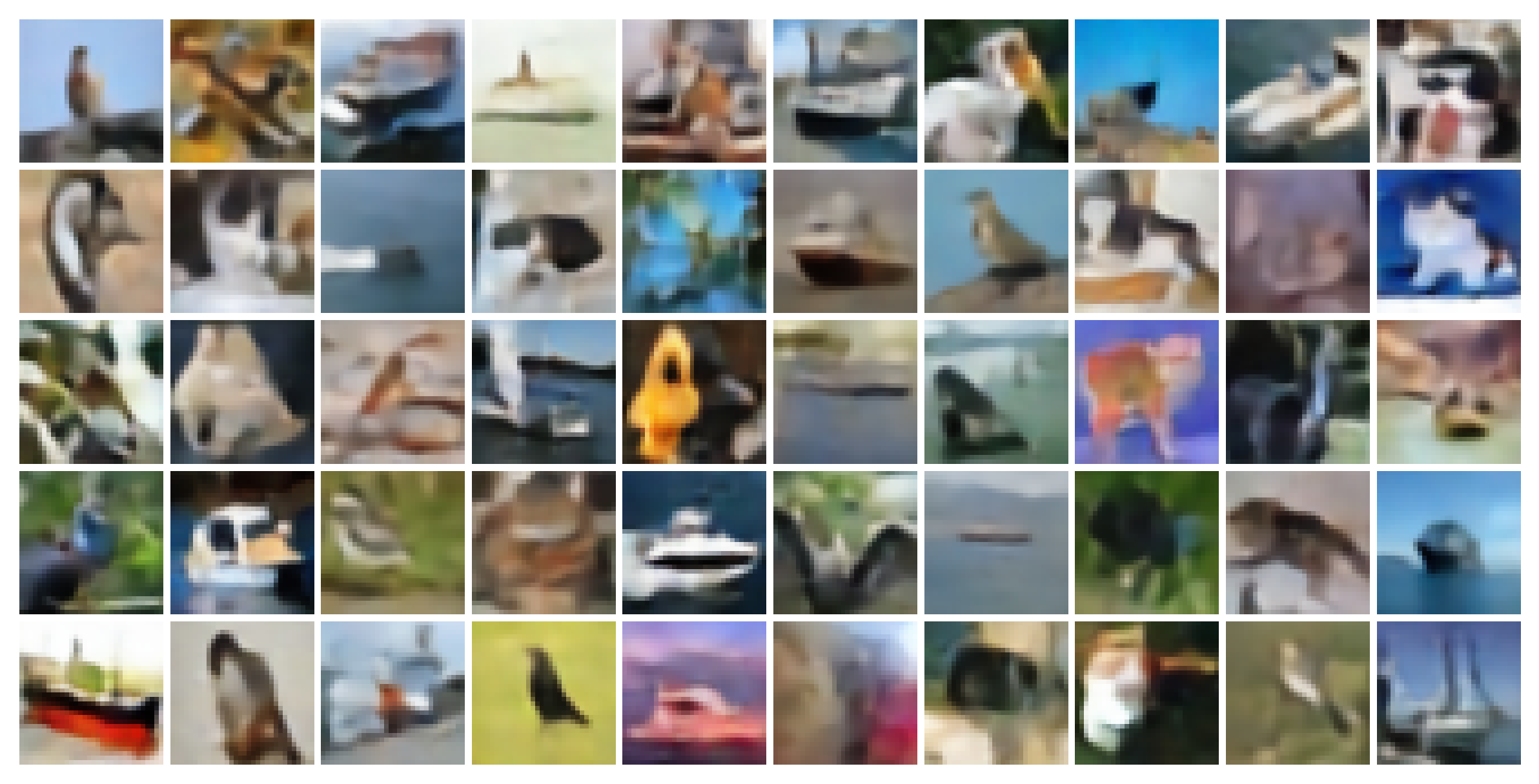}}
	\caption{Reconstructed images on CIFAR10-sub.  dim-$\bz = 512$ for all methods. 
Excluding  the reconstruction loss in the objective, the reconstruction of SWAE $(\beta = 1)$ is blurry. }
\label{fig:cifarsub_rec}
\end{figure}

\subsection{Denoising Effect with SWAE ($\beta = 1$)}

\begin{figure}[hbt!]
	\centering
	\subfigure[Noisy real images]
	{\includegraphics[width=0.22\textwidth]{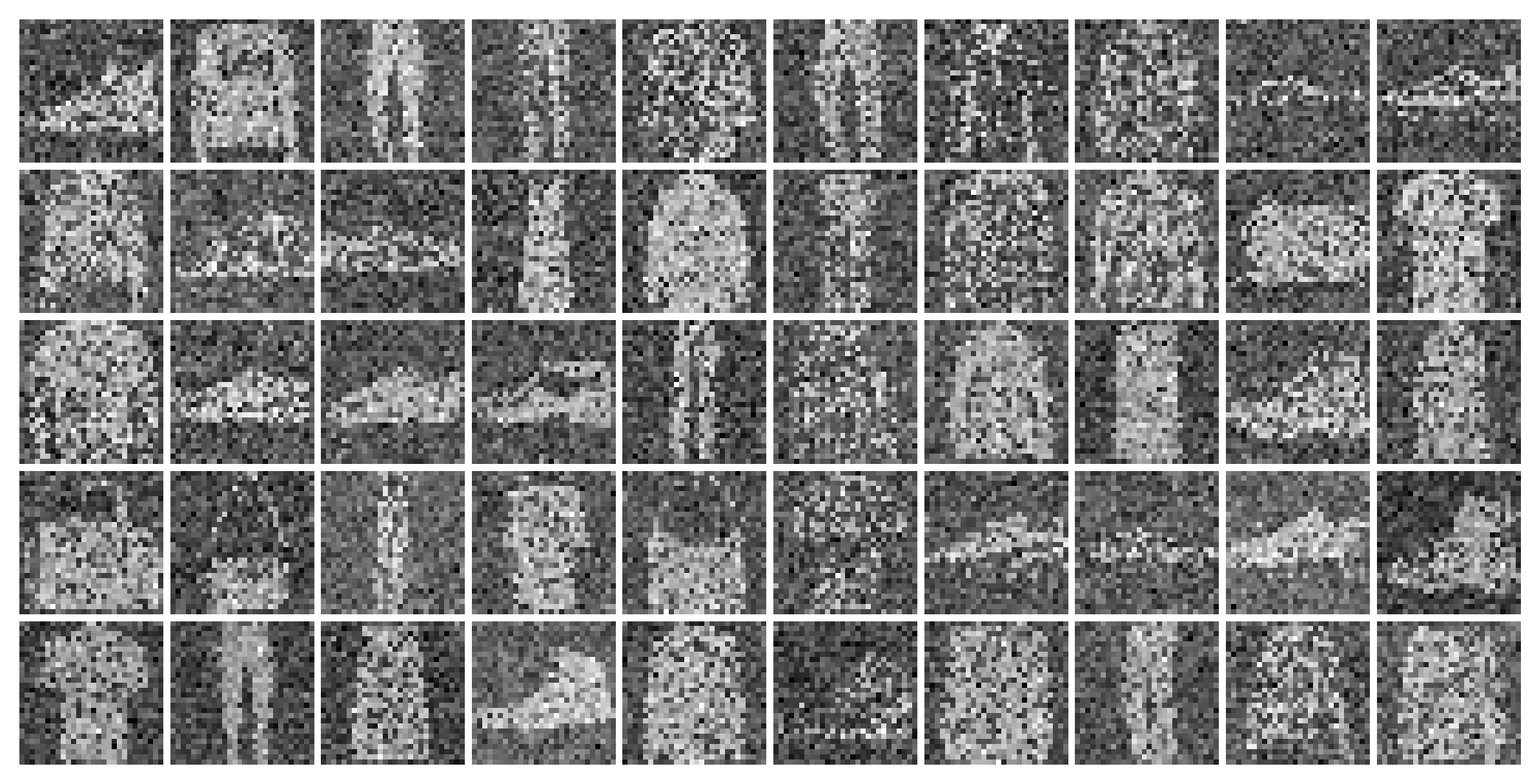}}	
	\vspace{-1mm}
	\subfigure[SWAE ($\beta = 1$)]
	{\includegraphics[width=0.22\textwidth]{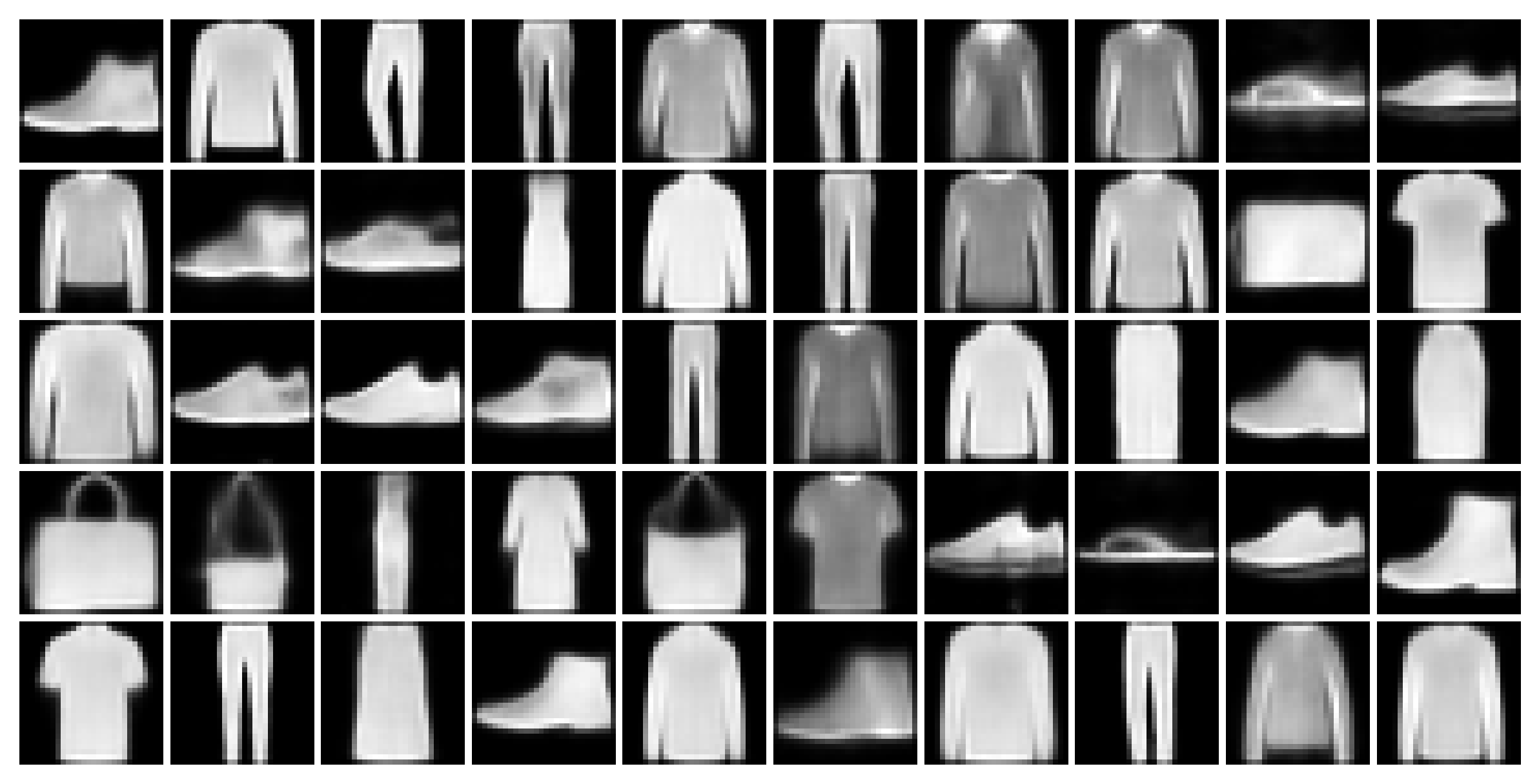}}	
	\vspace{-1mm}
	\subfigure[SWAE ($\beta = 0.5$)]
	{\includegraphics[width=0.22\textwidth]{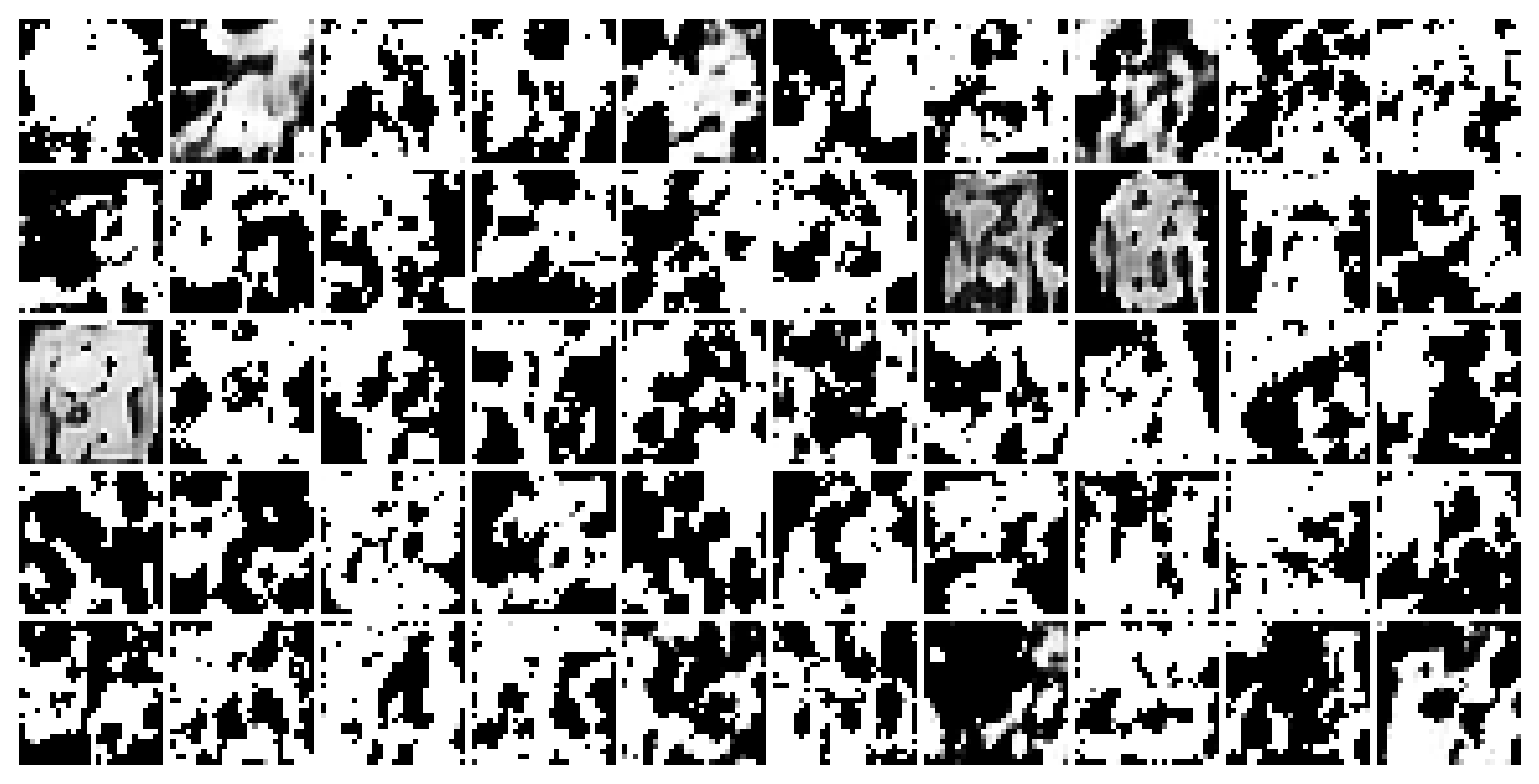}}	
	\vspace{-1mm}
	\subfigure[VAE]
	{\includegraphics[width=0.22\textwidth]{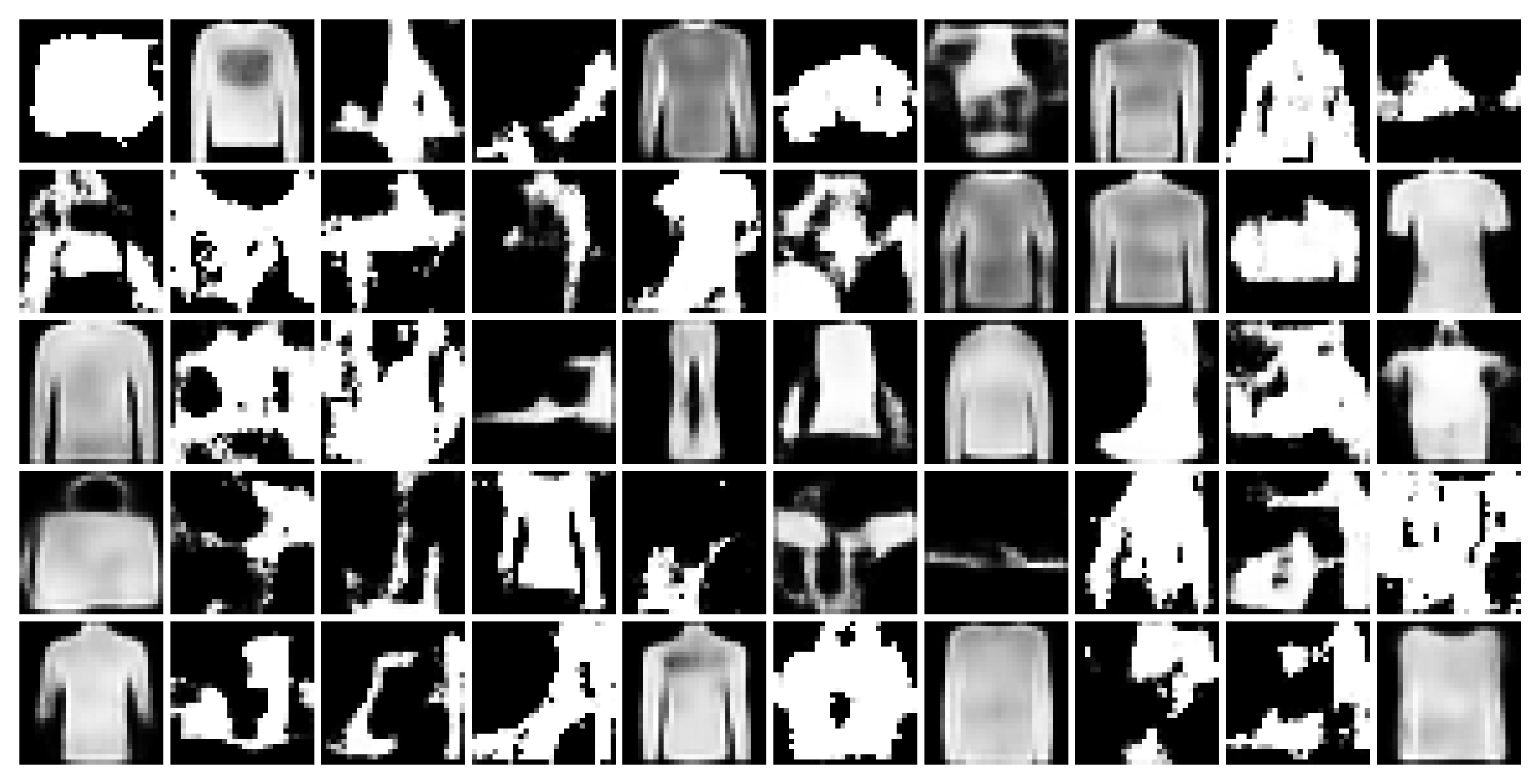}}	
	\vspace{-1mm}
	\subfigure[WAE-GAN]
	{\includegraphics[width=0.22\textwidth]{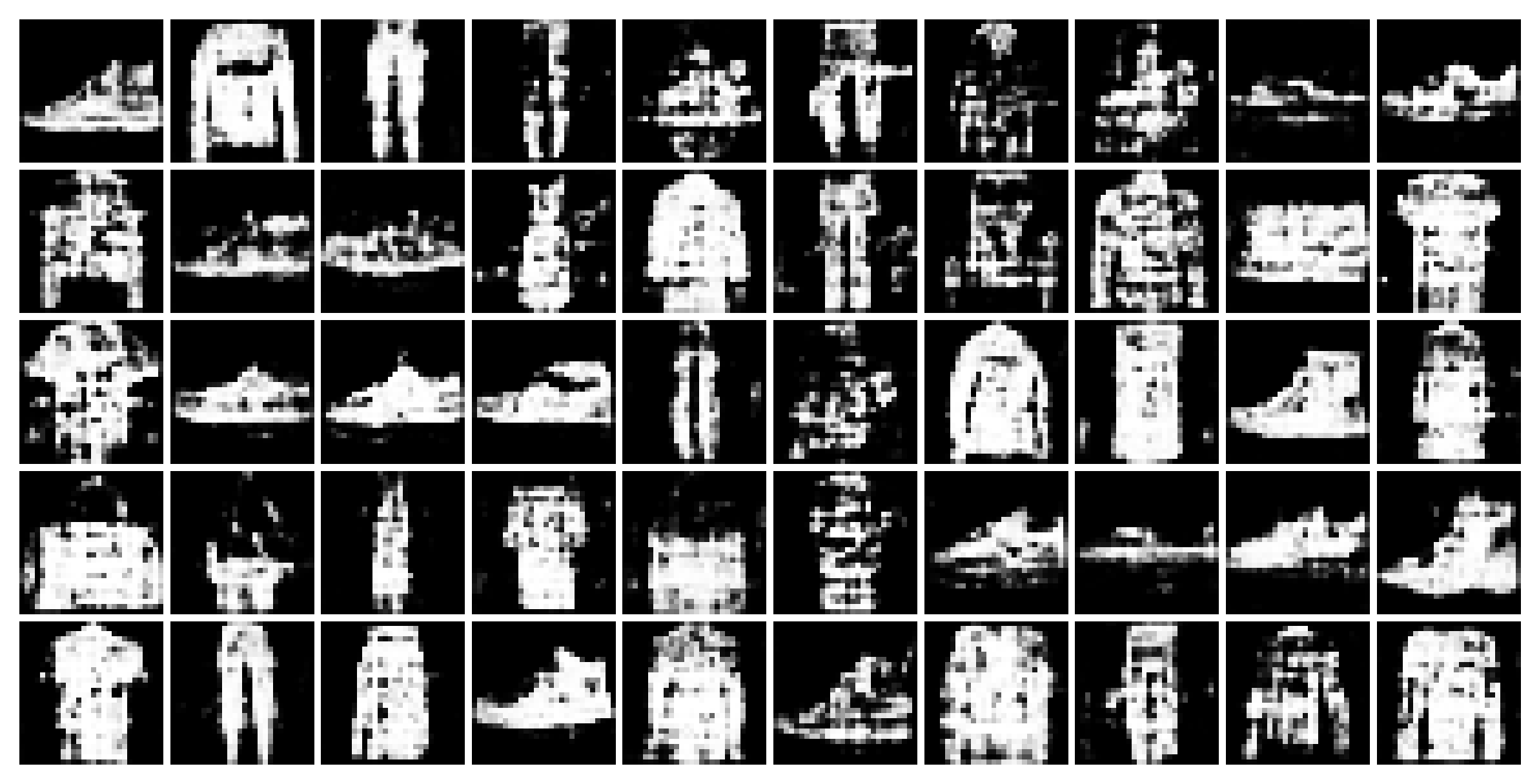}}	
	\subfigure[WAE-MMD]
	{\includegraphics[width=0.22\textwidth]{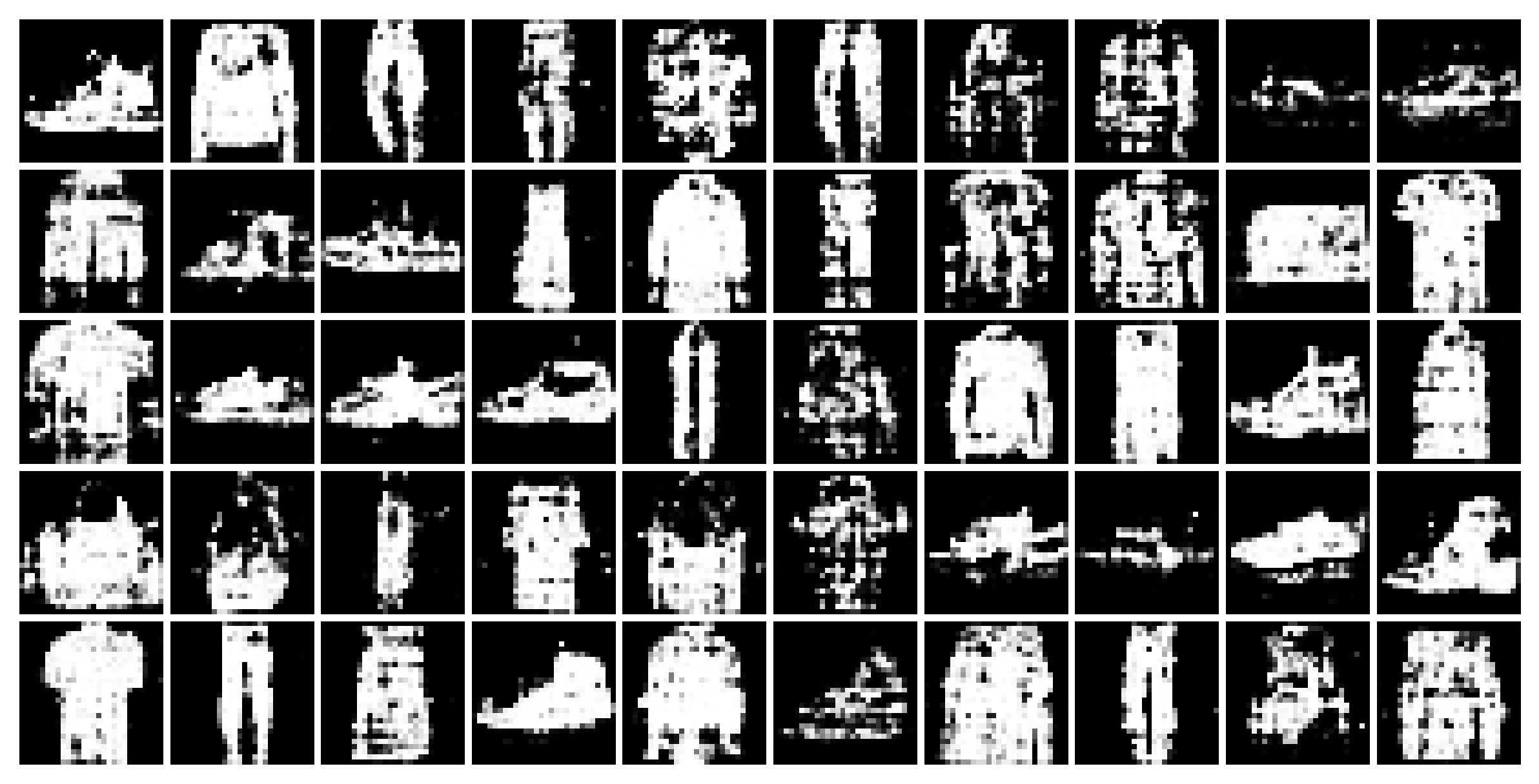}}	
	\subfigure[VampPrior]
	{\includegraphics[width=0.22\textwidth]{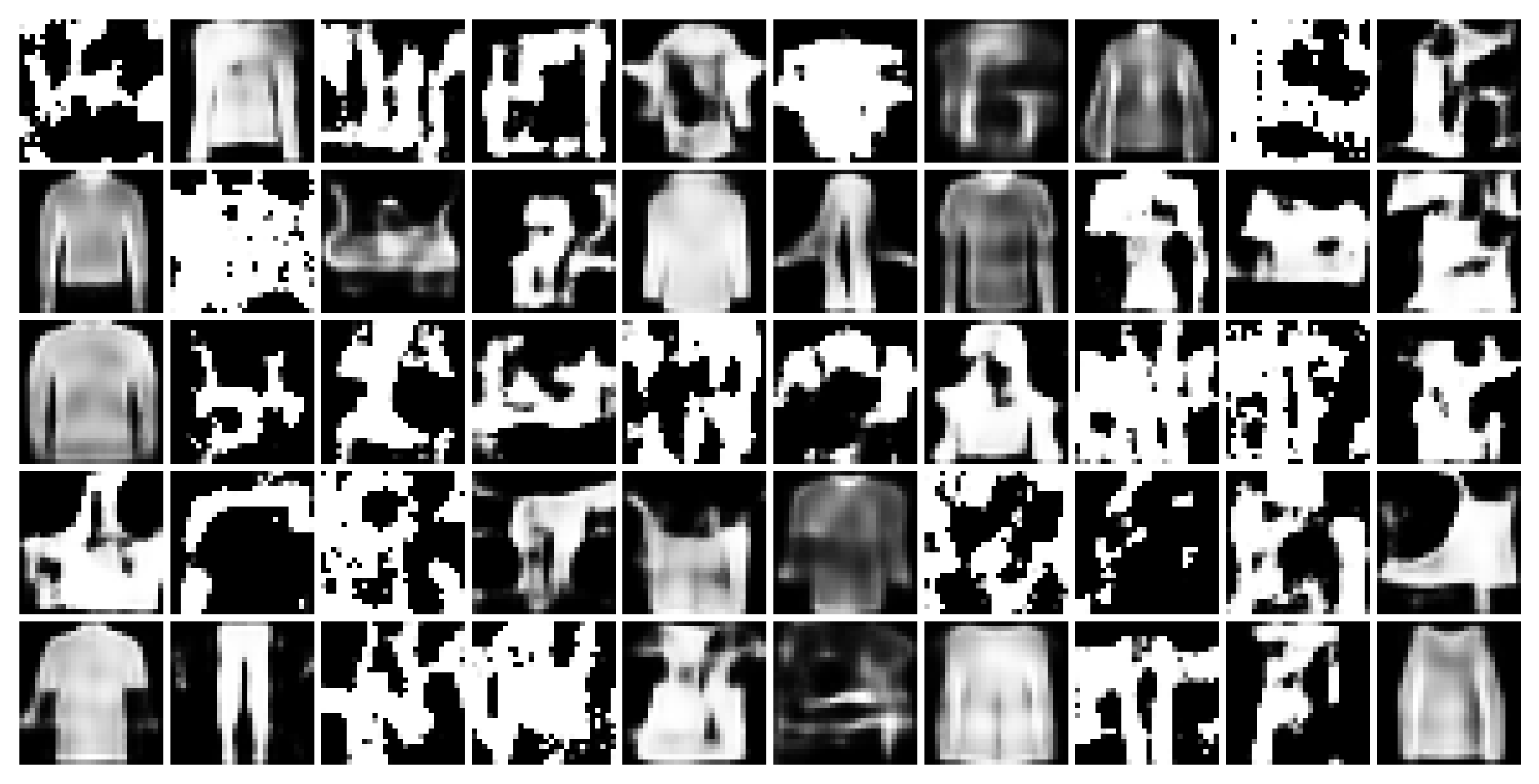}}
	\subfigure[MIM]
	{\includegraphics[width=0.22\textwidth]{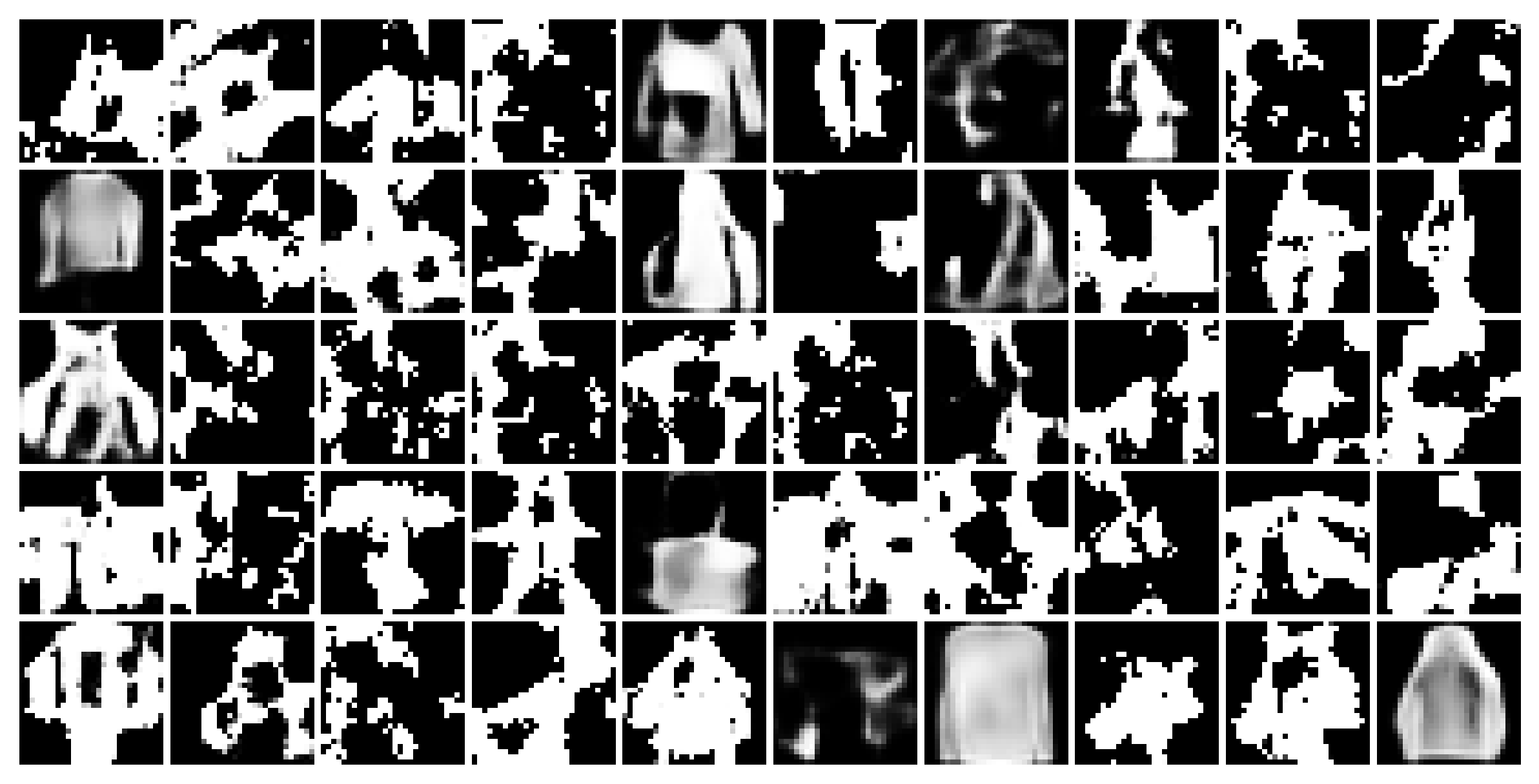}}
	\caption{Denoising effect: reconstructed images on Fashion-MNIST.  dim-$\bz = 80$ for all methods.}
	\label{fig:fash_denoise}
\end{figure}

As discussed in Section \ref{subsec:ot}, the $\bx$-loss has a close relationship to the objective of Denoising Autoencoders (DAs). After training, we feed the noisy images, which are obtained by adding the Gaussian random samples with 
mean zero and 
standard deviation $0.3$ to the clean test samples, to the encoder. 
In Figure \ref{fig:fash_denoise}, as an example, we show the reconstructed images  on Fashion-MNIST. Since the reconstruction loss is highly related to the dimension of the latent space, for fair comparison we set dim-$\bz$ to $80$ for all methods. We observe that only SWAE ($\beta = 1$) can recover clean images. This observation confirms the denoising effect induced by the $\bx$-loss, and thus the resultant latent representation is robust to partial destruction of the observed data. 

\vspace{-2mm}
\section{Related Work}
\vspace{-2mm}


The objective of VAEs uses the asymmetric KL divergence between the encoding and the decoding distributions.  To improve VAEs 
 \citep{livne2019mim,chen2018symmetric,pu2017adversarial} propose symmetric divergence measures 
instead of the asymmetric KL divergence in VAE-based generative models. For example, 
MIM \citep{livne2019mim} adopts the Jensen-Shannon (JS) divergence
between the encoding and the decoding distributions together with a regularizer maximizing the mutual information between the data and the latent representation. Due to the difficulty of estimating the mutual information and the unavailability of the data distribution, an upper bound of the desired loss is proposed.
AS-VAE \citep {pu2017adversarial} and the following work \citep{chen2018symmetric} propose a symmetric form of the KL divergence optimized with adversarial training. 
These methods typically  involve a difficult objective either depending on (unstable) adversarial training or containing the mutual information that requires further approximation. In contrast, the proposed SWAEs yield a simple expression of 
objective and do not involve adversarial training. 


Compared to VAEs, GANs lack an efficient inference model thus are incapable of providing the corresponding latent representation given the observed data. To bridge the gap between VAEs and GANs, recent works attempt to integrate an inference mechanism into GANs by symmetrically treating the observed data and the latent representation, \ie, the discriminator is trained to discriminate the joint samples in both the data and the latent spaces. In particular, the JS divergence between the encoding and the decoding distributions is deployed 
in ALI \citep{dumoulin2016adversarially} and BiGANs \citep{donahue2016adversarial}. 
To address the non-identifiability issue in ALI (\eg, unfaithful reconstruction), later ALICE \citep{li2017alice} proposes to regularize ALI using conditional entropy.

Generative modelling is closely related to minimizing a dissimilarity measure between two distributions. 
As opposed to many other commonly adopted dissimilarity measures, \eg, the JS and the KL divergences, the Wasserstein distances that arise from the OT problem provide a weaker distance  between probability distributions (see \citep{santambrogio2015optimal,peyre2019computational,kolouri2017optimal} for more background on OT). This is crucial as in
many applications the observed data are essentially supported on a low dimensional manifold. In such cases, common dissimilarity measures 
may fail to provide a useful gradient for training.
Consequently, the Wasserstein distances have received a 
surge 
of attention for learning generative models \citep{arjovsky2017wasserstein, balaji2019entropic,sanjabi2018convergence,kolouri2018sliced,patrini2019sinkhorn,tolstikhin2018wasserstein,deshpande2019max,nguyen2020distributional}.
Particularly,  the VAE-based models \citep{tolstikhin2018wasserstein,kolouri2018sliced,patrini2019sinkhorn} are all based on minimizing the OT cost of the marginal distributions in the data space  with the difference of how to measure the divergence in the latent space: \citep{tolstikhin2018wasserstein} proposes the GAN-based and the MMD-based divergences, \citep{kolouri2018sliced} adopts the sliced-Wasserstein distance, and \citep{patrini2019sinkhorn} exploits the Sinkhorn divergence. Unlike these works, our proposed SWAEs directly minimize the OT cost of the joint distributions of the observed data and the latent representation with the inclusion of a reconstruction loss for further improving the generative model.



\section{Conclusion and Future Work}

We contributed a novel family of generative autoencoders, termed Symmetric Wasserstein Autoencoders (SWAEs) under the framework of OT. We proposed to symmetrically match the encoding and the decoding distributions with the inclusion of a reconstruction loss for further improving the generative model. We conducted empirical studies on  benchmark tasks to confirm the superior performance of SWAEs over  state-of-the-art generative autoencoders.

We believe that symmetrically aligning the encoding and the decoding distributions with a proper regularizer is crucial to improving the performance of generative models. To further enhance the performance of SWAEs, it is worthwhile to exploit other methods for the prior design,  \eg, the flow-based approaches \citep{rezende2015variational,yoshuabengio2014nice,dinh2016density}, and other forms of the reconstruction loss, \eg, the cross entropy.

\newpage
\clearpage
\bibliography{sun_146.bib}

\clearpage
\appendix

\section{Datasets and Network Architectures}\label{app:setup}
In this section, we briefly describe the datasets, the network architectures, and the hyperparameters that are used in our training algorithm. 
\begin{itemize}
    \item MNIST: The dataset includes $70,000$ binarized
    images of numeric digits from $0$ to $9$, each of the size $28\times 28$. There are $7,000$ images per class. The training set contains $50,000$ images, the validation set contains $10,000$ images for choosing the best model based on the loss function, and the test set contains $10,000$ images. 
    \item Fashion-MNIST: The dataset includes $70,000$ binarized images of fashion products in $10$ classes. This dataset has the same image size and the split of the training, validation, and test sets as in MNIST. 
    \item Coil20: The dataset includes gray-scale images of $20$ objects, each image of the size $32\times 32$. The training set contains $1040$ images, the validation set contains $200$ images for choosing the best model based on the loss function, and the test set contains $200$ images.
    \item CIFAR10-sub: The CIFAR-10 dataset consists of $60,000$ $32\times 32$ colour images in $10$ classes with $6,000$ images per class. 
There are $40,000$ training, $10,000$ validation, and $10,000$ test images. We randomly select three classes to form the CIFAR10-sub dataset, namely {\em bird, cat}, and {\em ship}.
\end{itemize}

Network architecture of SWAE: The building block of the network structure of SWAE is based on VampPrior, called GatedConv2d. GatedConv2d contains two convolutional layers with the gating mechanism utilized as an element-wise non-linearity. The parameters in the function GatedConv2d() represent the number of the input channels, the number of the output channels, kernel size, stride, and padding, respectively.   The conditional prior network outputs the mean and the log-variance of a Gaussian distribution, based on which the latent prior is sampled.

\begin{itemize}
    \item The structure of the encoder network: GatedConv2d(1,32,7,1,3)-GatedConv2d(32,32,3,2,1)-GatedConv2d(32,64,5,1,2)-GatedConv2d(64,64,3,2,1)-GatedConv2d(64,6,3,1,1), followed by one fully-connected layer with no activation function. 
    \item The structure of the conditional prior network: The layers of GatedConv2d are the same as those in the encoder network, which are followed by two fully-connected layers. One produces the mean, and the other produces the log-variance with the activation function Hardtanh.
    \item The structure of the decoder network: Two fully-connected layers with the gating mechanism, followed by GatedConv2d(1,64,3,1,1)-GatedConv2d(64,64,3,1,1)-GatedConv2d(64,64,3,1,1)-GatedConv2d(64,64,3,1,1), followed by a convolutional layer with the activation function Sigmoid.
\end{itemize}

\section{More Experimental Results}\label{app:exp}
In this section, we show more experimental results based on the comparison with the benchmarks. 

\begin{figure*}[h]
	\centering
	\subfigure[SWAE ($\beta^* = 0.02$)]
	{\includegraphics[width=0.3\textwidth]{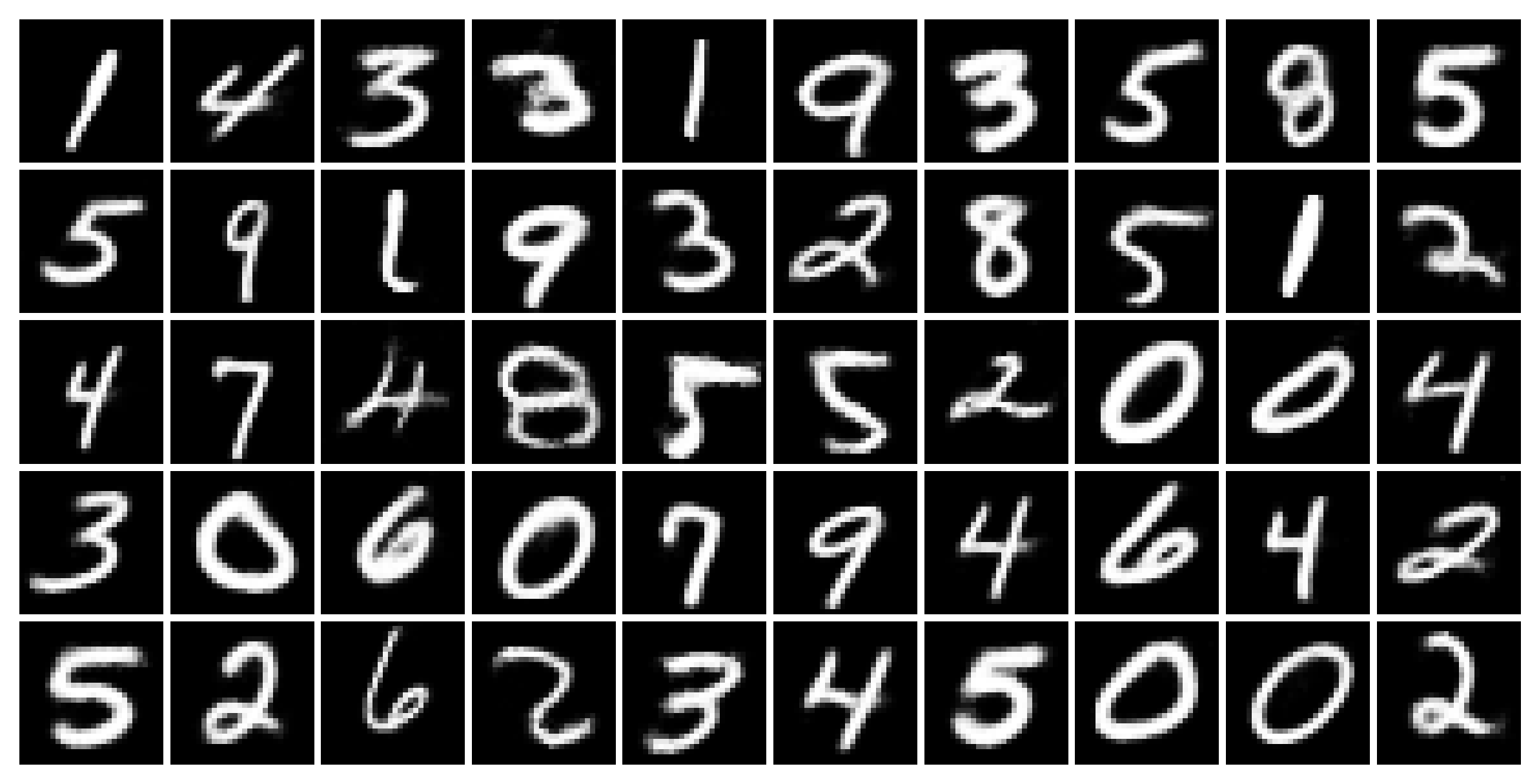}}
	\subfigure[SWAE ($\beta = 1$)]
	{\includegraphics[width=0.3\textwidth]{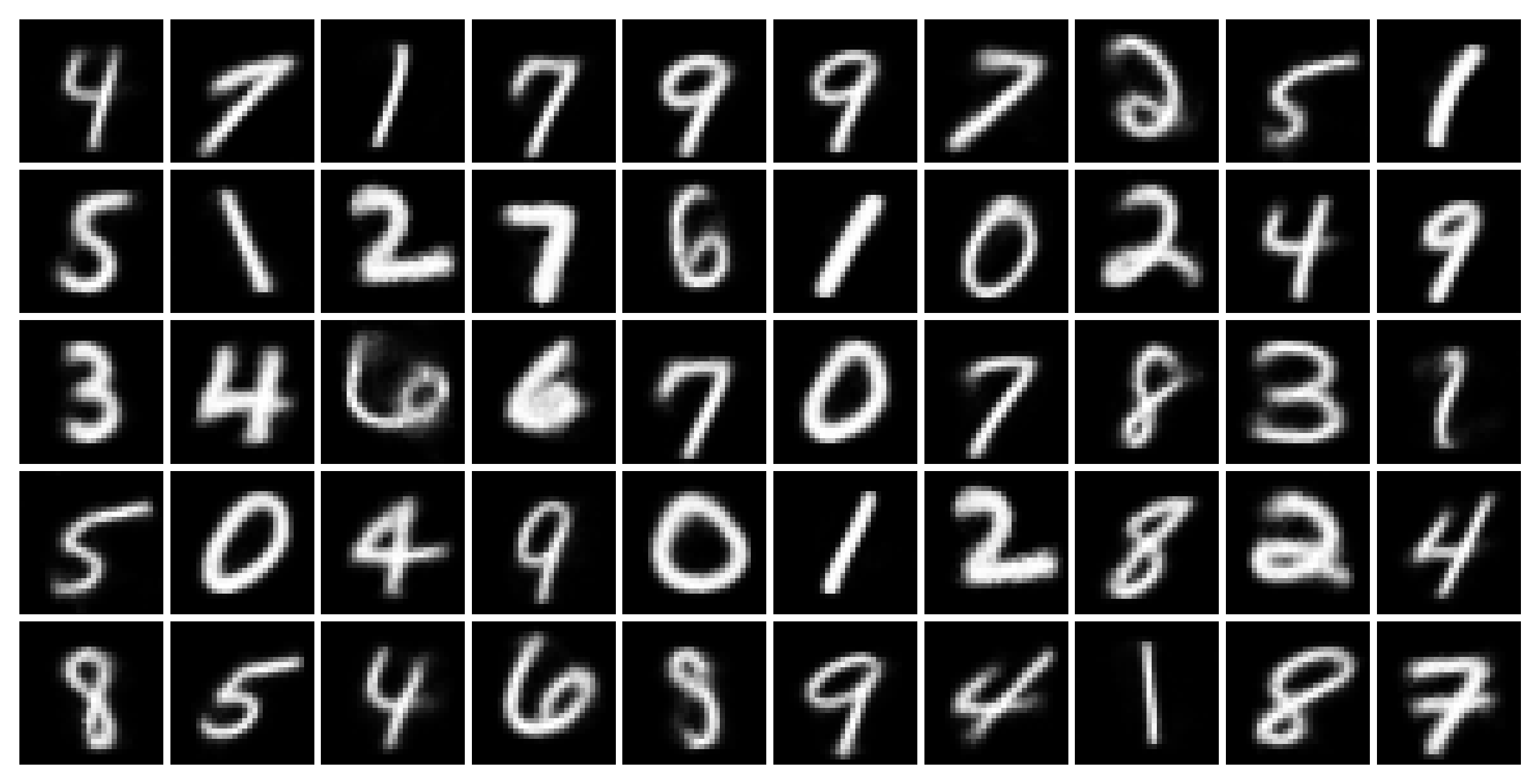}}	
	\subfigure[SWAE ($\beta = 0$)]
	{\includegraphics[width=0.3\textwidth]{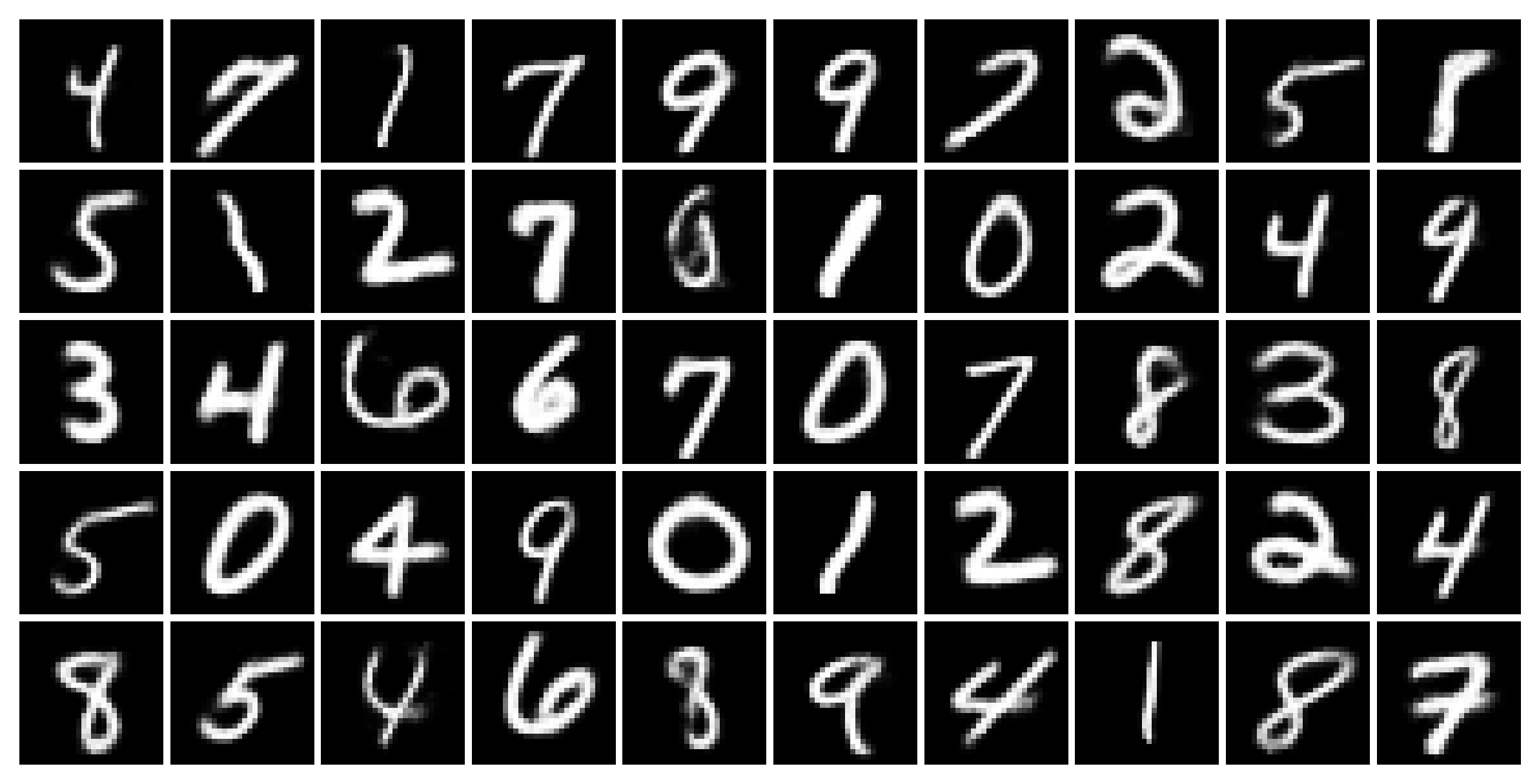}}	
		\vspace{-1mm}
	\subfigure[VAE]
	{\includegraphics[width=0.3\textwidth]{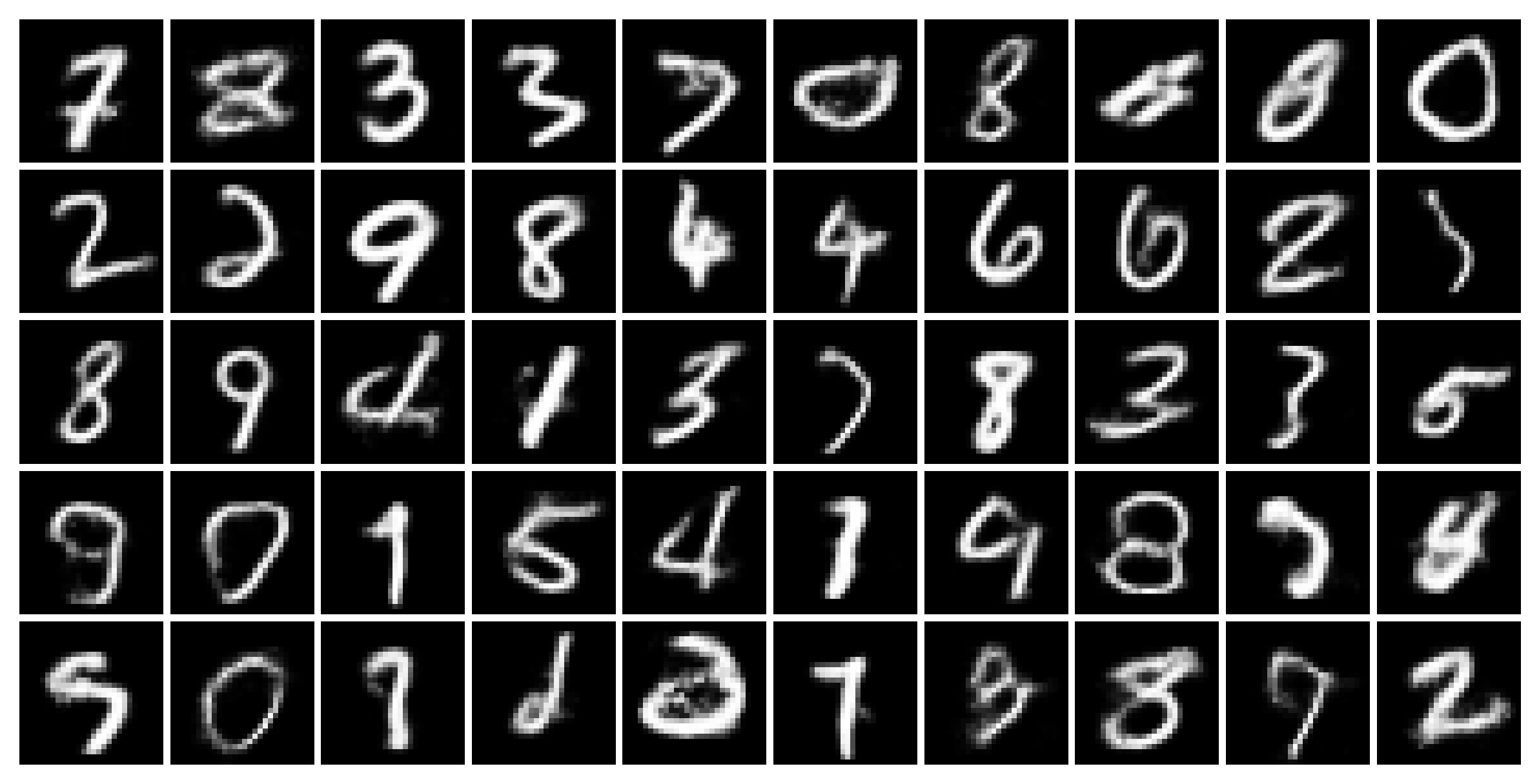}}
		\vspace{-1mm}
	\subfigure[WAE-GAN]
	{\includegraphics[width=0.3\textwidth]{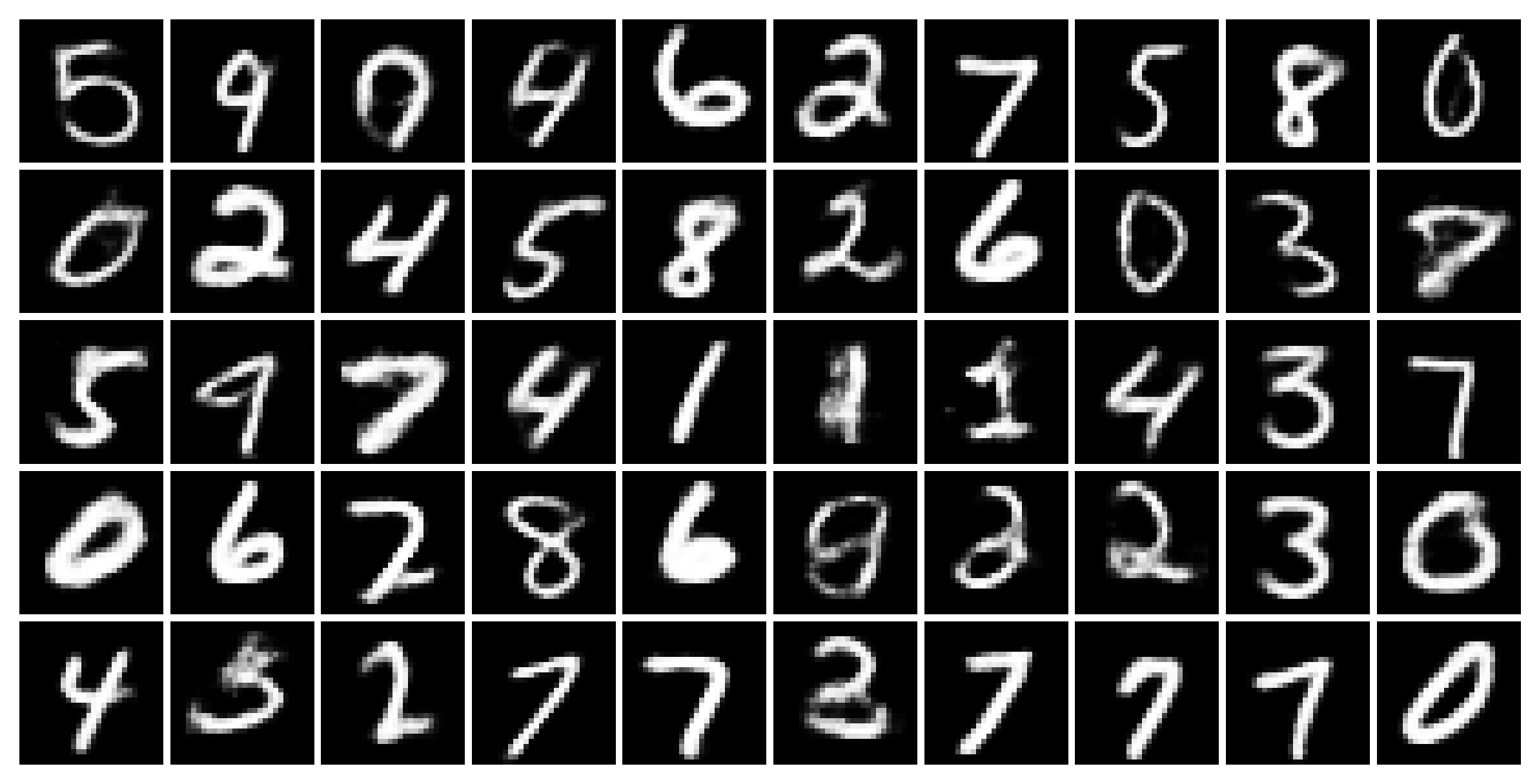}}	
	\subfigure[WAE-MMD]
	{\includegraphics[width=0.3\textwidth]{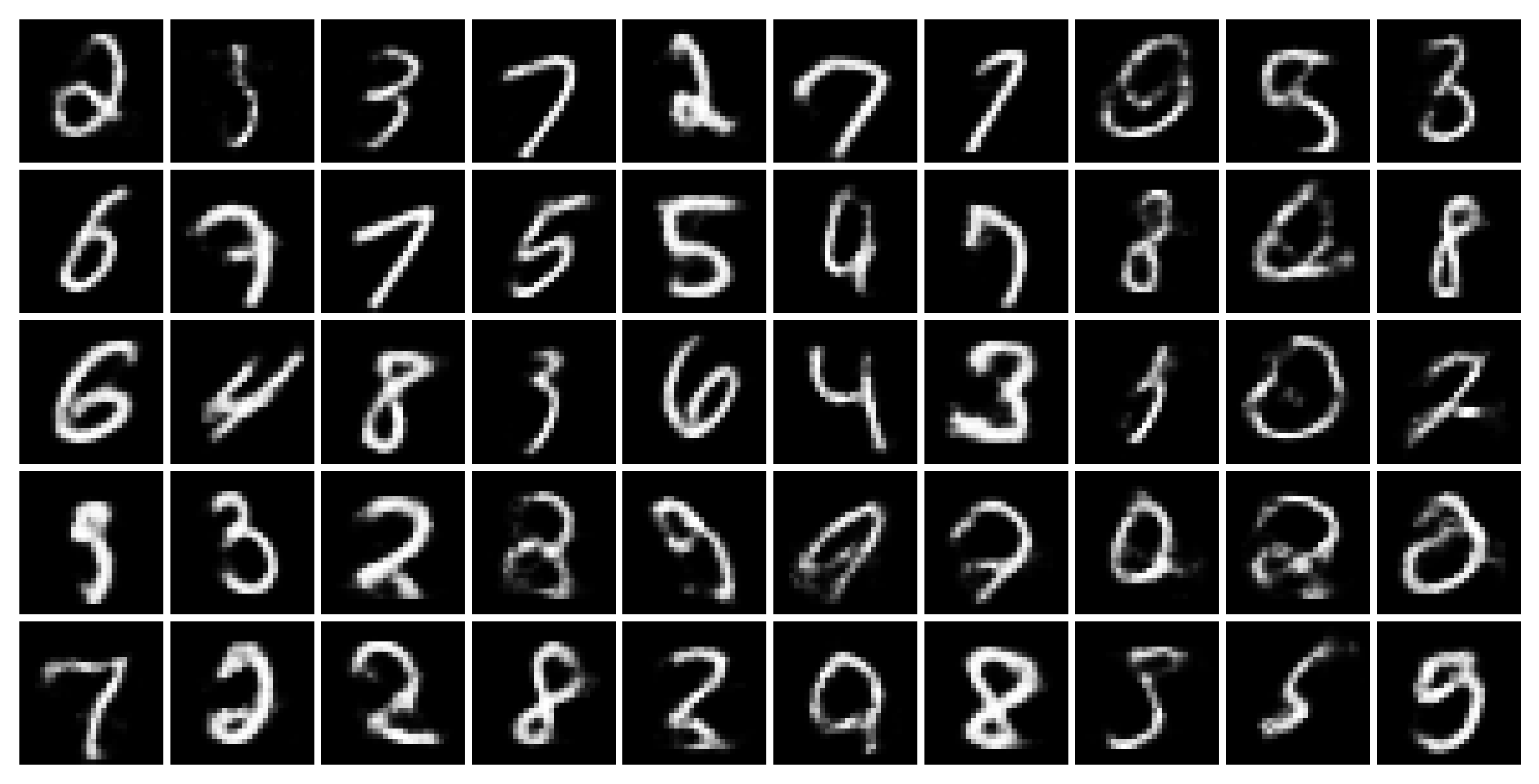}}	
	\subfigure[VampPrior]
	{\includegraphics[width=0.3\textwidth]{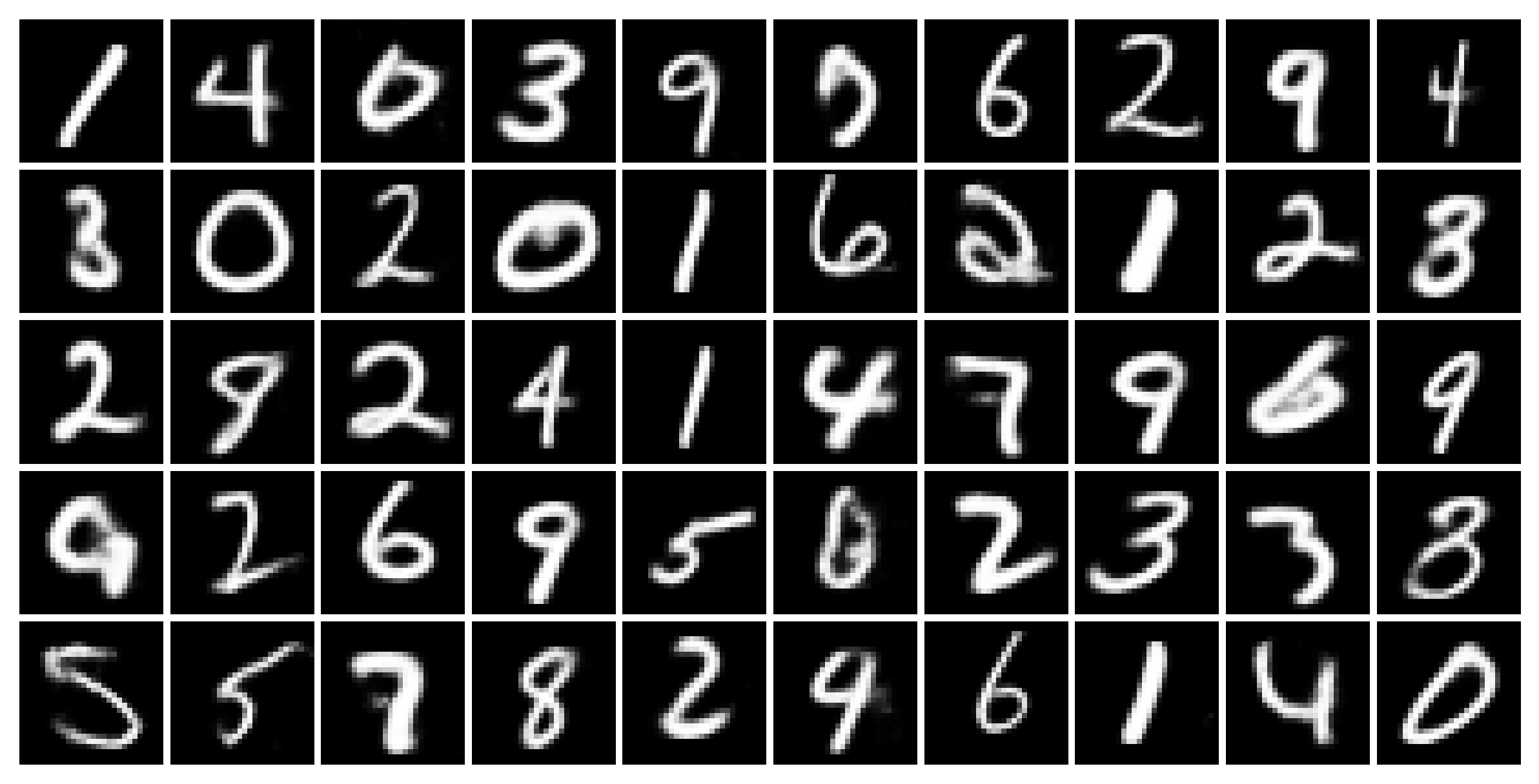}}
	\subfigure[MIM]
	{\includegraphics[width=0.3\textwidth]{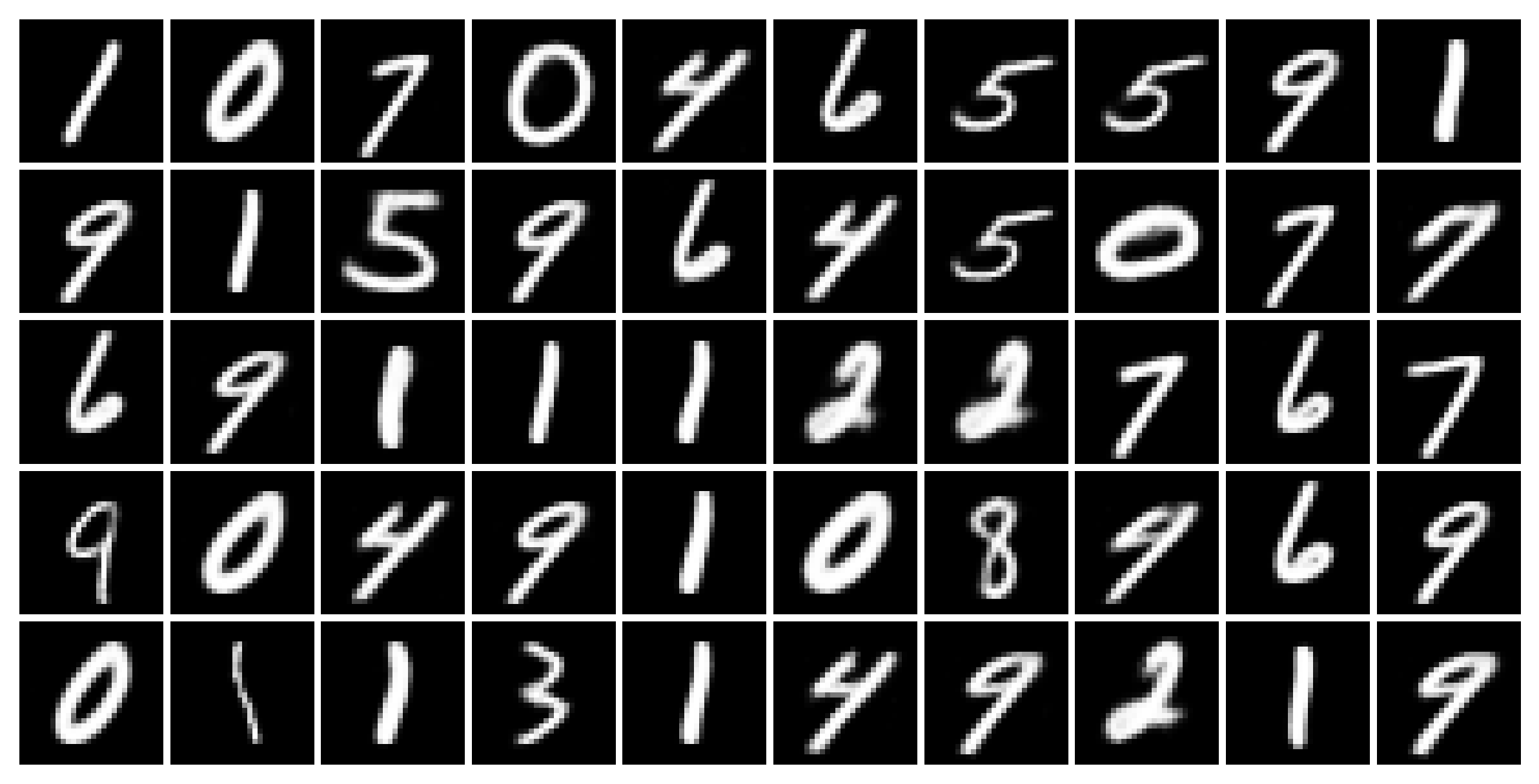}	}
	\vspace{-0.2cm}
	\caption{Generated new samples on MNIST. dim-$\bz = 8$ for all methods.
	\label{fig:mnist_gen}}

\end{figure*}

\vspace{-2mm}

\begin{figure*}[h]
	\centering
	\subfigure[SWAE ($\beta^* = 0.05$)]
	{\includegraphics[width=0.3\textwidth]{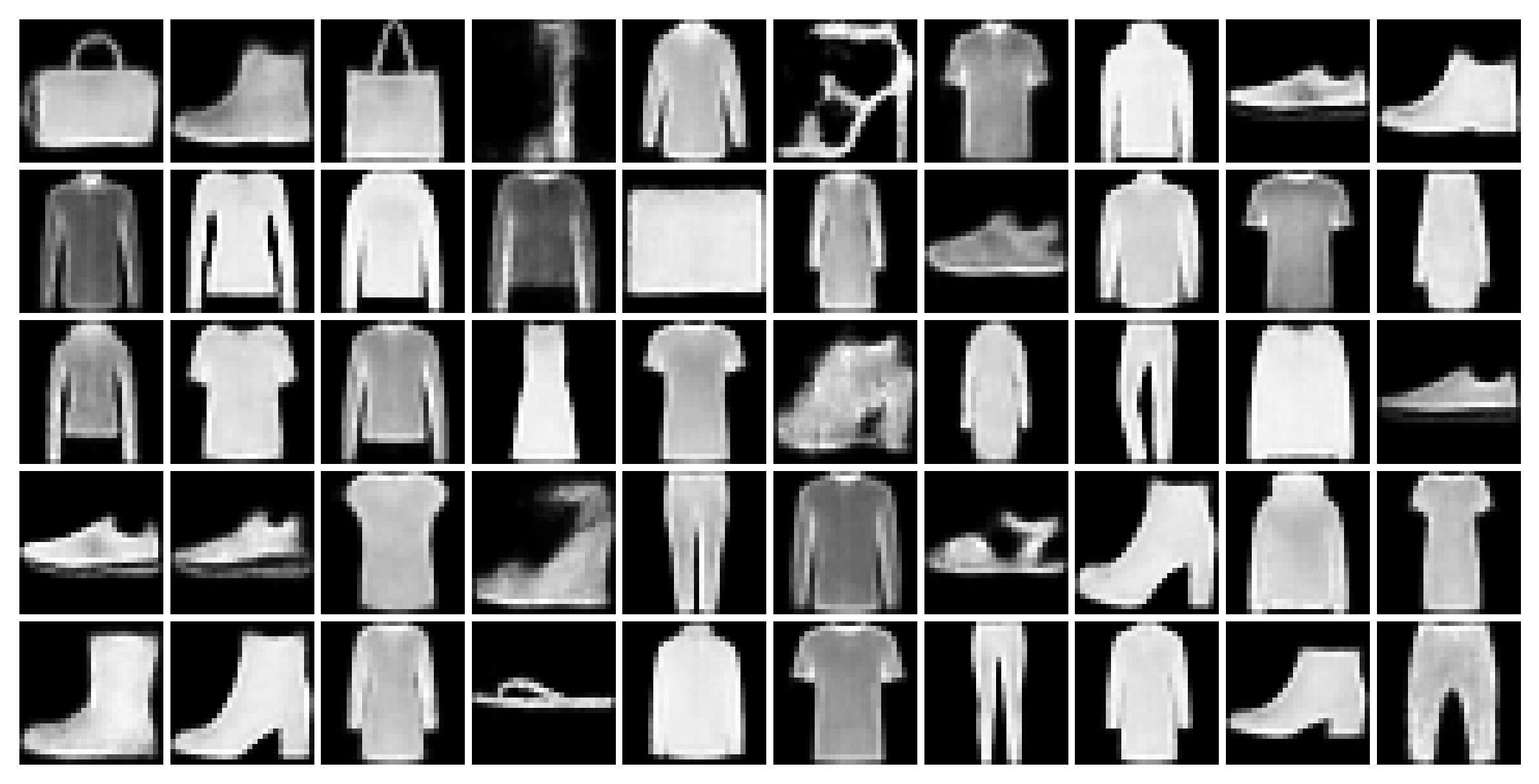}}	
	\subfigure[SWAE ($\beta = 1$)]
	{\includegraphics[width=0.3\textwidth]{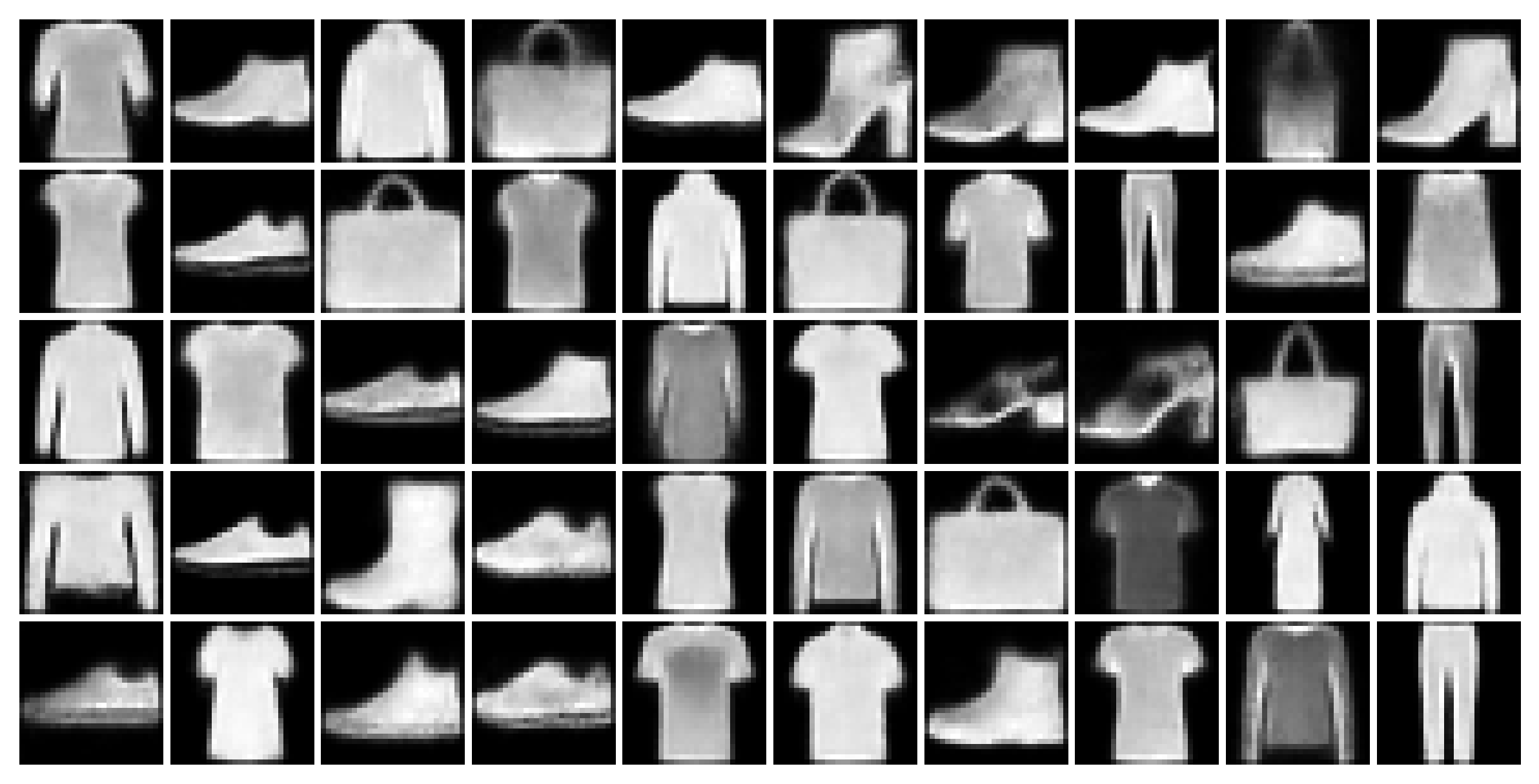}}	
	\subfigure[SWAE ($\beta = 0$)]
	{\includegraphics[width=0.3\textwidth]{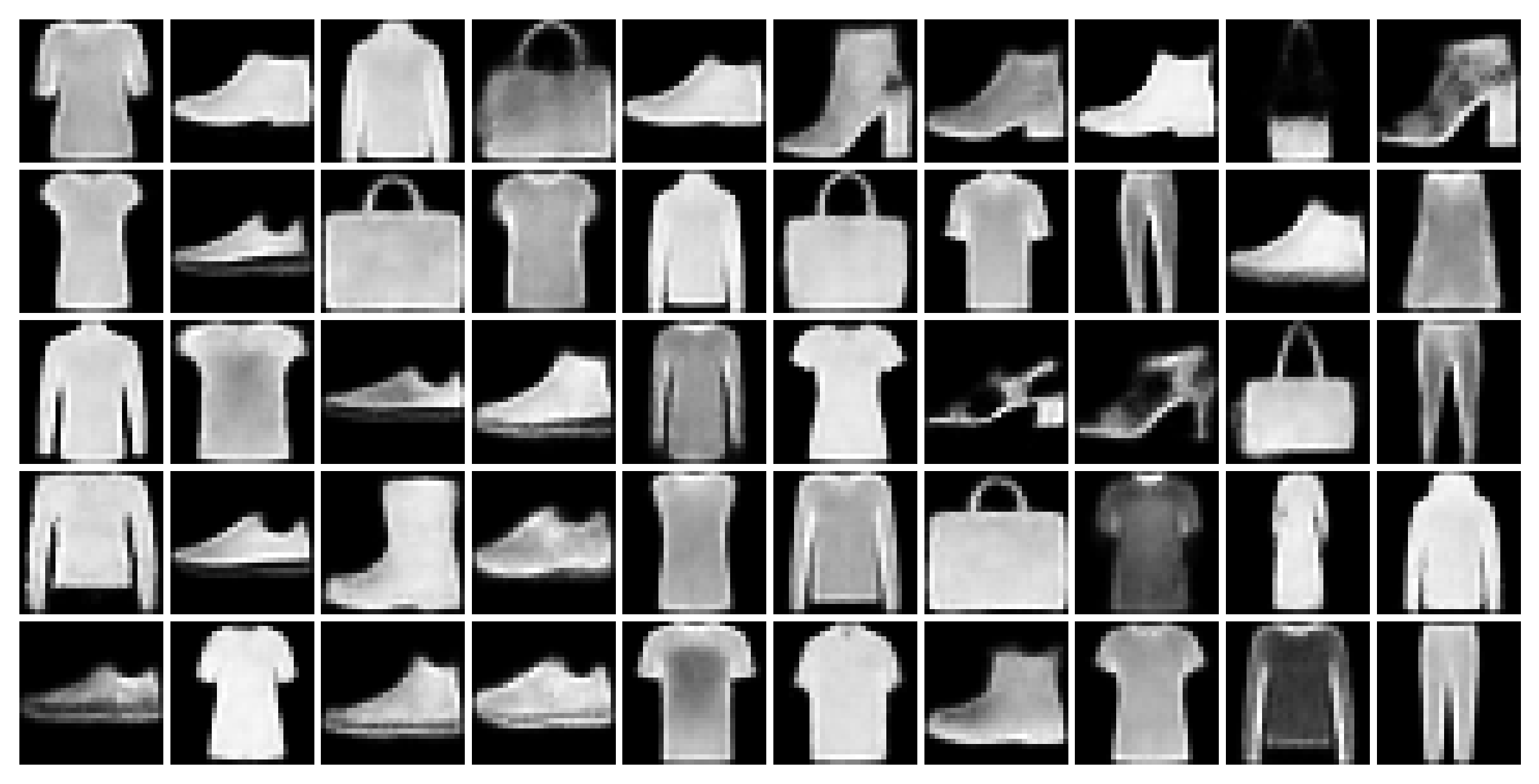}}	
		\vspace{-1mm}
	\subfigure[VAE]
	{\includegraphics[width=0.3\textwidth]{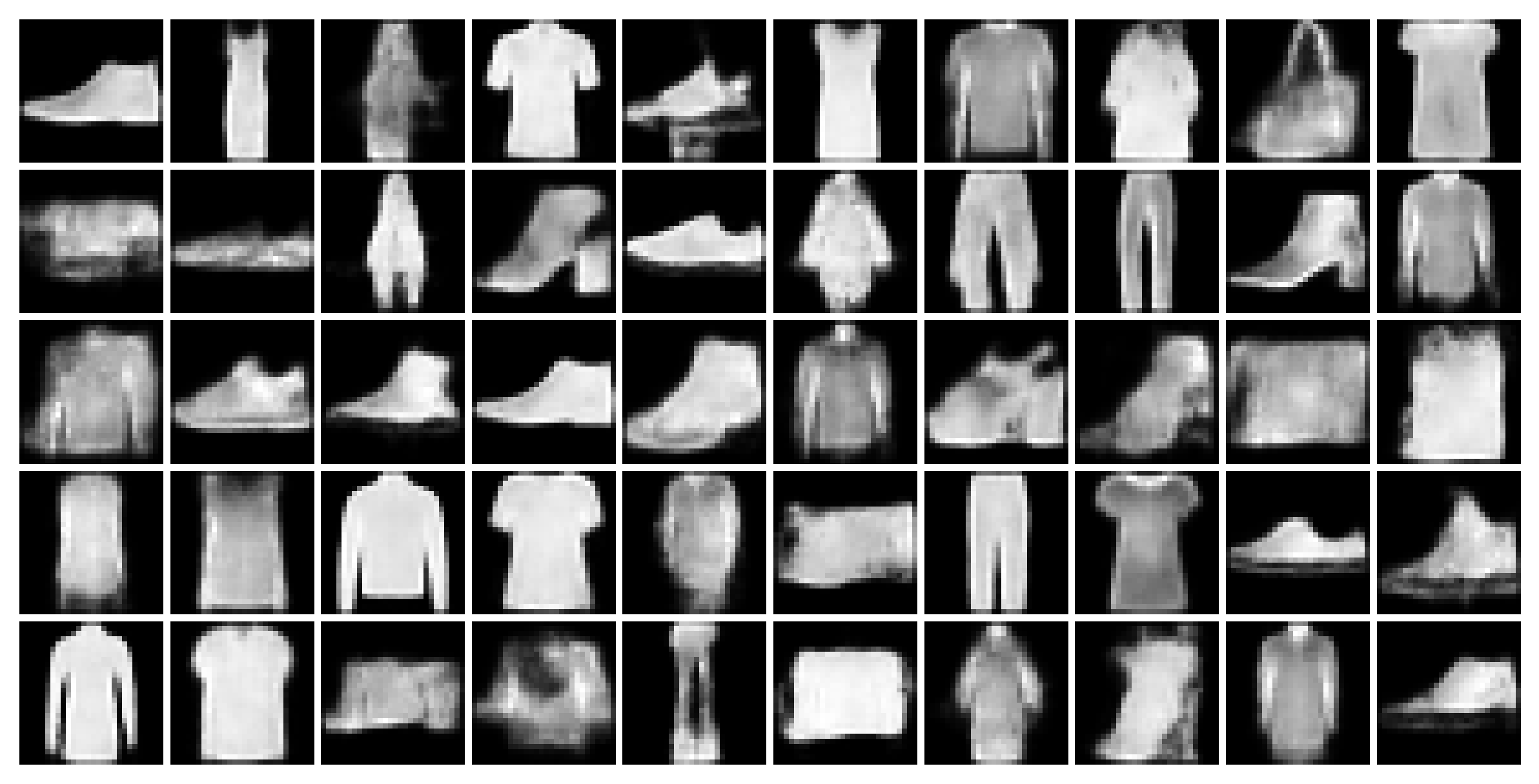}}
		\vspace{-1mm}
	\subfigure[WAE-GAN]
	{\includegraphics[width=0.3\textwidth]{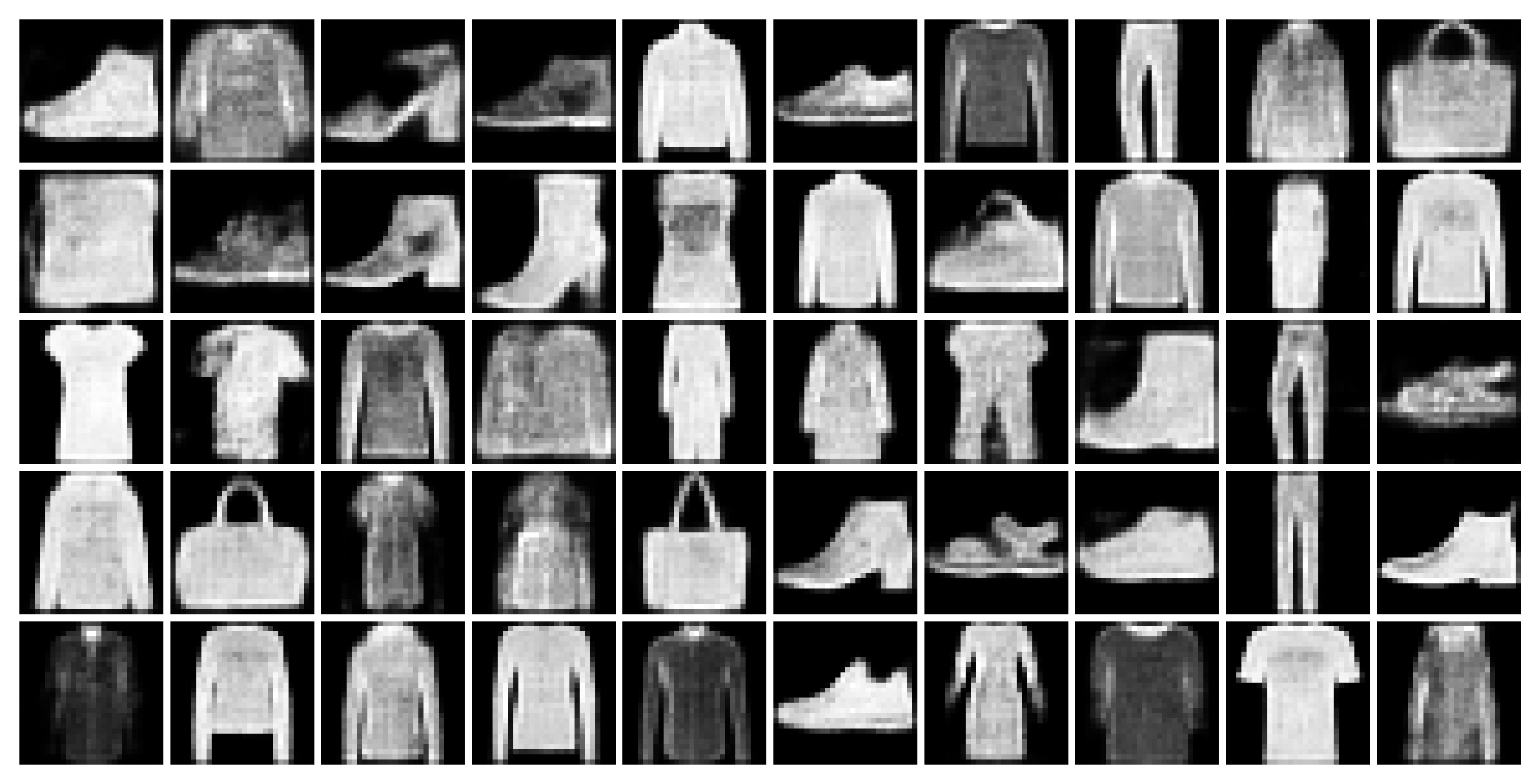}}	
	\subfigure[WAE-MMD]
	{\includegraphics[width=0.3\textwidth]{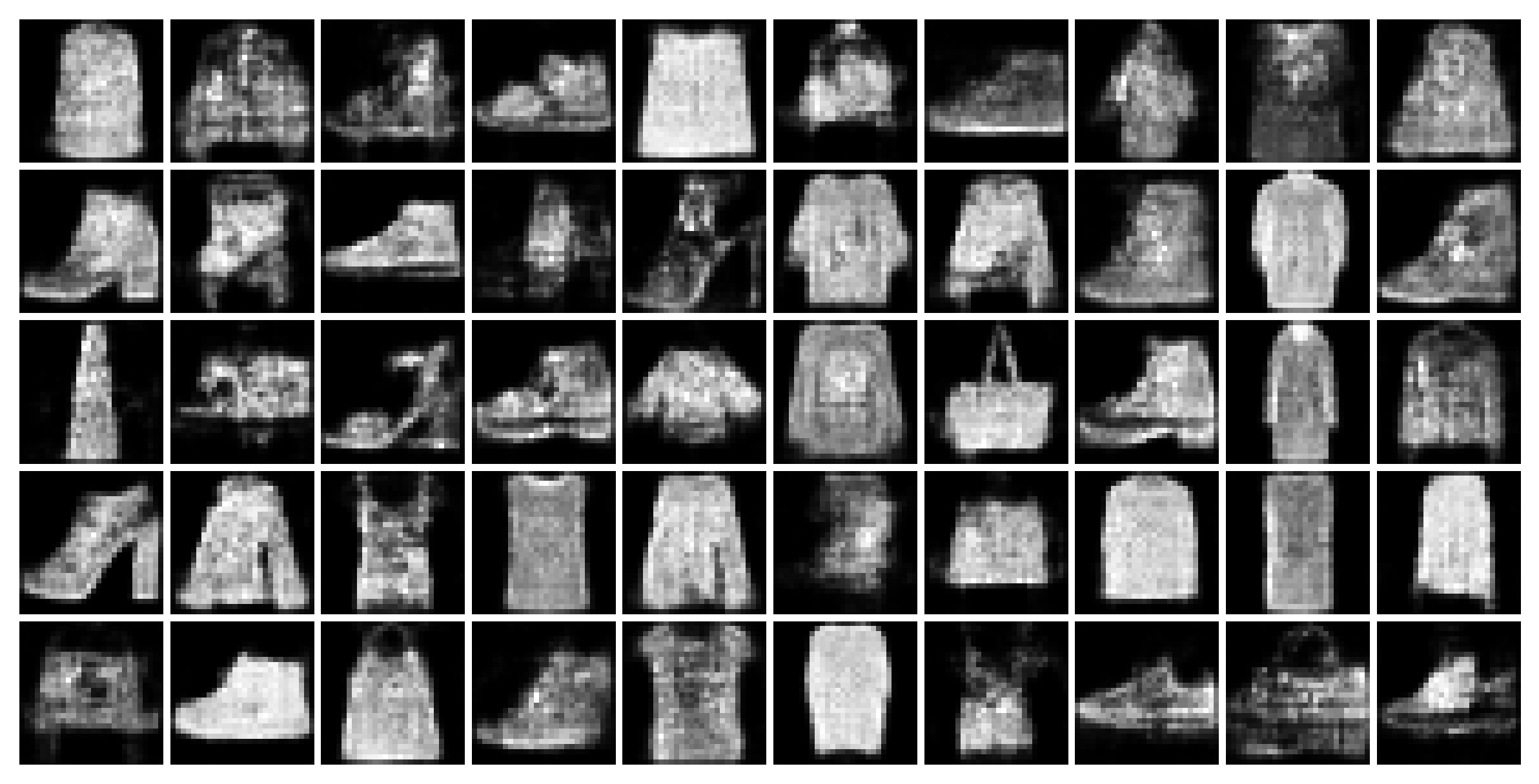}}	
	\subfigure[VampPrior]
	{\includegraphics[width=0.3\textwidth]{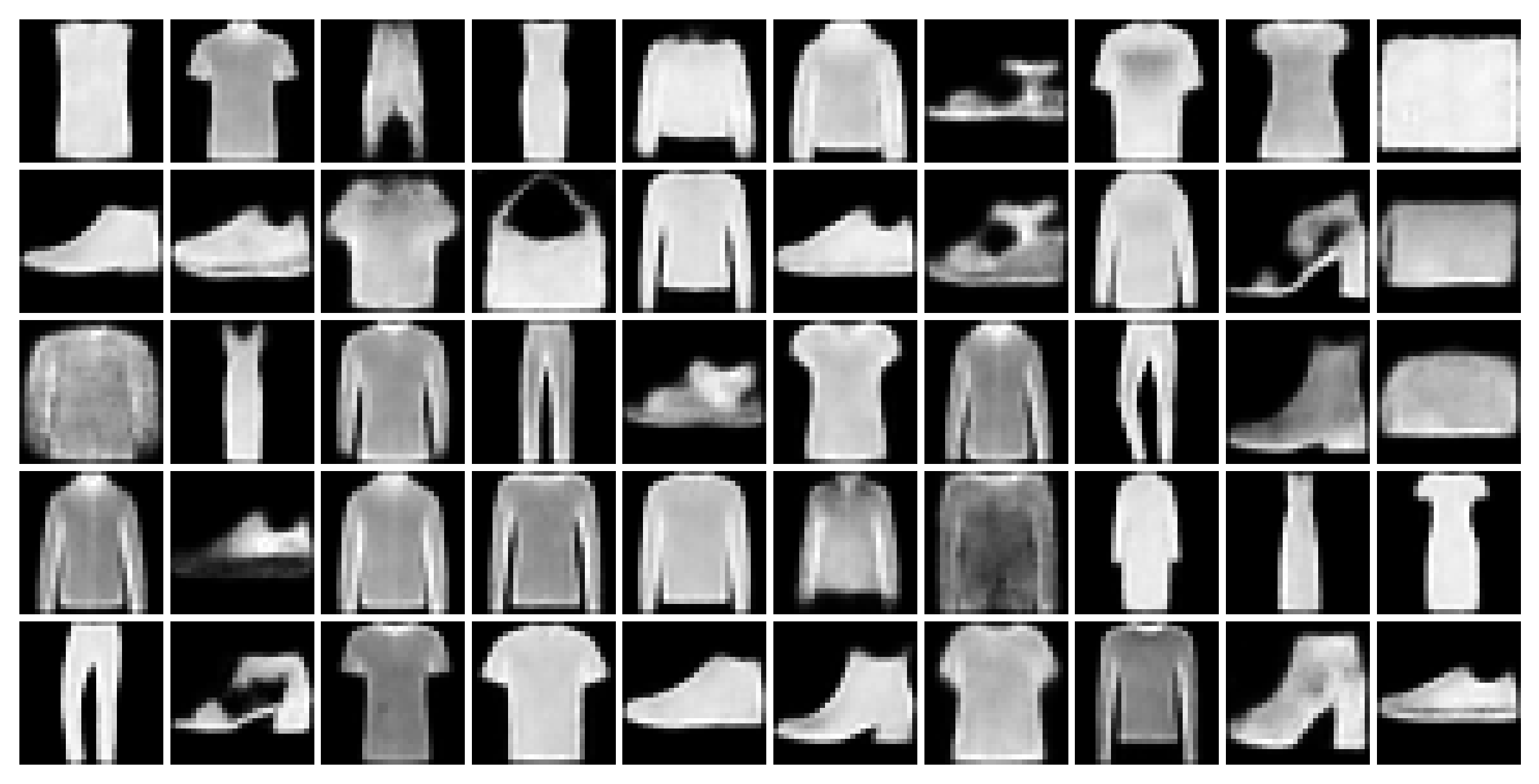}}
	\subfigure[MIM]
	{\includegraphics[width=0.3\textwidth]{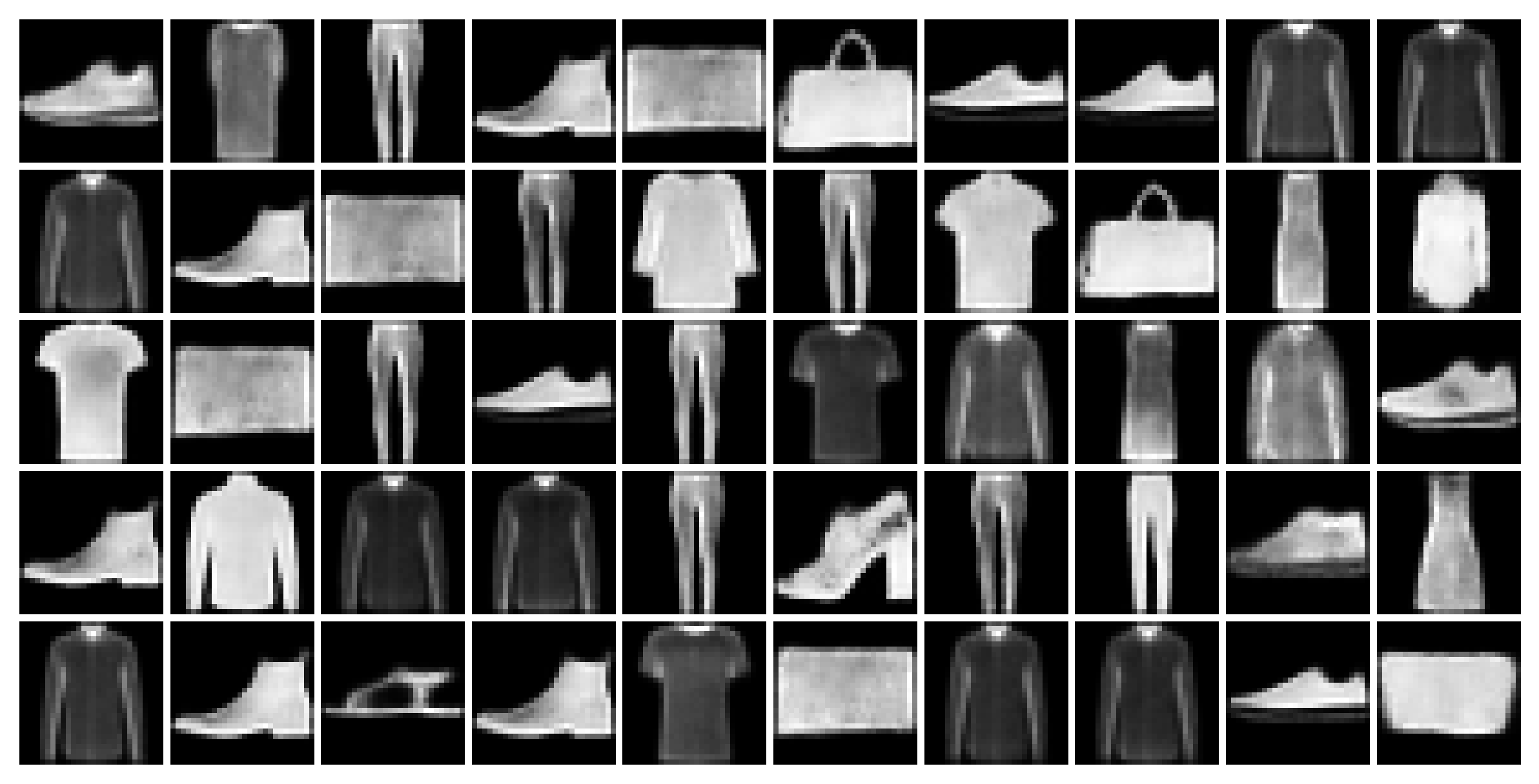}}
	\vspace{-0.2cm}
	\caption{Generated new samples on Fashion-MNIST. dim-$\bz = 8$ for all methods.}
	\label{fig:fash_gen}
\end{figure*}

\vspace{-2mm}

\begin{figure*}[h]
	\centering
	\subfigure[SWAE ($\beta^* = 0.5$)]
	{\includegraphics[width=0.3\textwidth]{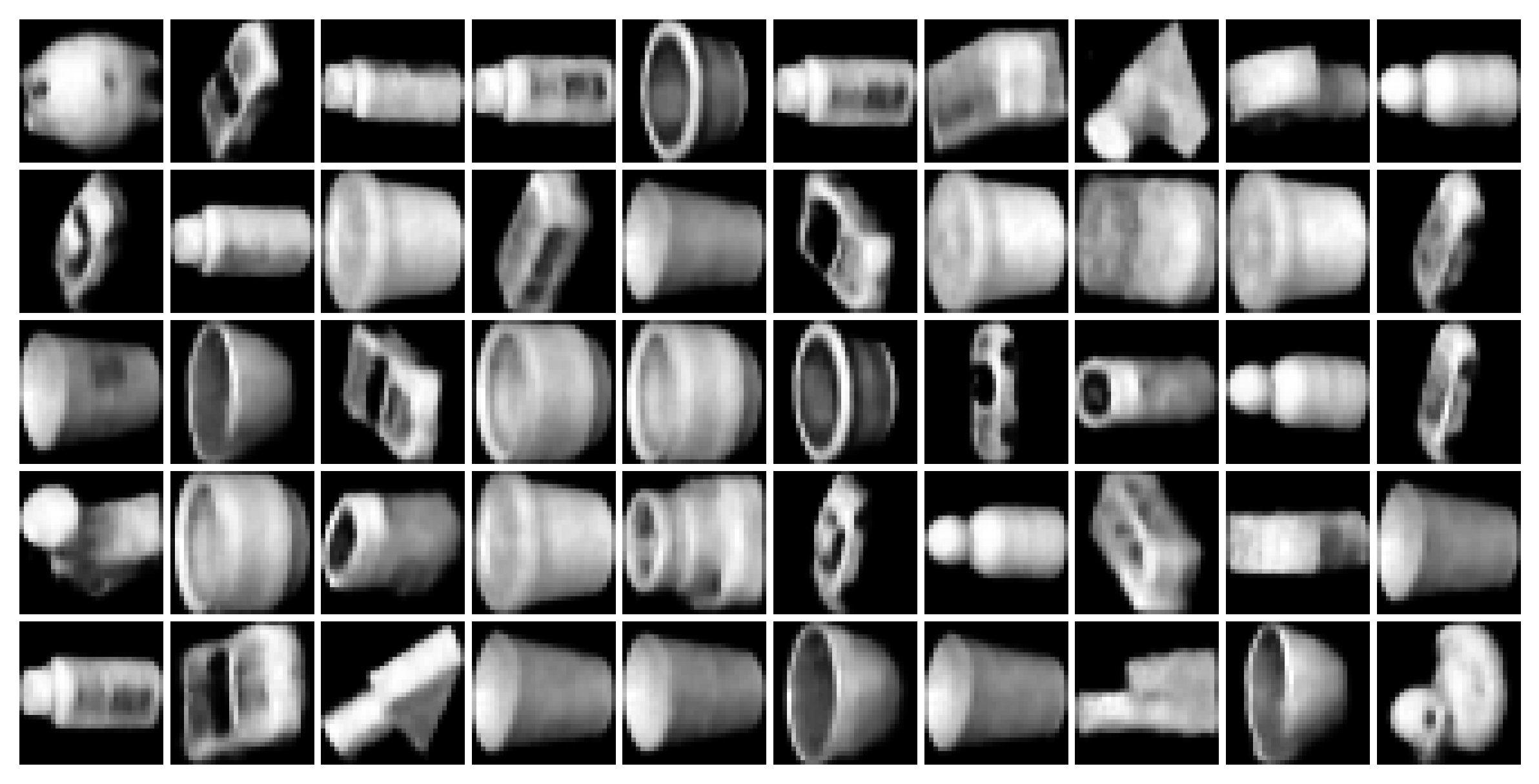}}	
		\subfigure[SWAE ($\beta = 1$)]
	{\includegraphics[width=0.3\textwidth]{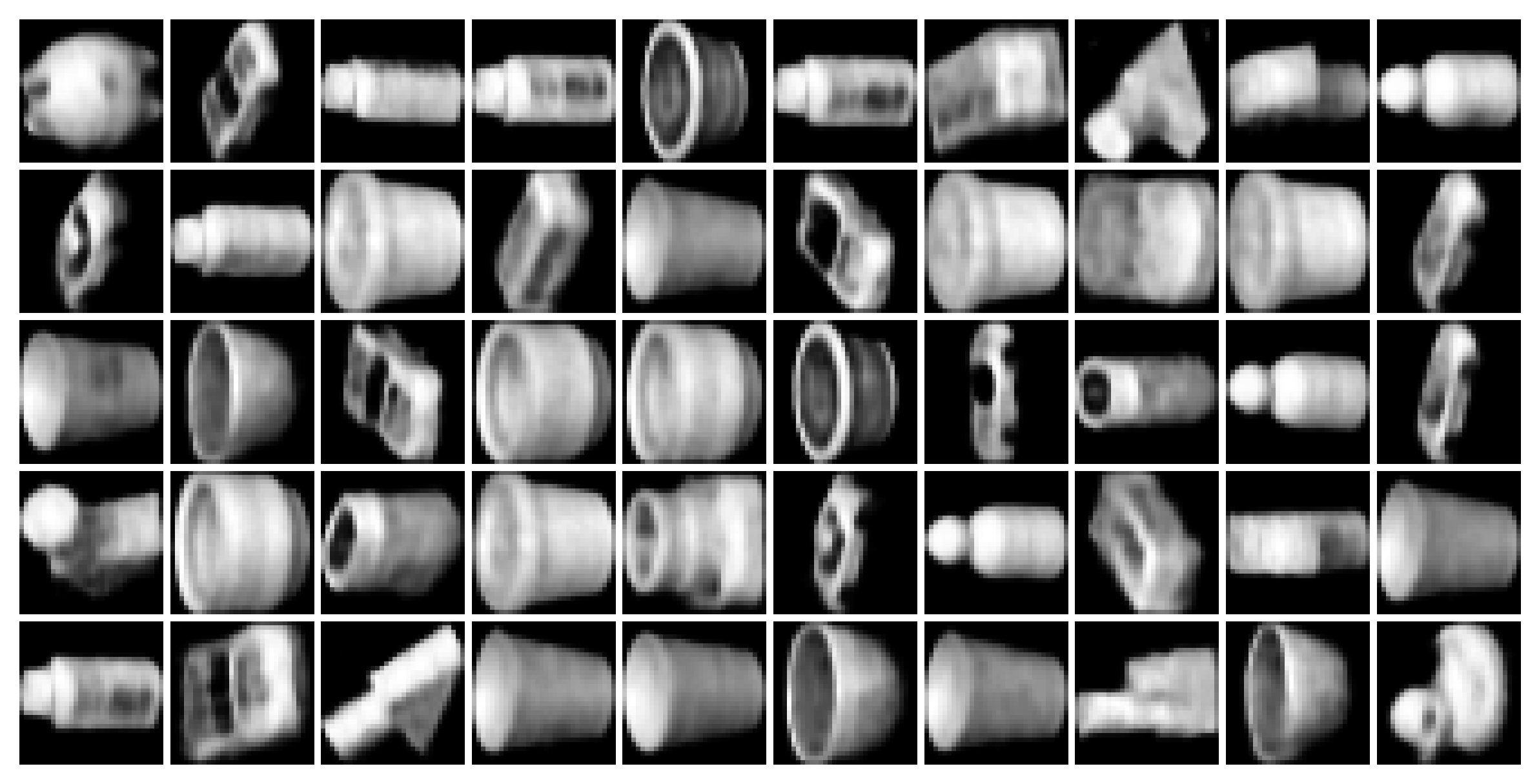}}	
	\subfigure[SWAE ($\beta = 0$)]
	{\includegraphics[width=0.3\textwidth]{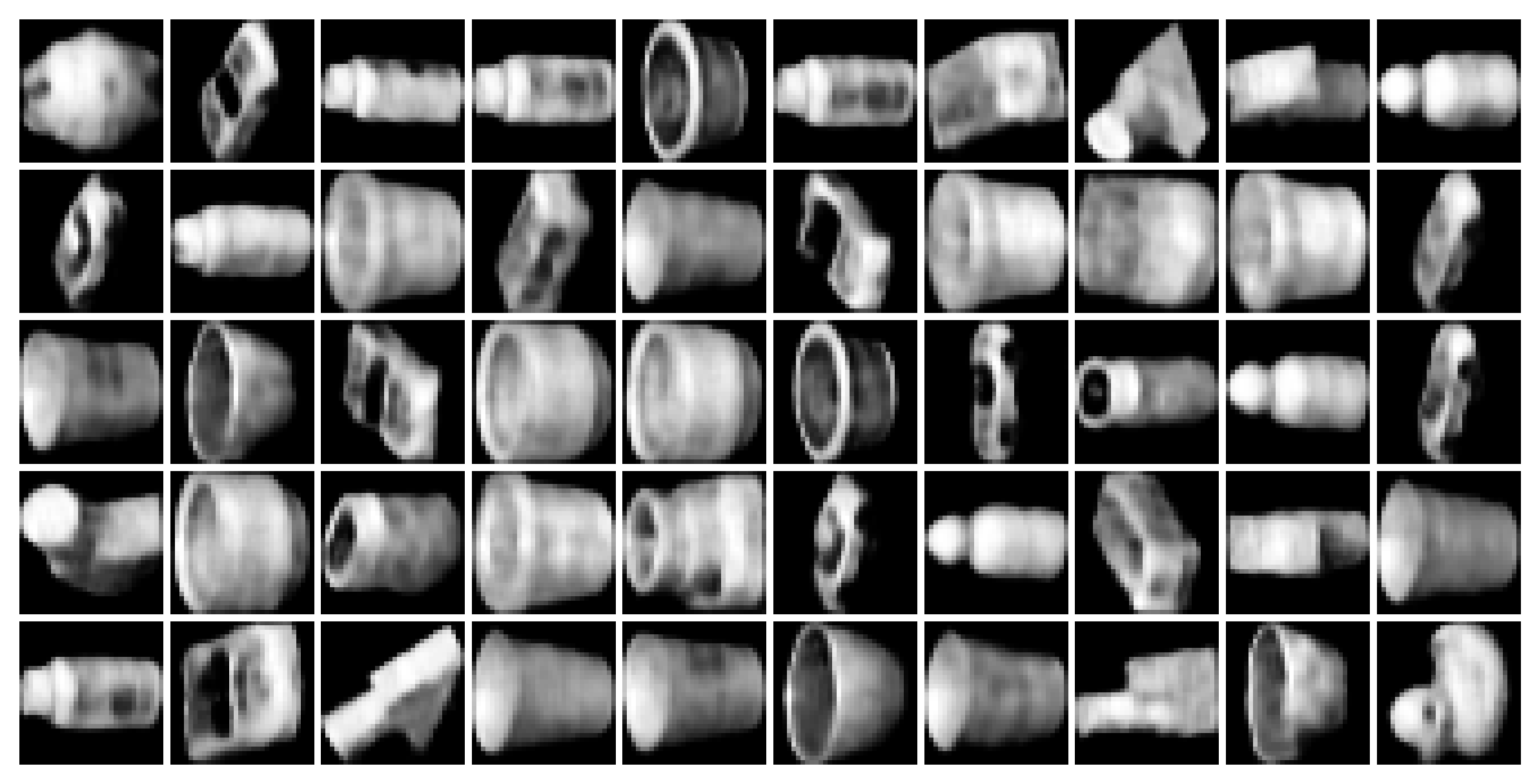}}	
		\vspace{-1mm}
	\subfigure[VAE]
	{\includegraphics[width=0.3\textwidth]{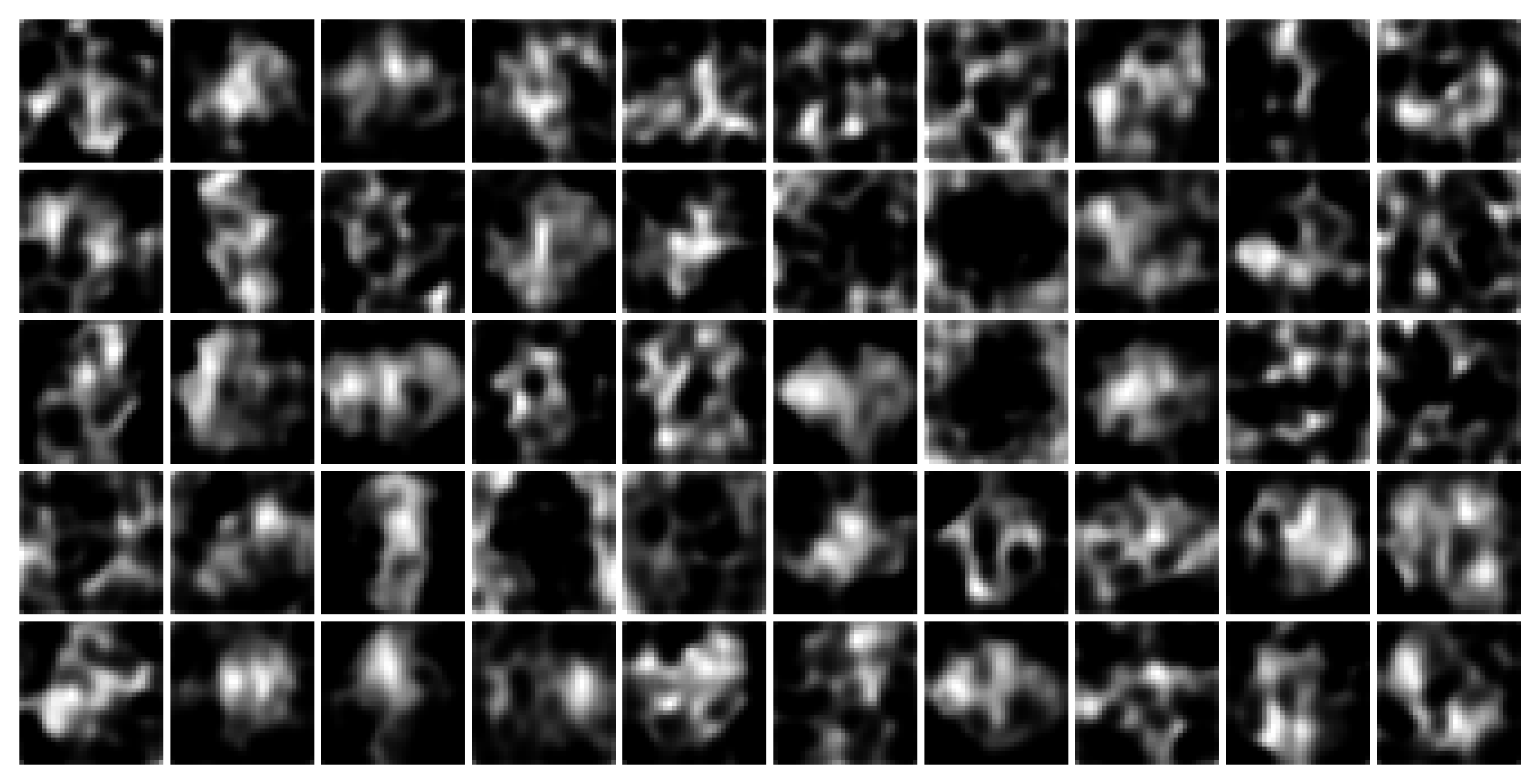}}	
		\vspace{-1mm}
	\subfigure[WAE-GAN]
	{\includegraphics[width=0.3\textwidth]{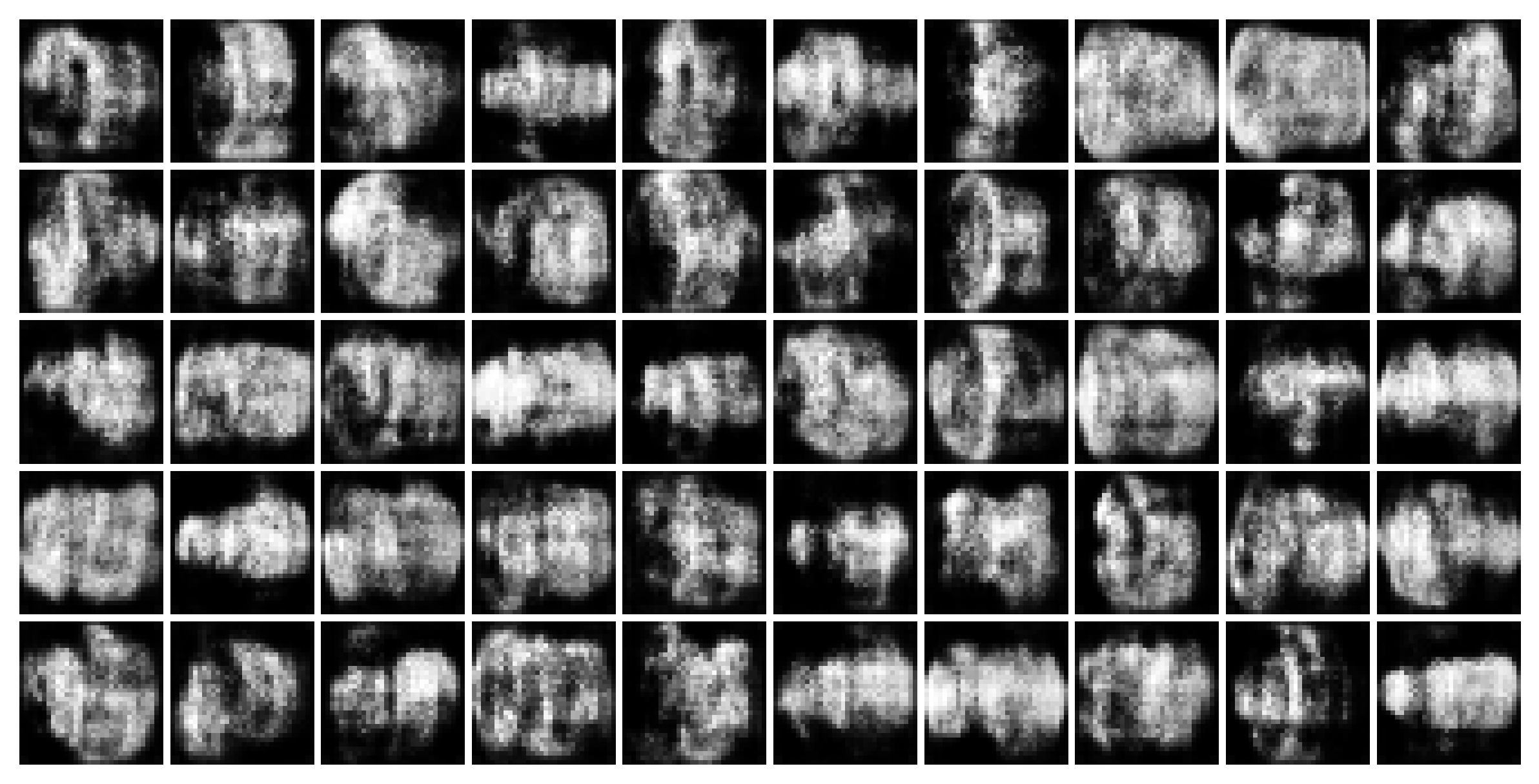}}	
	\subfigure[WAE-MMD]
	{\includegraphics[width=0.3\textwidth]{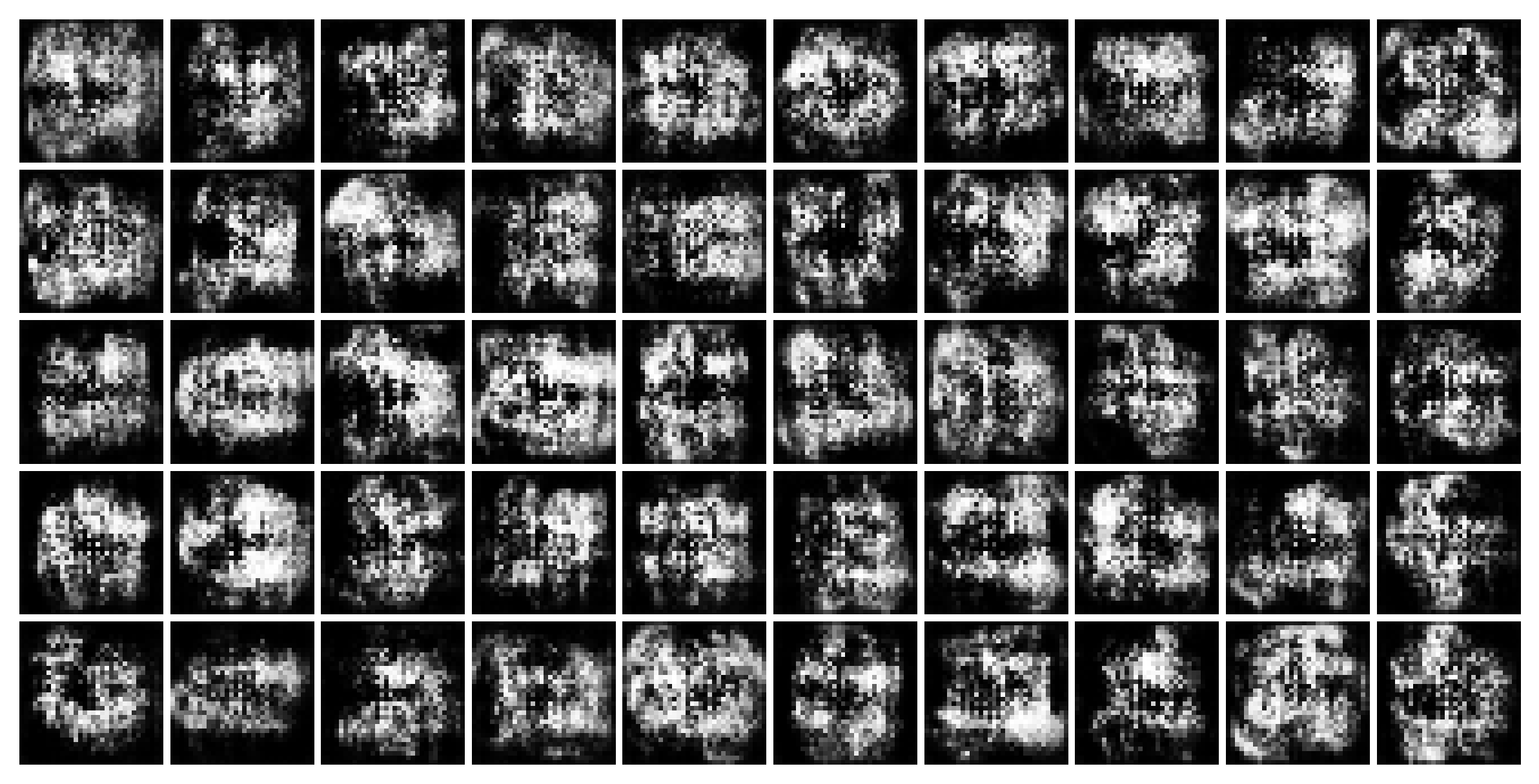}}	
	\subfigure[VampPrior]
	{\includegraphics[width=0.3\textwidth]{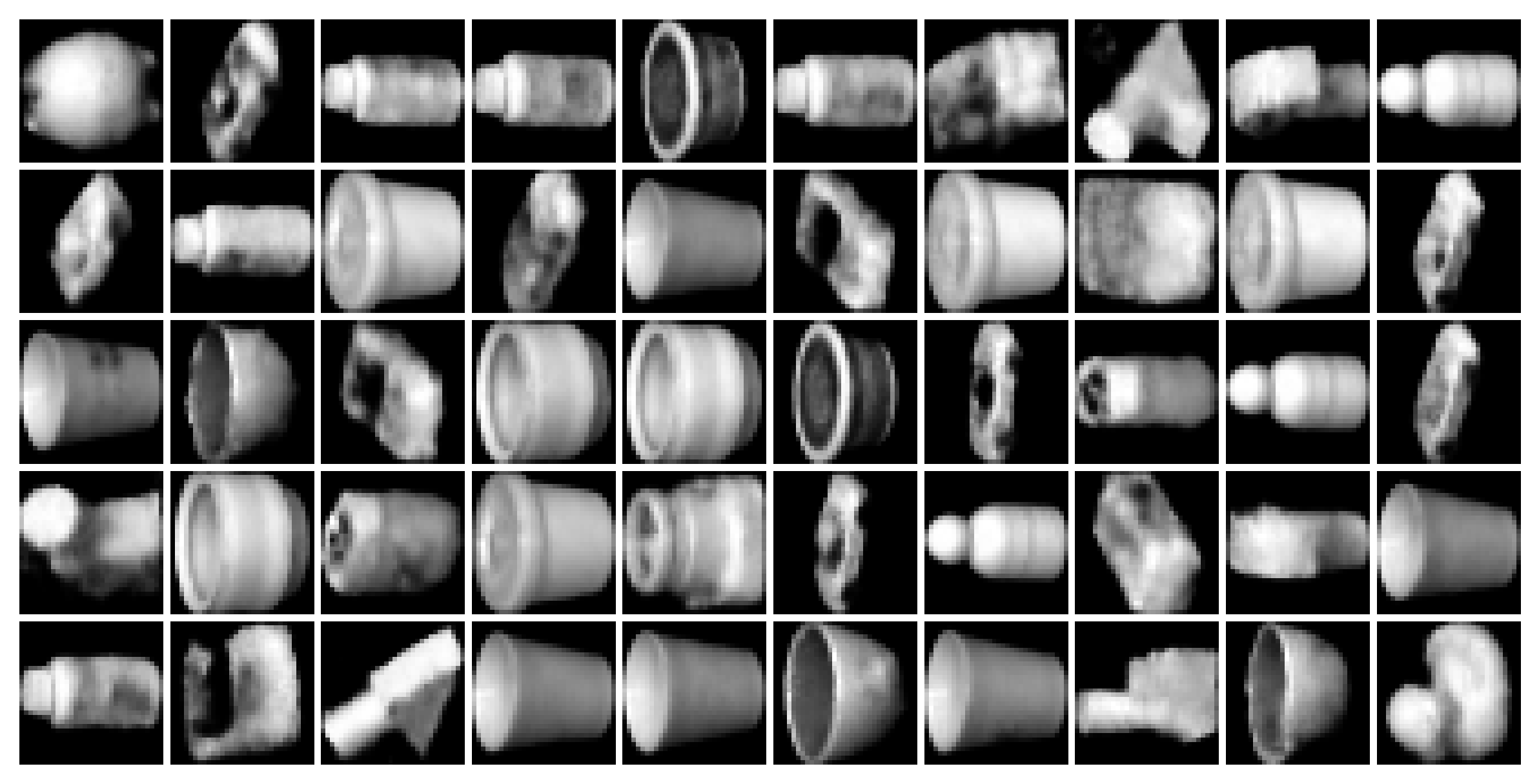}}
	\subfigure[MIM]
	{\includegraphics[width=0.3\textwidth]{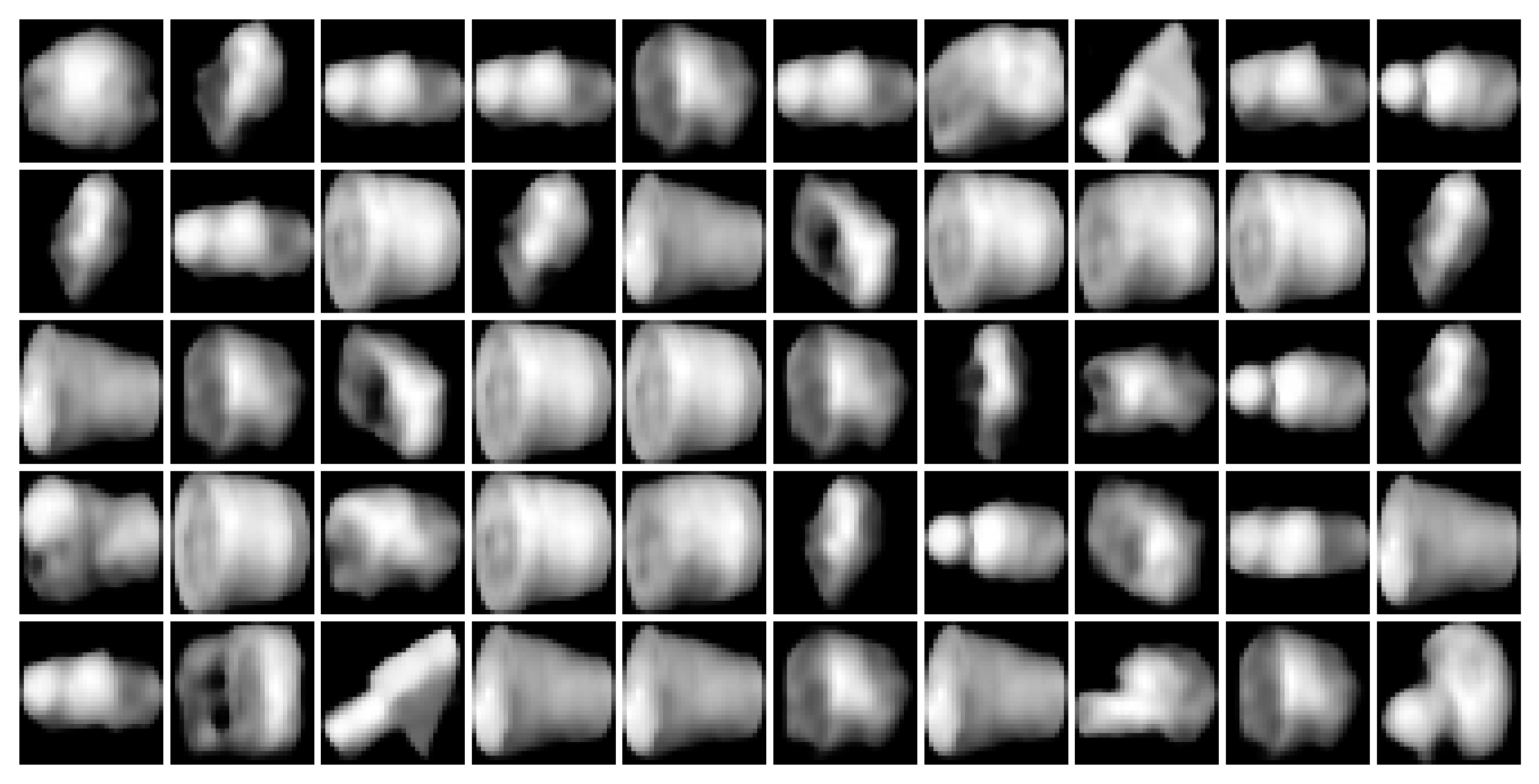}	}
	\vspace{-0.2cm}
	\caption{Generated new samples on Coil20. dim-$\bz = 80$ for all methods.
	\label{fig:coil20_gen}}
\end{figure*}

\begin{figure*}[h]
	\centering
		\subfigure[Real images]
	{\includegraphics[width=0.24\textwidth]{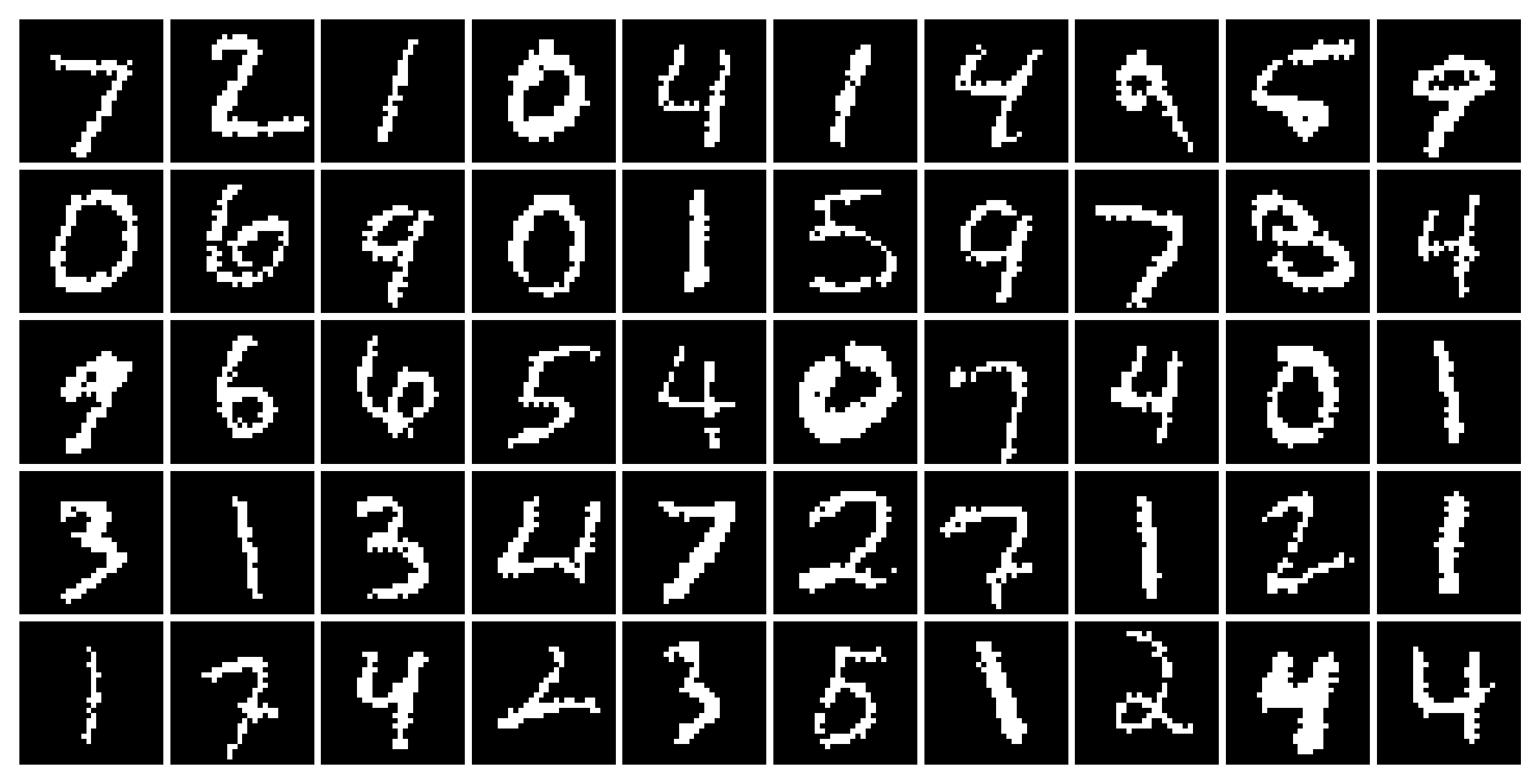}}	
	\subfigure[SWAE ($\beta = 1$)]
	{\includegraphics[width=0.24\textwidth]{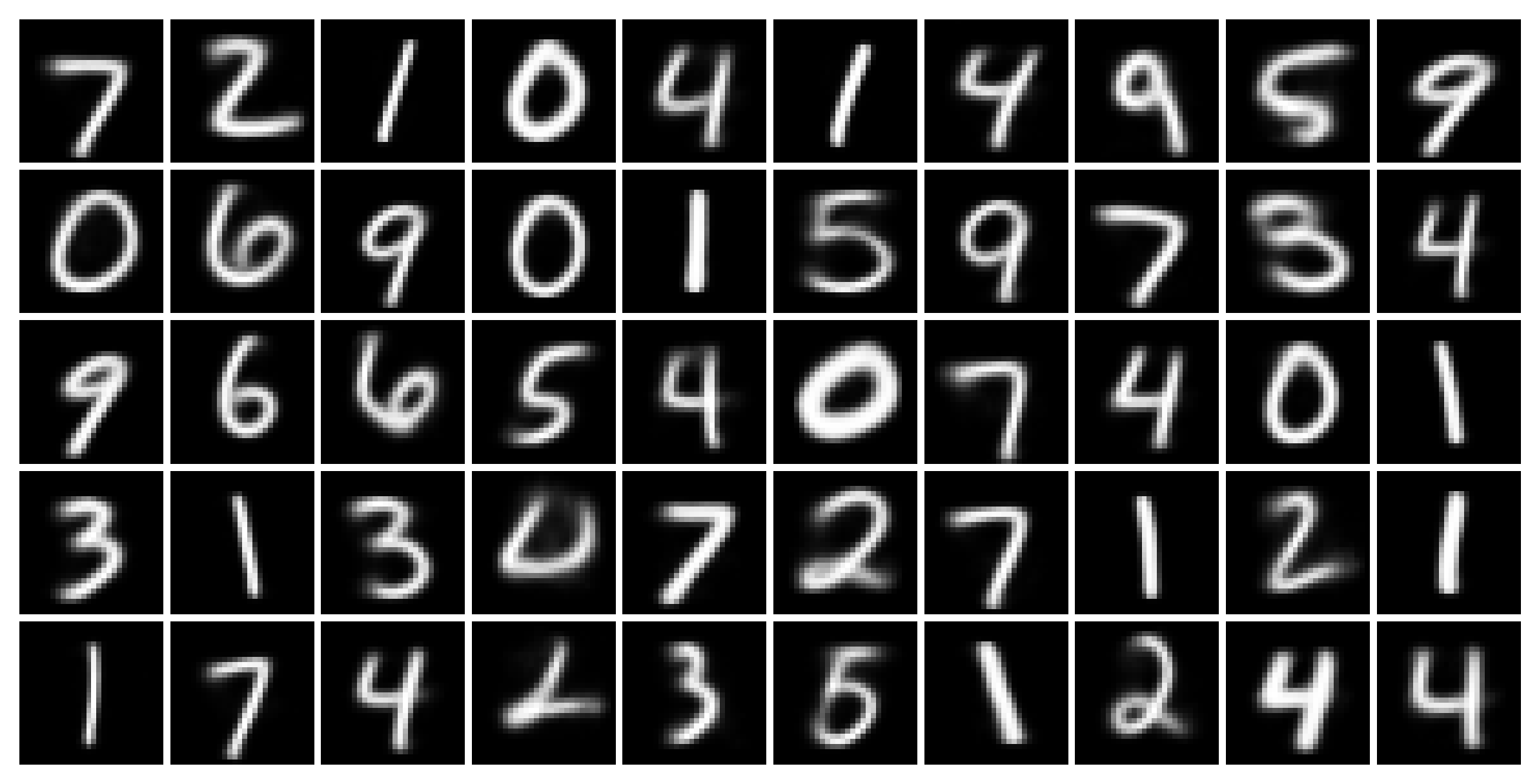}}	
	\subfigure[SWAE ($\beta = 0.5$)]
	{\includegraphics[width=0.24\textwidth]{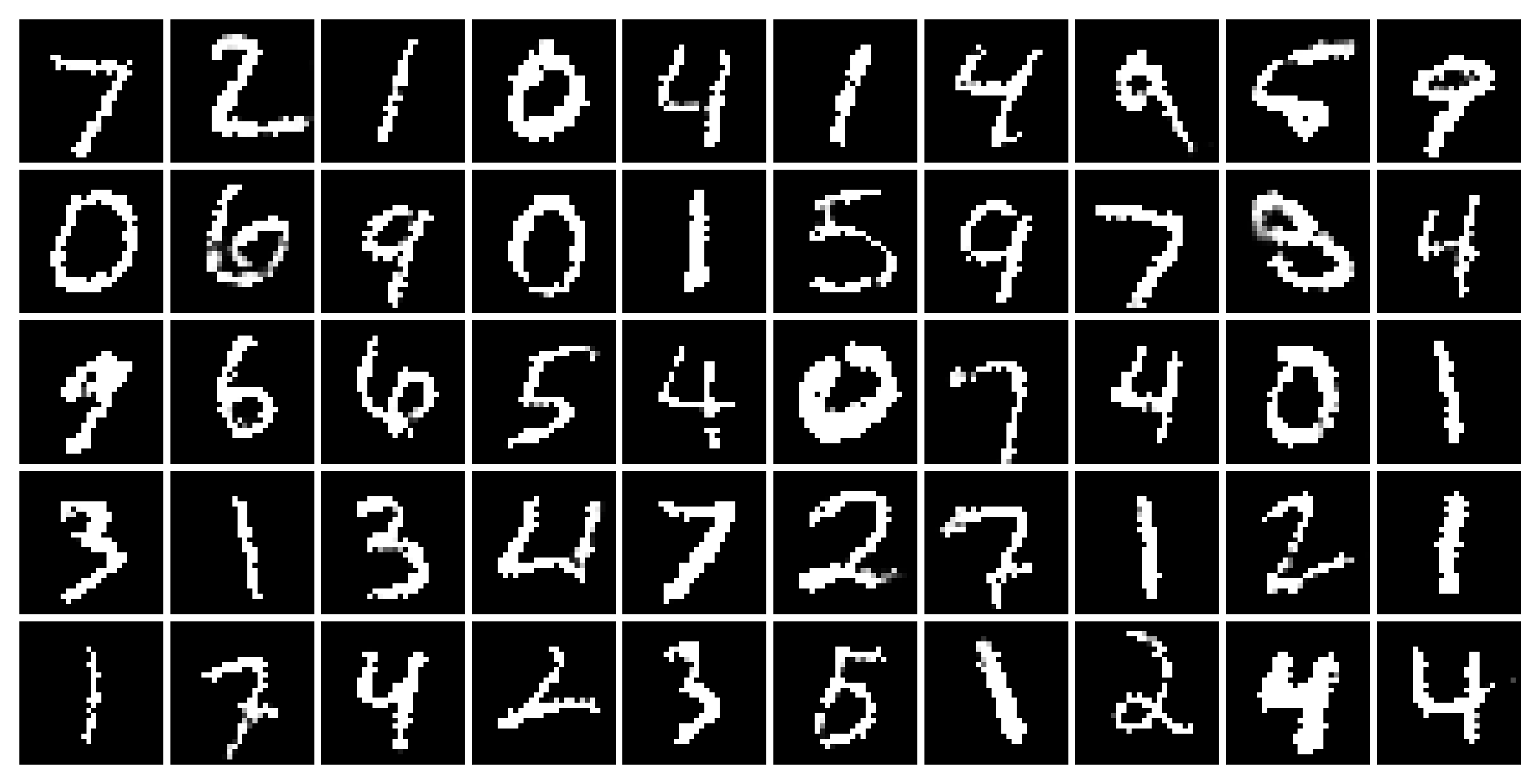}}	
	\subfigure[SWAE ($\beta = 0$)]
	{\includegraphics[width=0.24\textwidth]{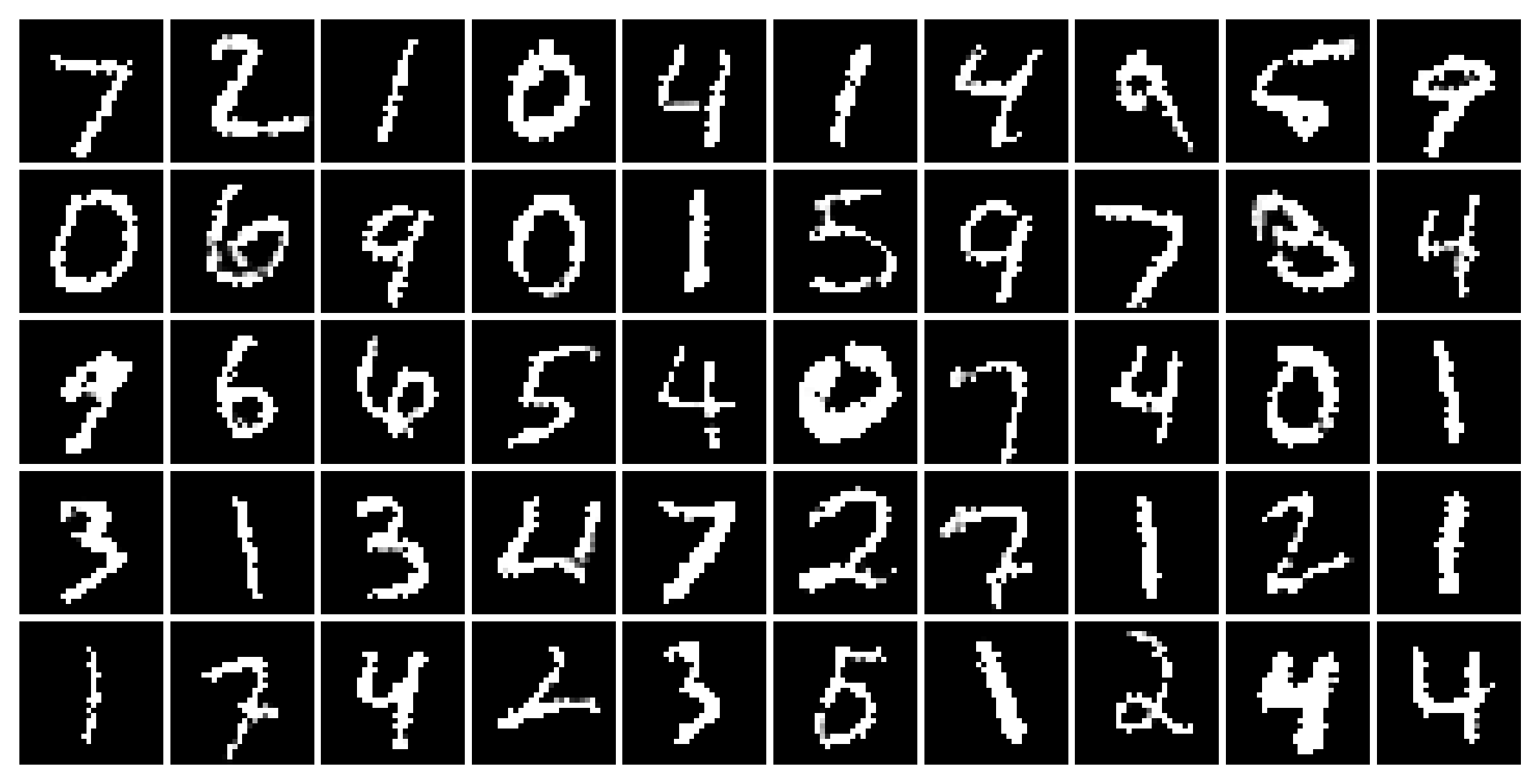}}
	\subfigure[VAE]
	{\includegraphics[width=0.24\textwidth]{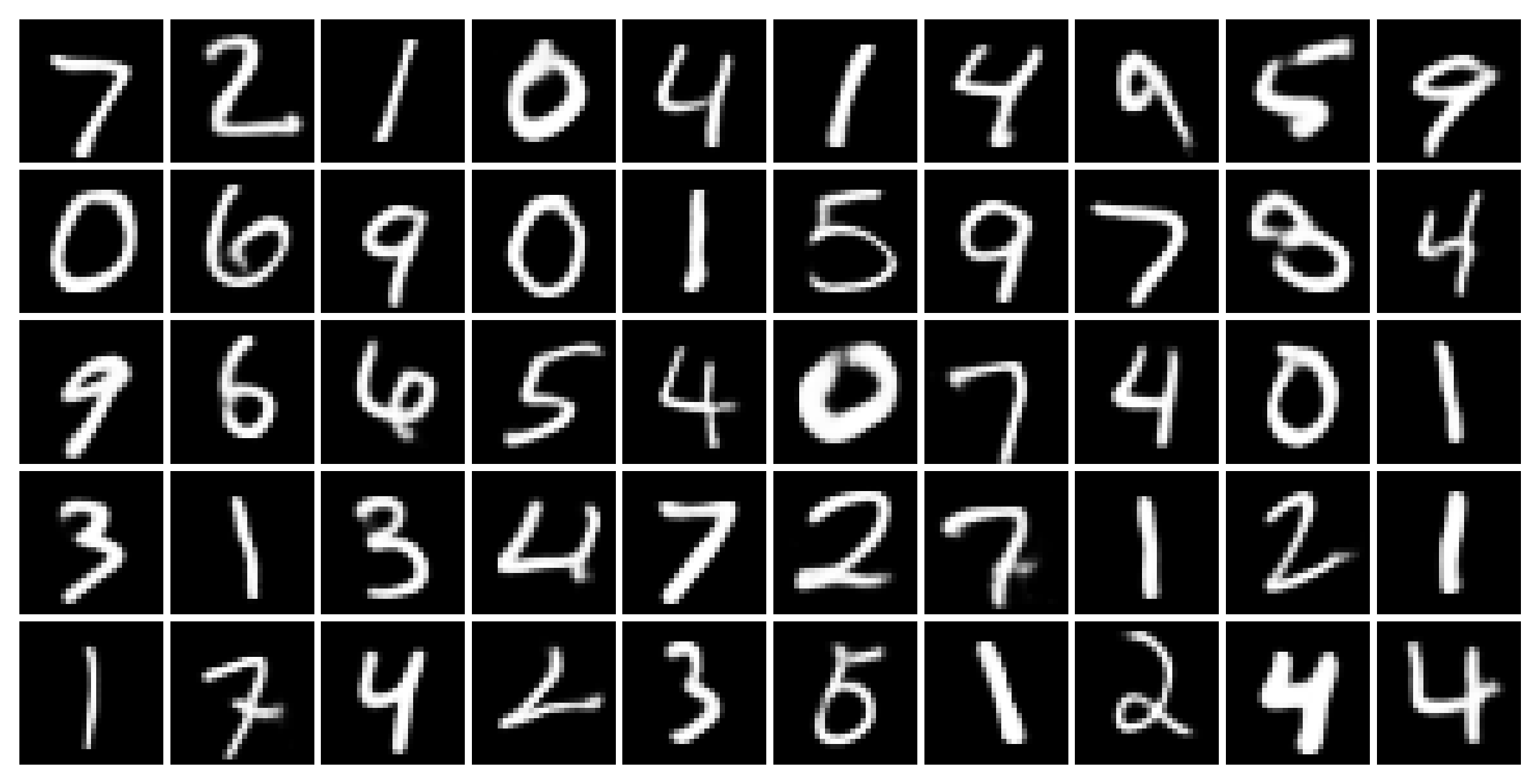}}
		\subfigure[WAE-GAN]
	{\includegraphics[width=0.24\textwidth]{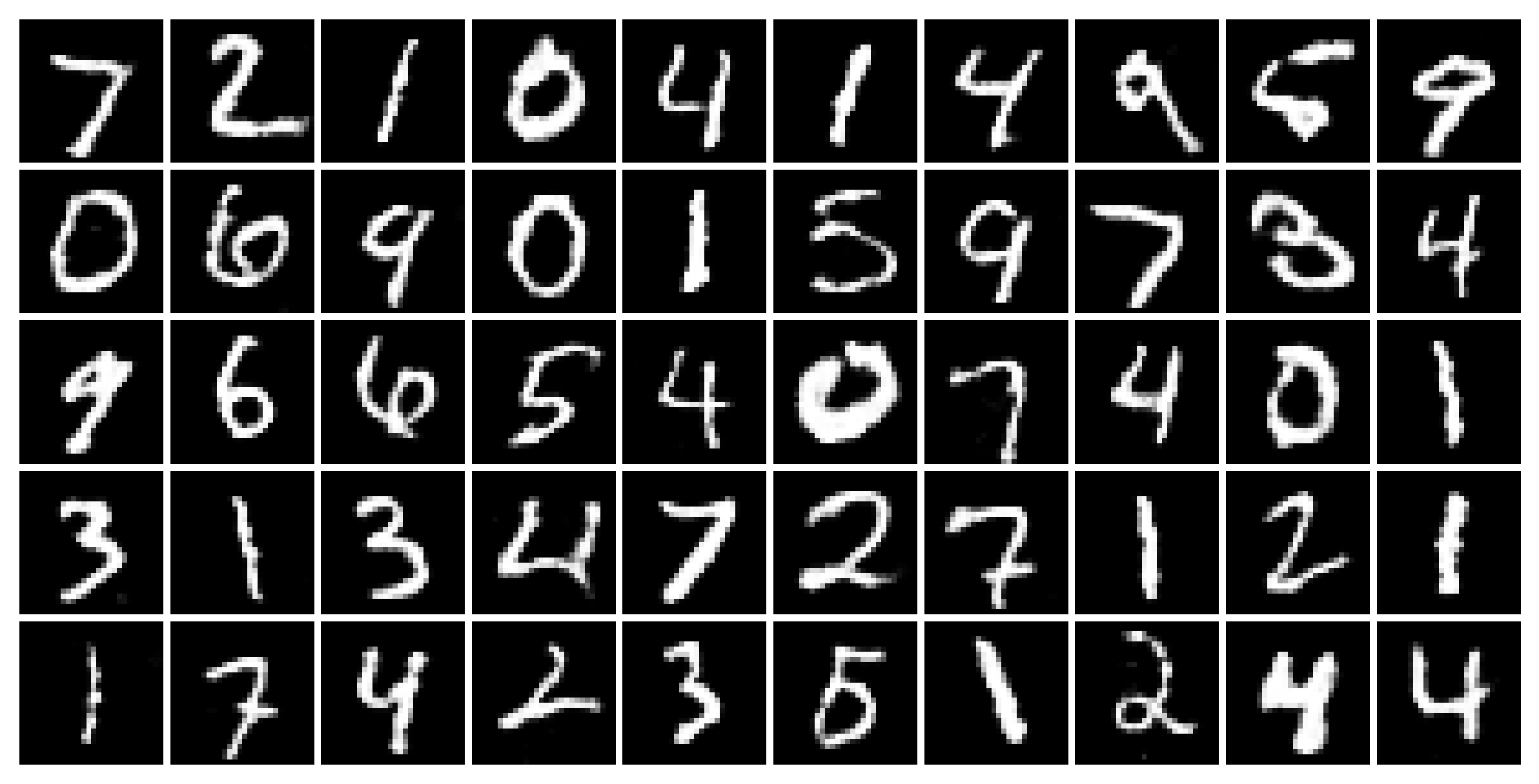}}
		\subfigure[WAE-MMD]
	{\includegraphics[width=0.24\textwidth]{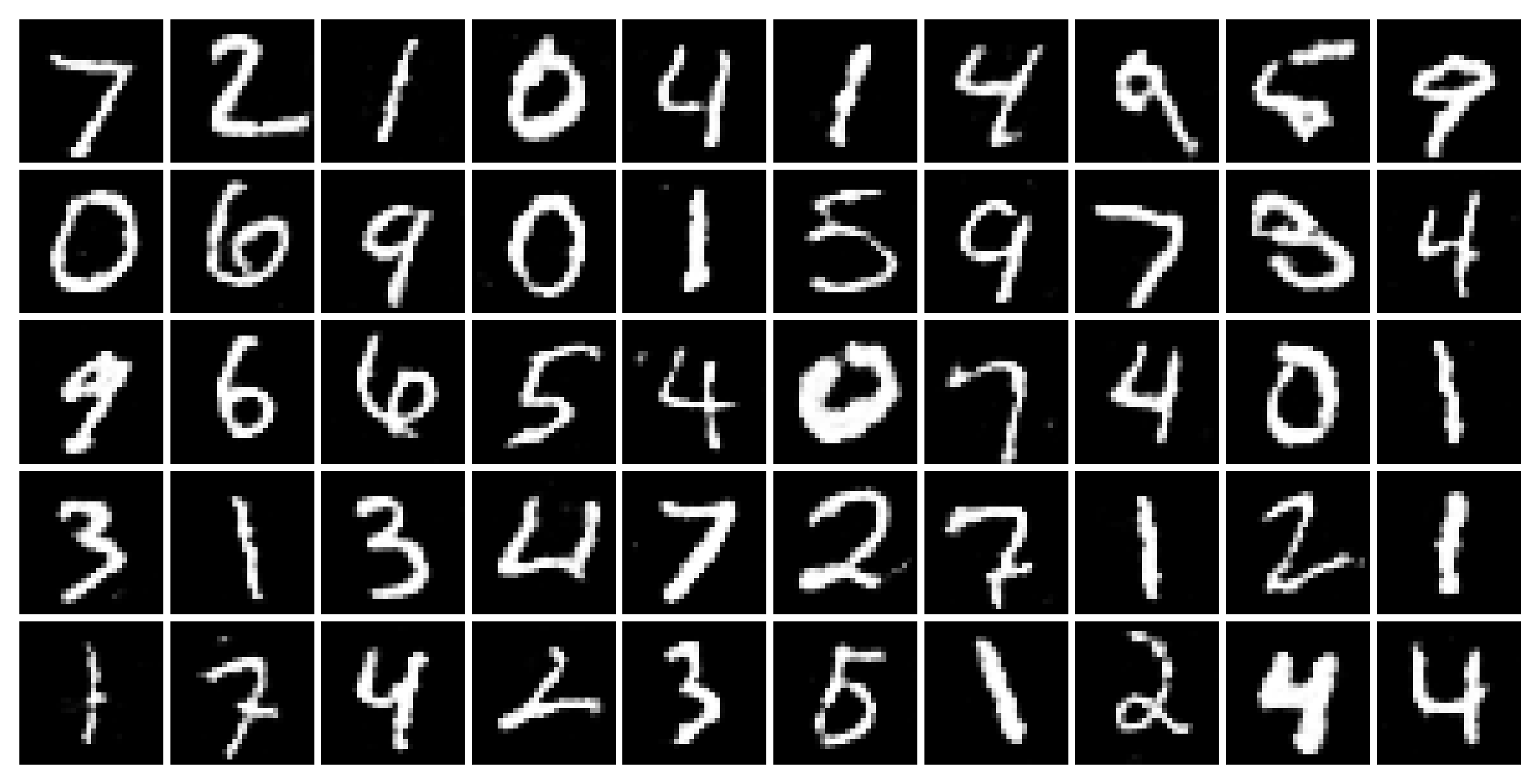}}
		\subfigure[VampPrior]
	{\includegraphics[width=0.24\textwidth]{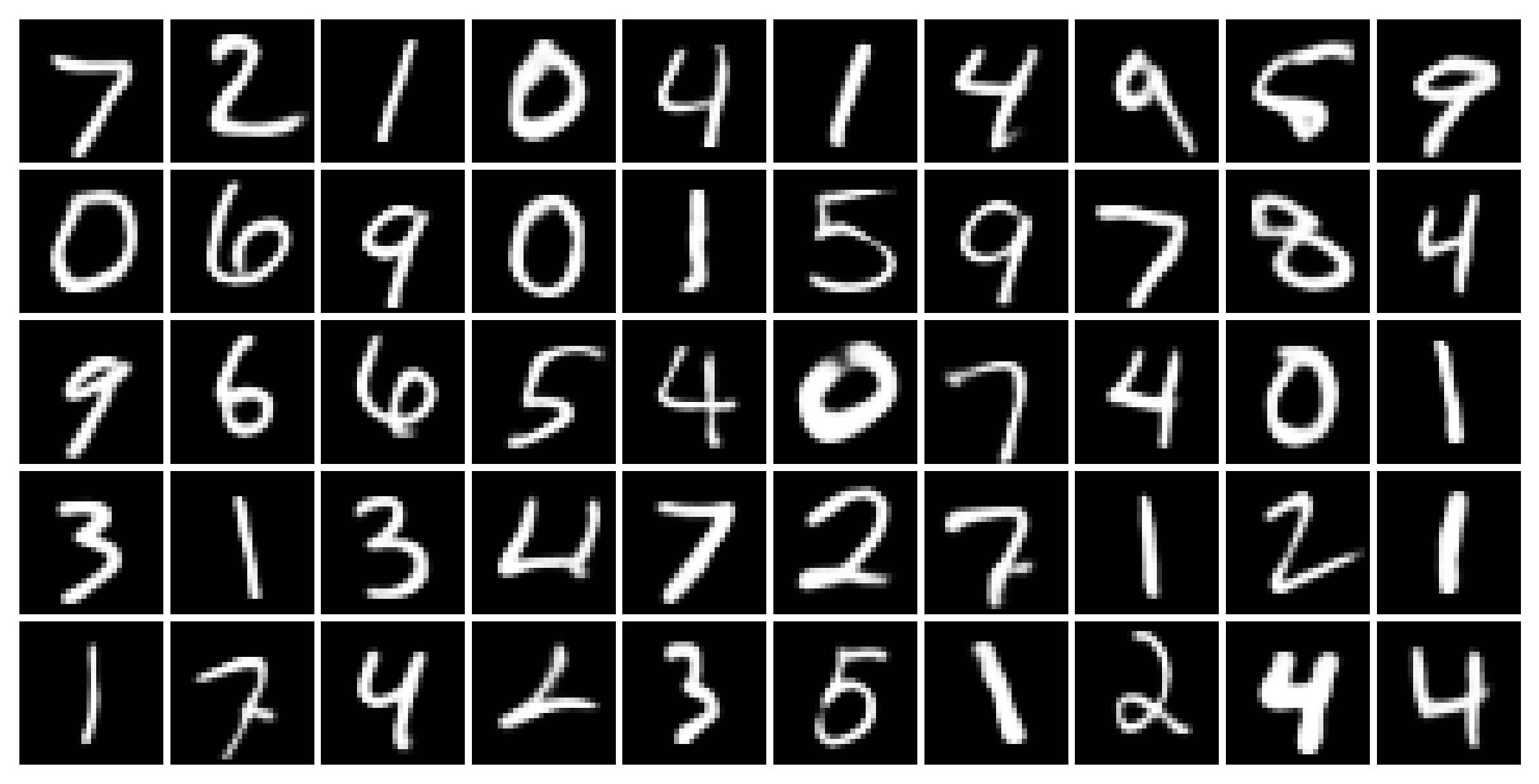}}
		\subfigure[MIM]
	{\includegraphics[width=0.24\textwidth]{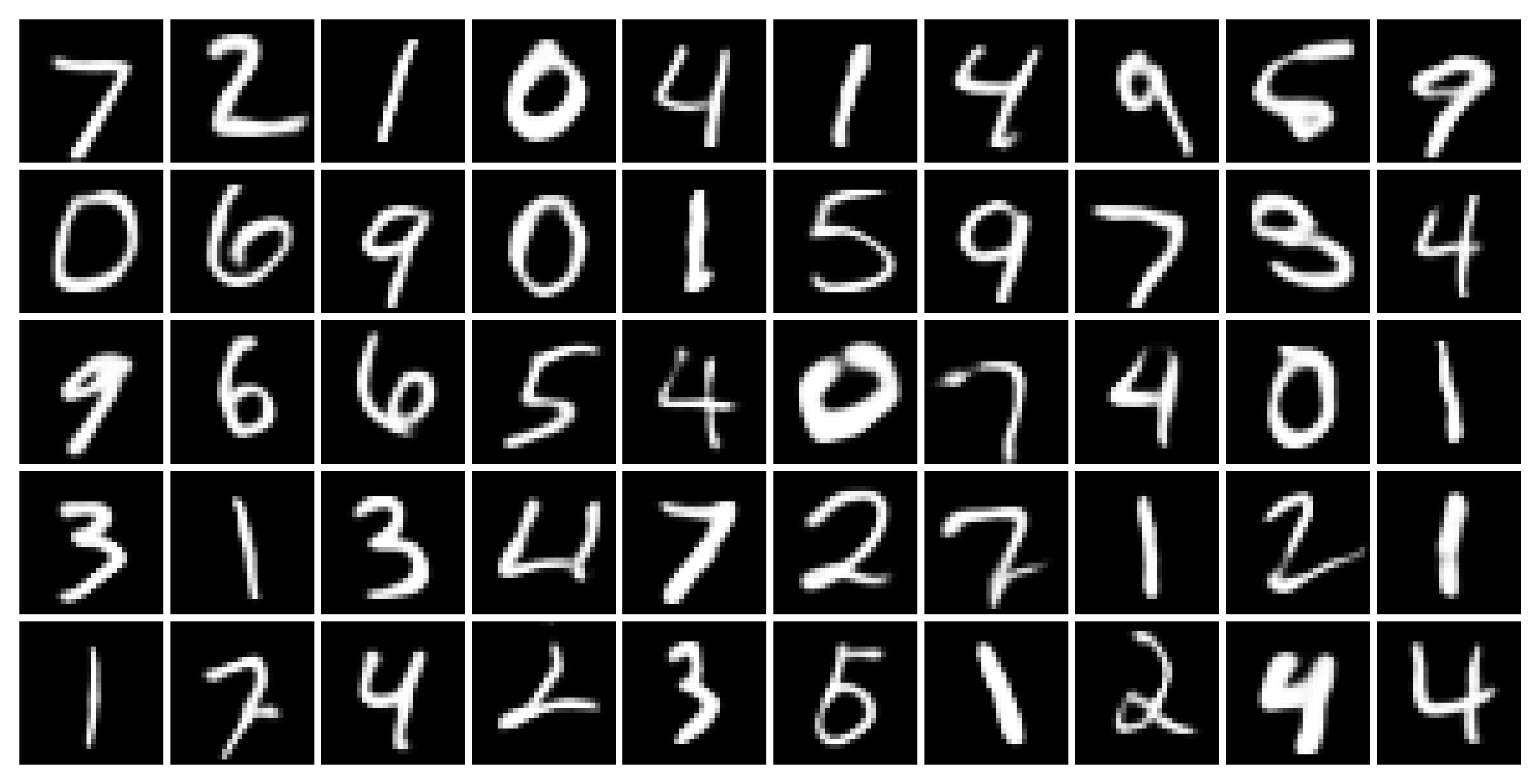}}
	\caption{Reconstructed images on MNIST. dim-$\bz = 80$ for all methods. As expected, for SWAEs a smaller $\beta$ leads to a higher quality of reconstruction.}
\end{figure*}

\begin{figure*}[h]
	\centering
	\subfigure[Real images]
	{\includegraphics[width=0.24\textwidth]{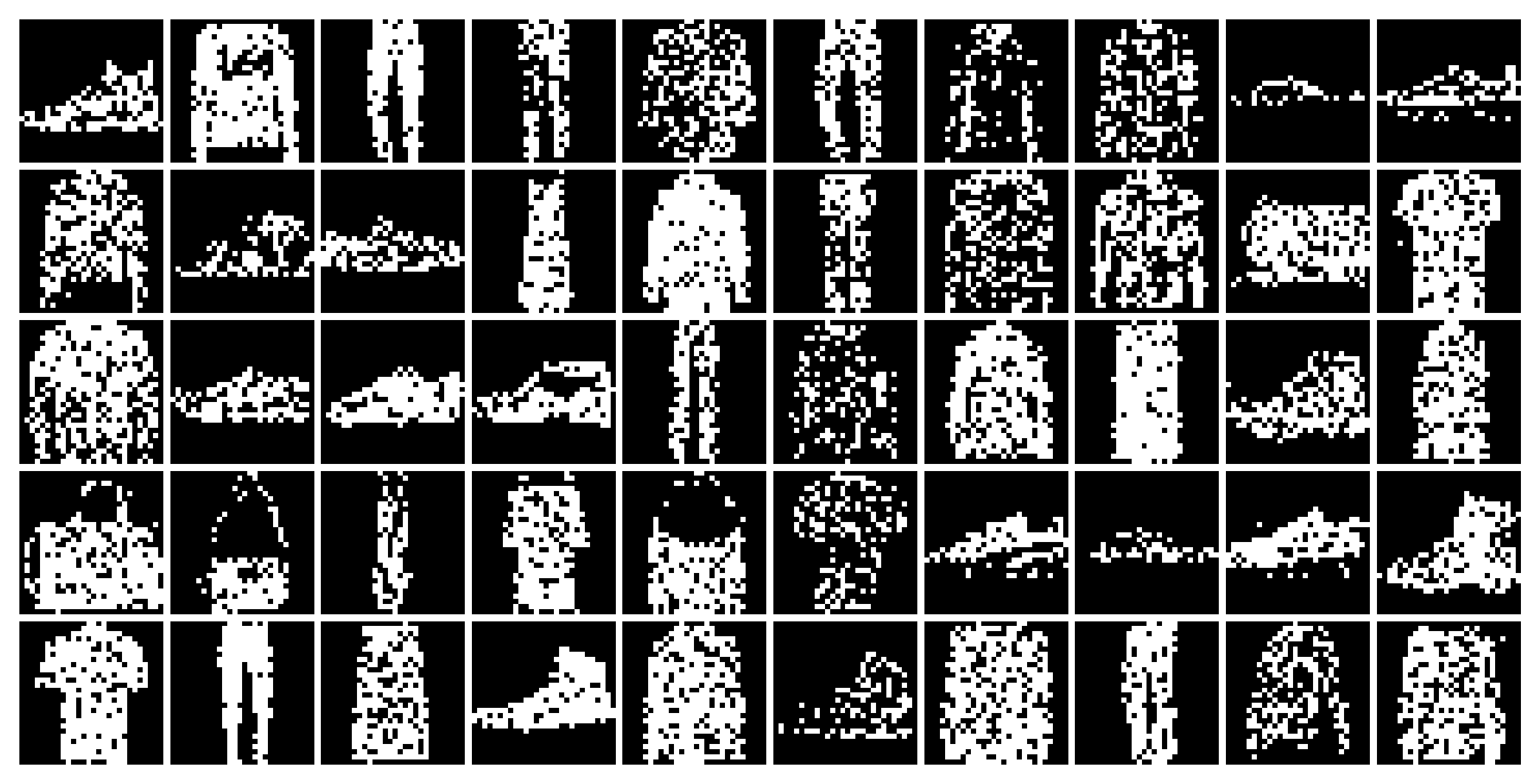}}	
	\subfigure[SWAE ($\beta = 1$)]
	{\includegraphics[width=0.24\textwidth]{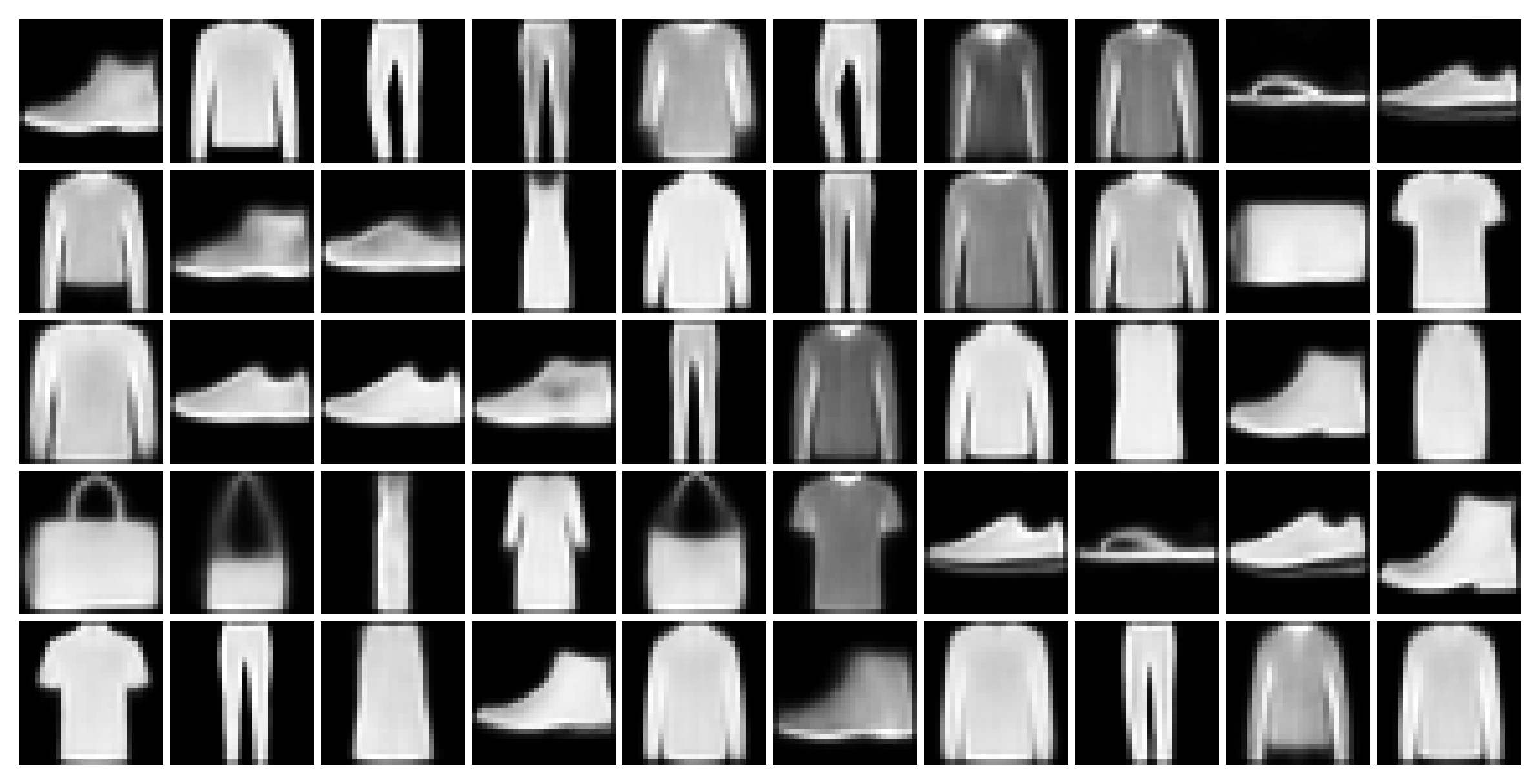}}	
	\subfigure[SWAE ($\beta = 0.5$)]
	{\includegraphics[width=0.24\textwidth]{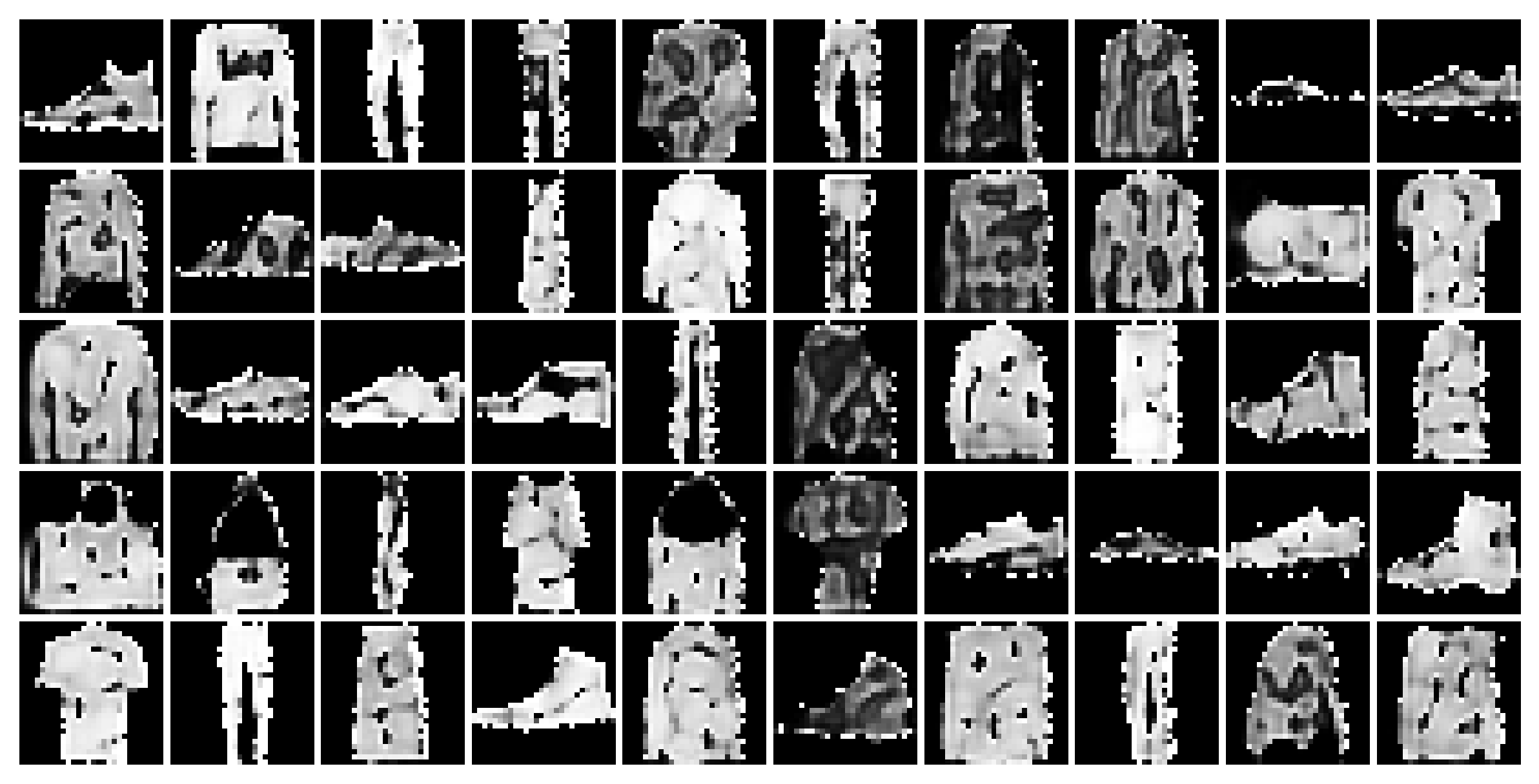}}	
	\subfigure[SWAE ($\beta = 0$)]
	{\includegraphics[width=0.24\textwidth]{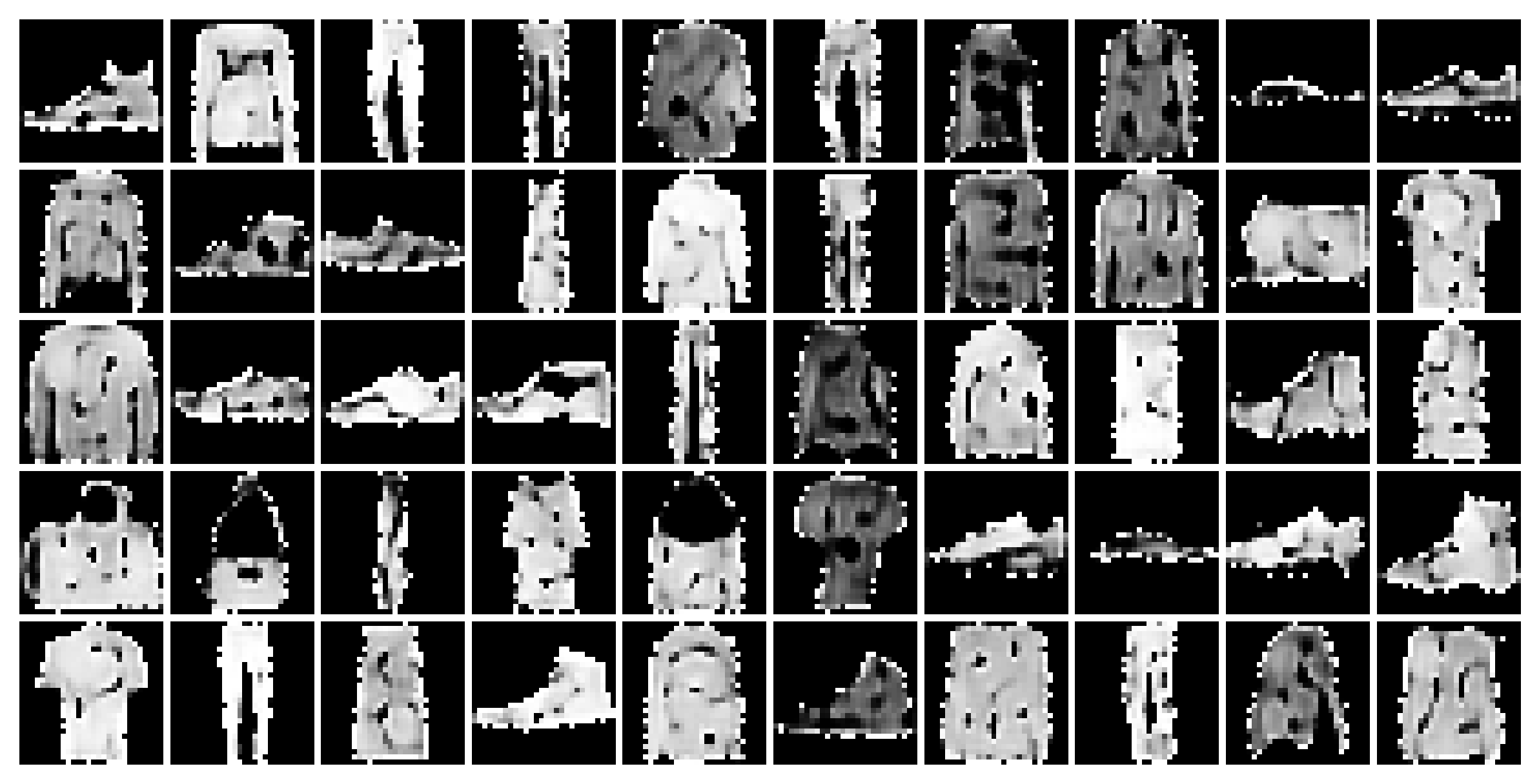}}
	\subfigure[VAE]
	{\includegraphics[width=0.24\textwidth]{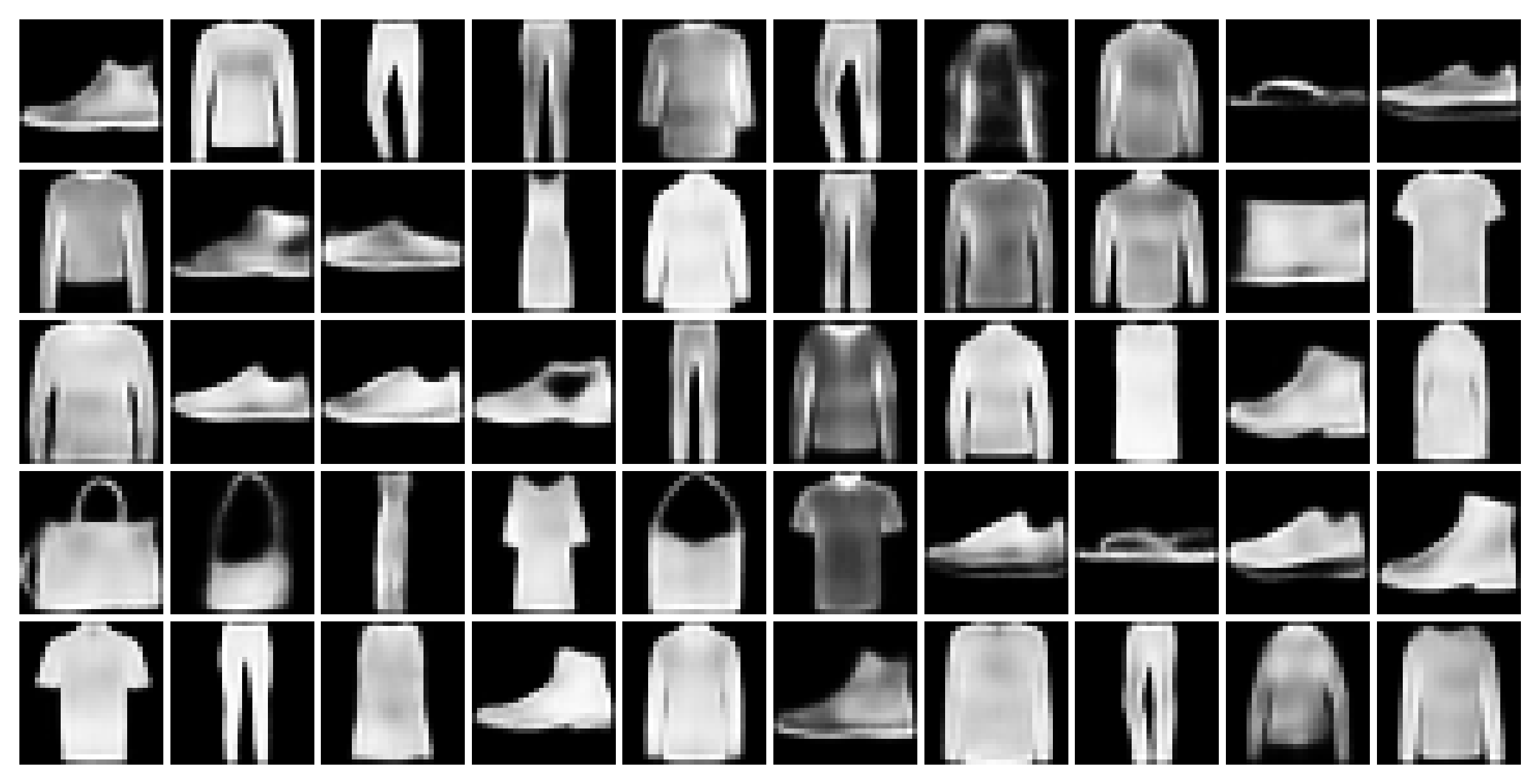}}
	\subfigure[WAE-GAN]
	{\includegraphics[width=0.24\textwidth]{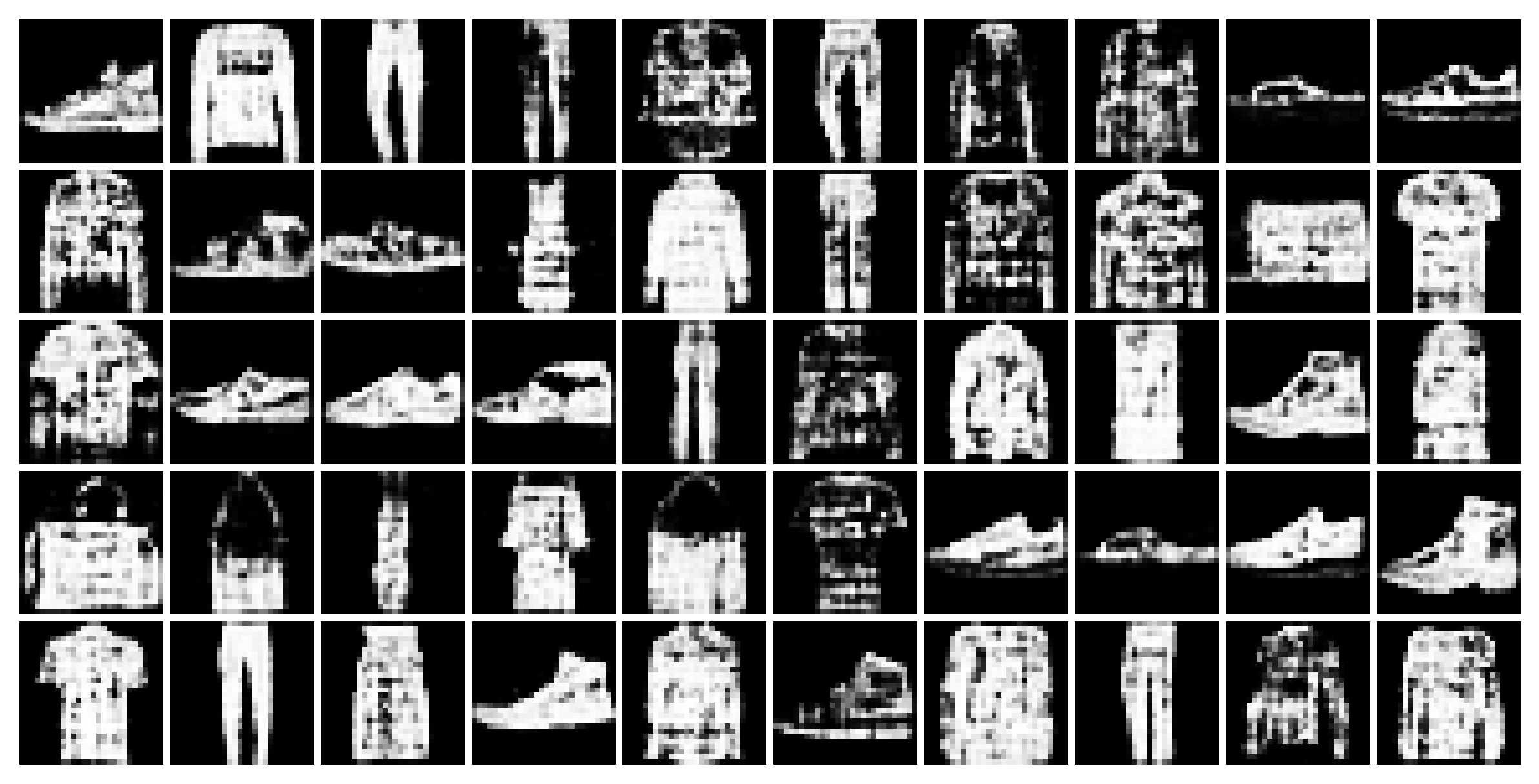}}
	\subfigure[WAE-MMD]
	{\includegraphics[width=0.24\textwidth]{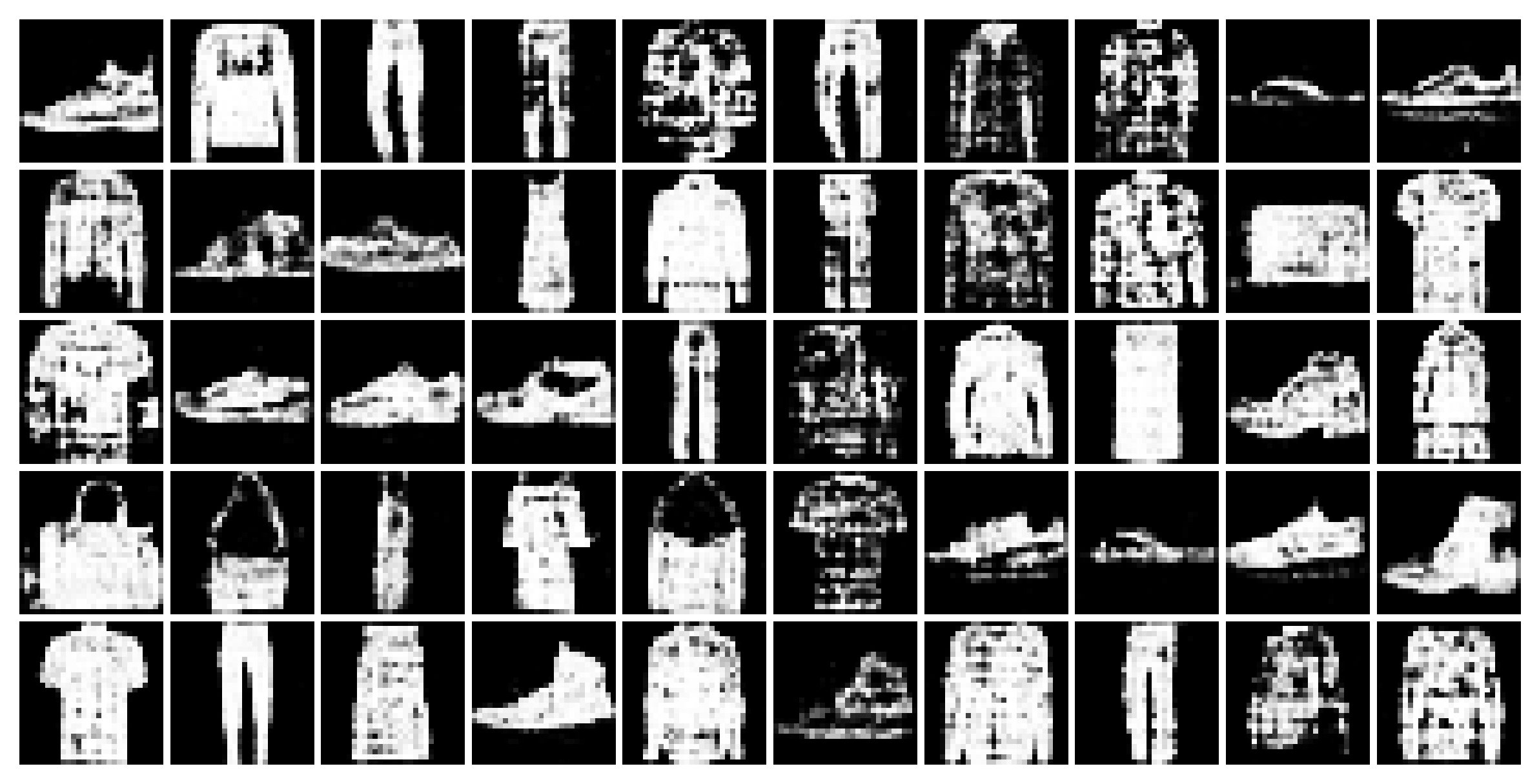}}
	\subfigure[VampPrior]
	{\includegraphics[width=0.24\textwidth]{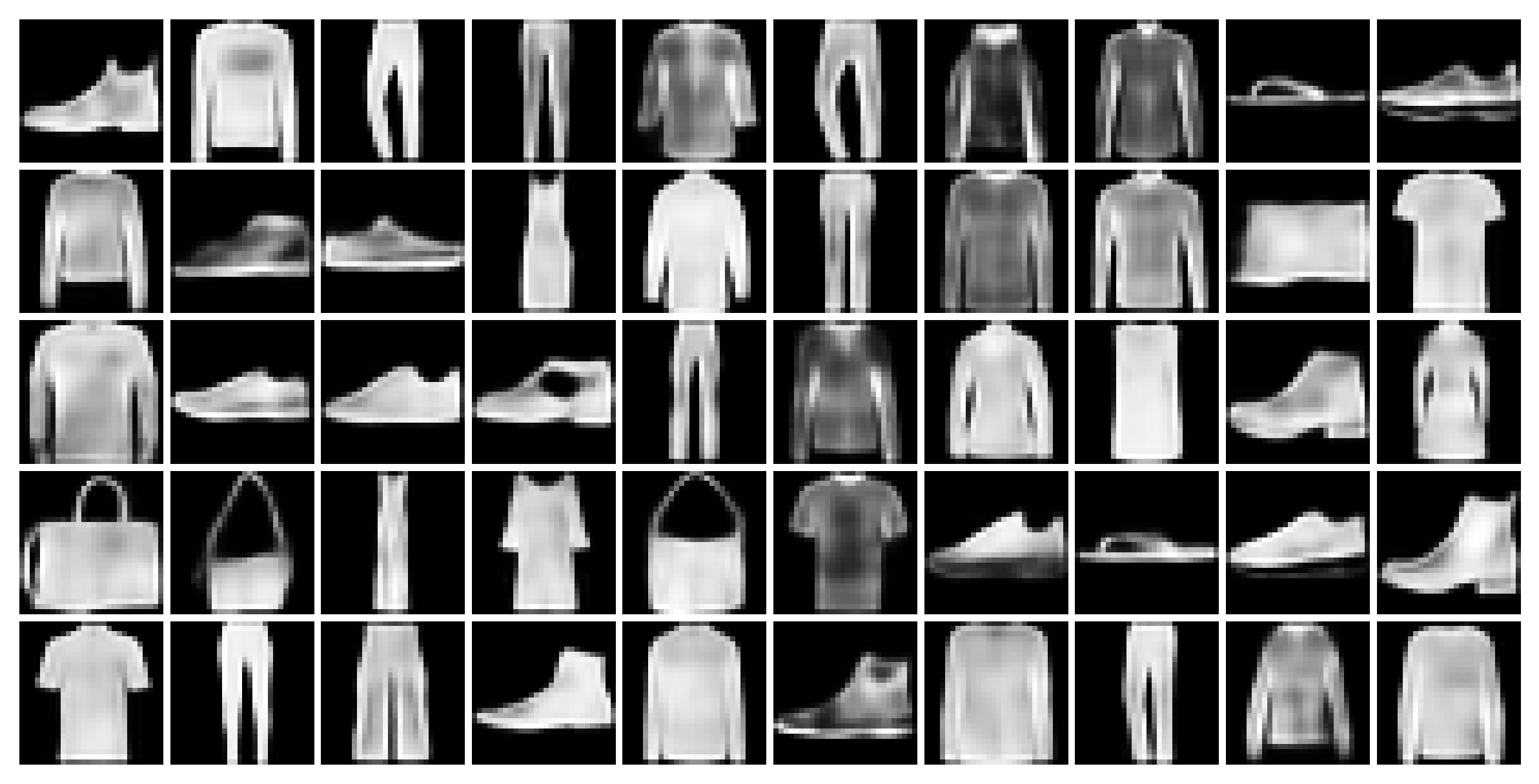}}
	\subfigure[MIM]
	{\includegraphics[width=0.24\textwidth]{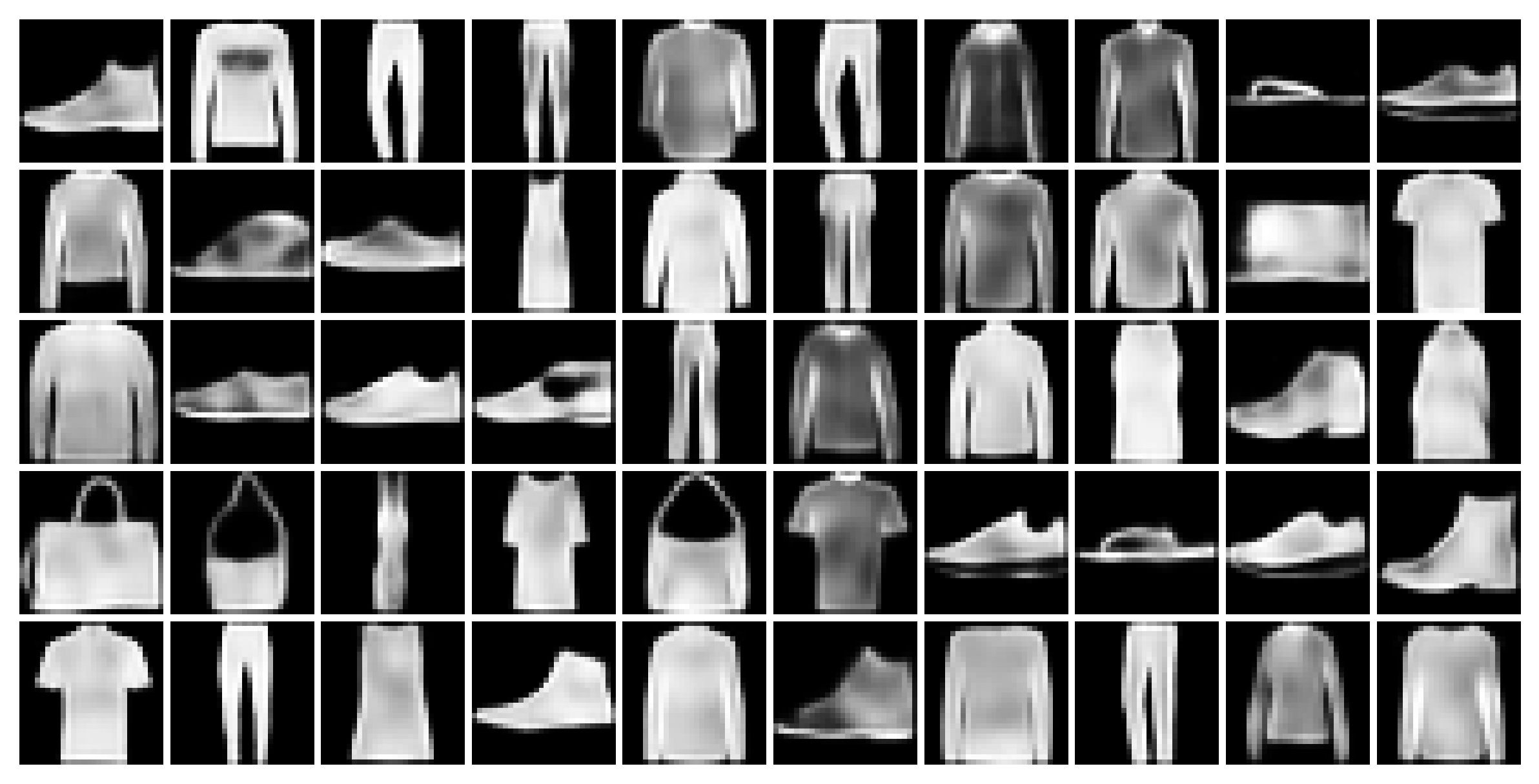}}
	\caption{Reconstructed images on Fashion-MNIST. dim-$\bz = 80$ for all methods.}
\end{figure*}

\begin{figure*}[h]
	\centering
	\subfigure[Real images]
	{\includegraphics[width=0.24\textwidth]{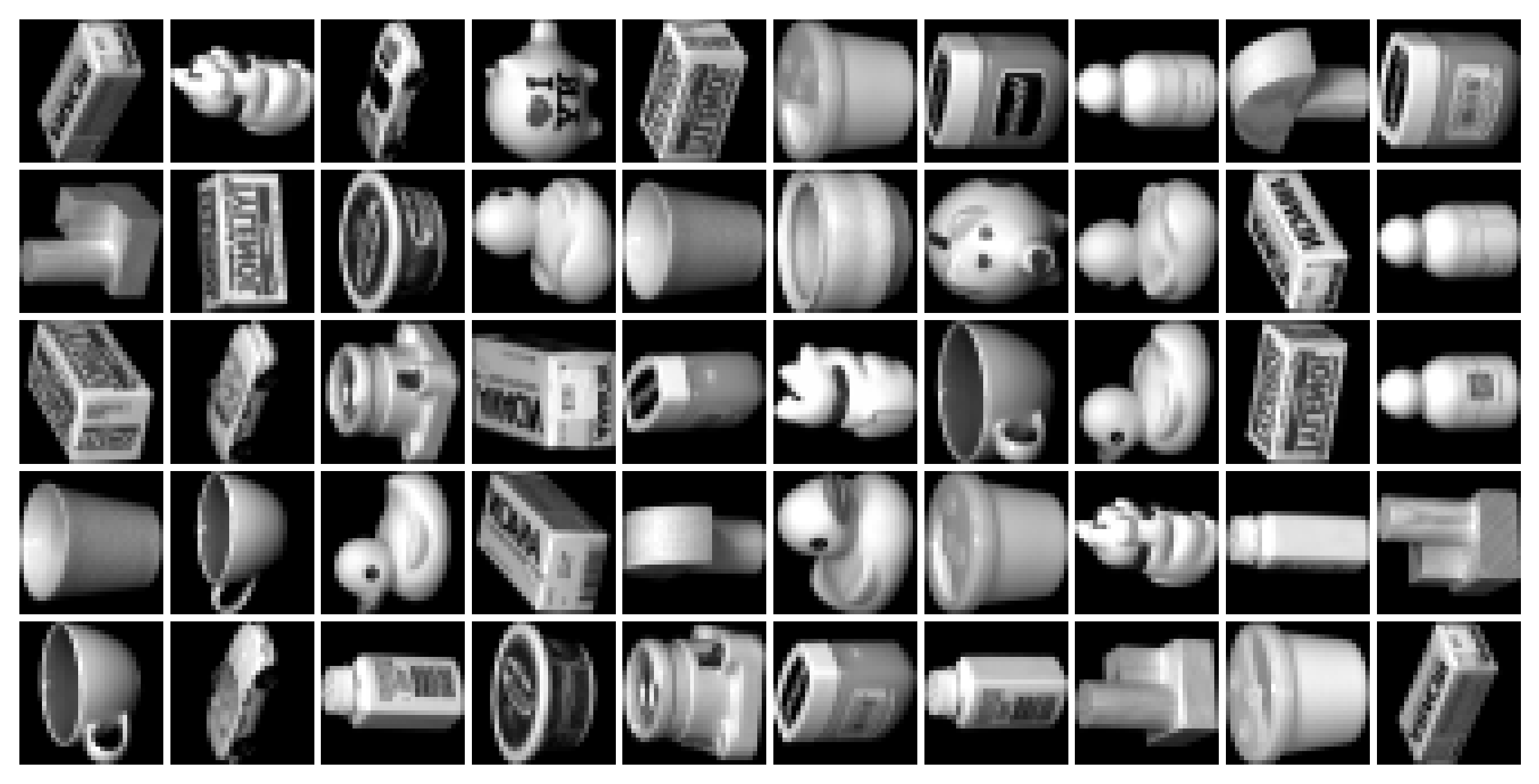}}	
	\subfigure[SWAE ($\beta = 1$)]
	{\includegraphics[width=0.24\textwidth]{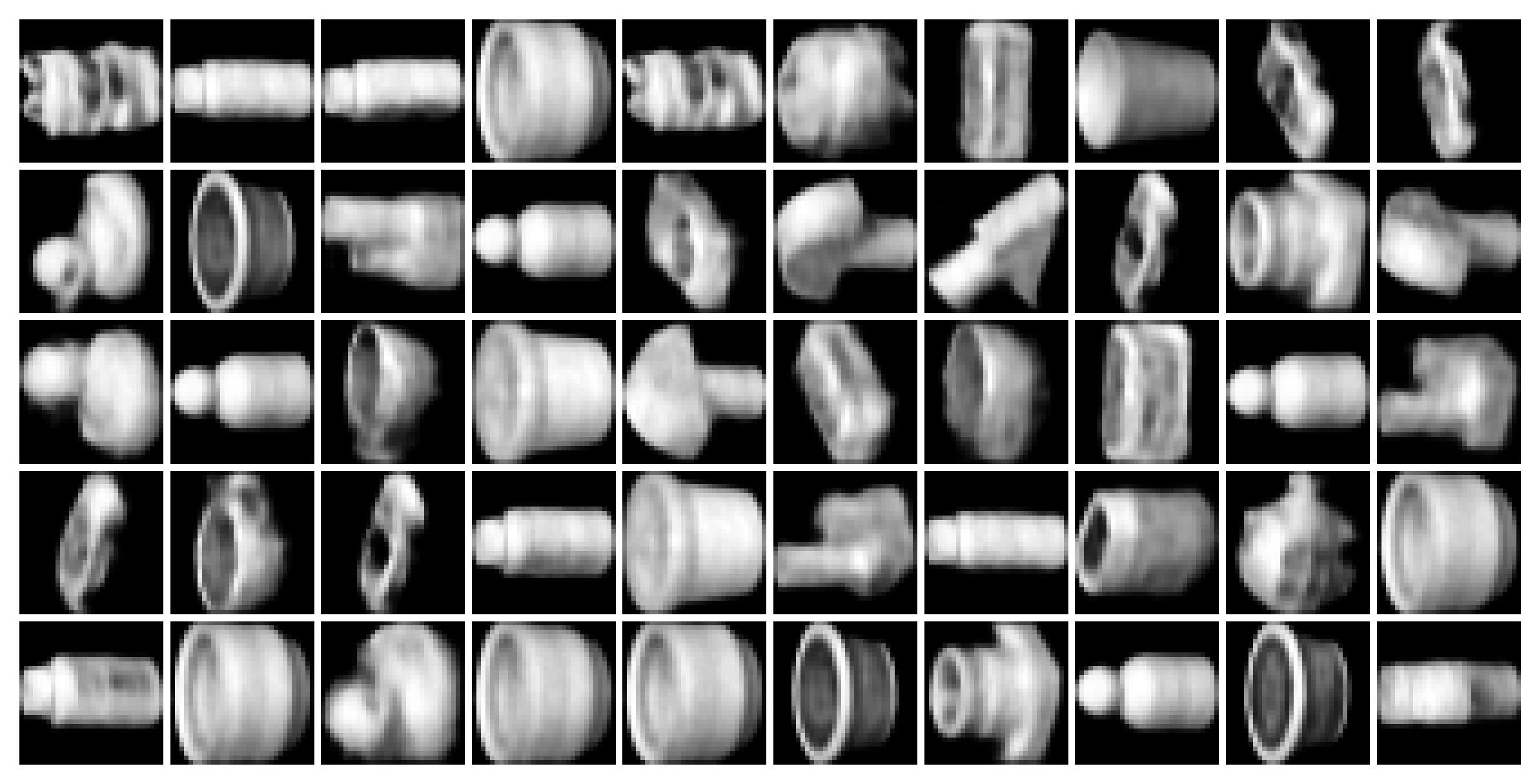}}	
	\subfigure[SWAE ($\beta = 0.5$)]
	{\includegraphics[width=0.24\textwidth]{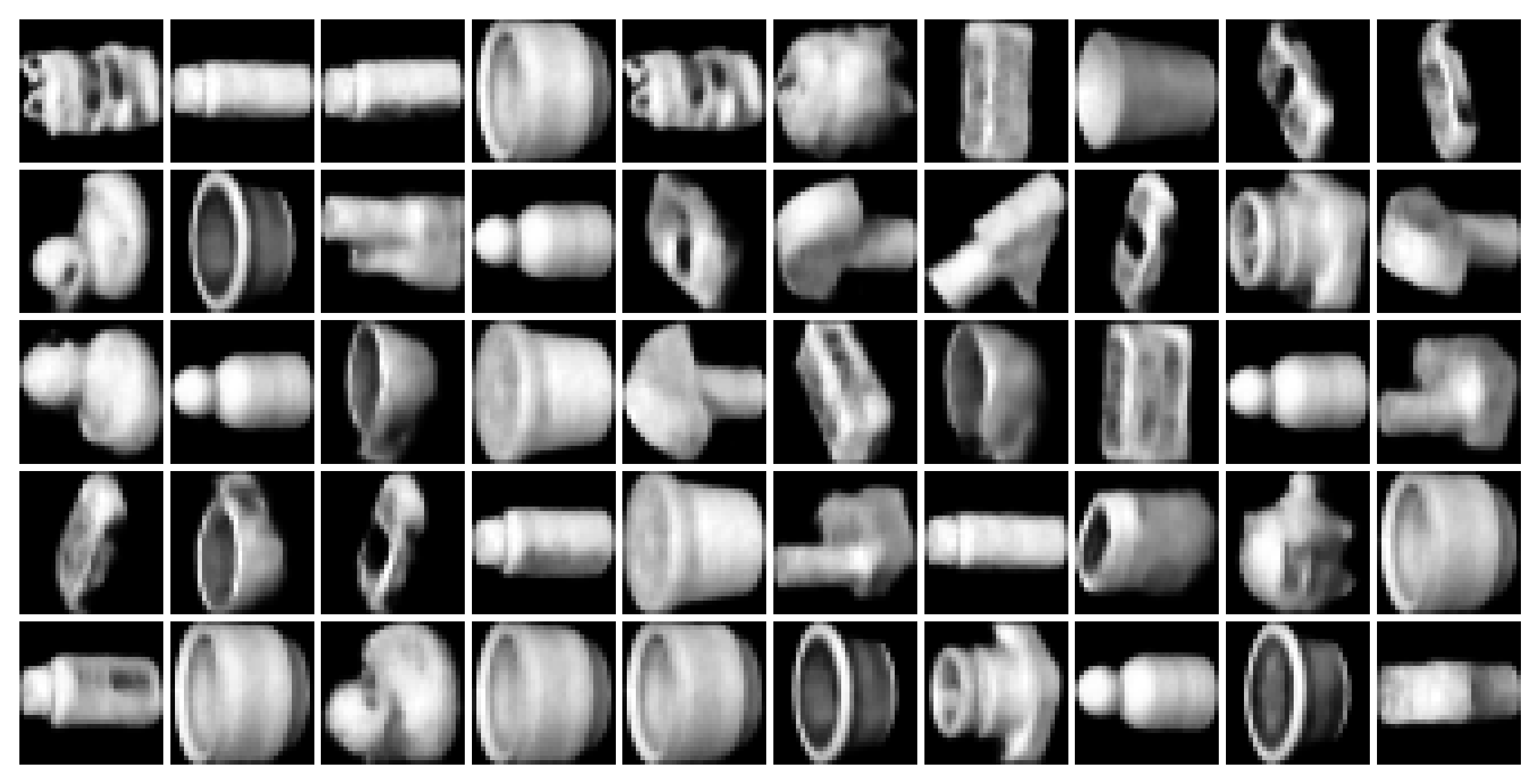}}	
	\subfigure[SWAE ($\beta = 0$)]
	{\includegraphics[width=0.24\textwidth]{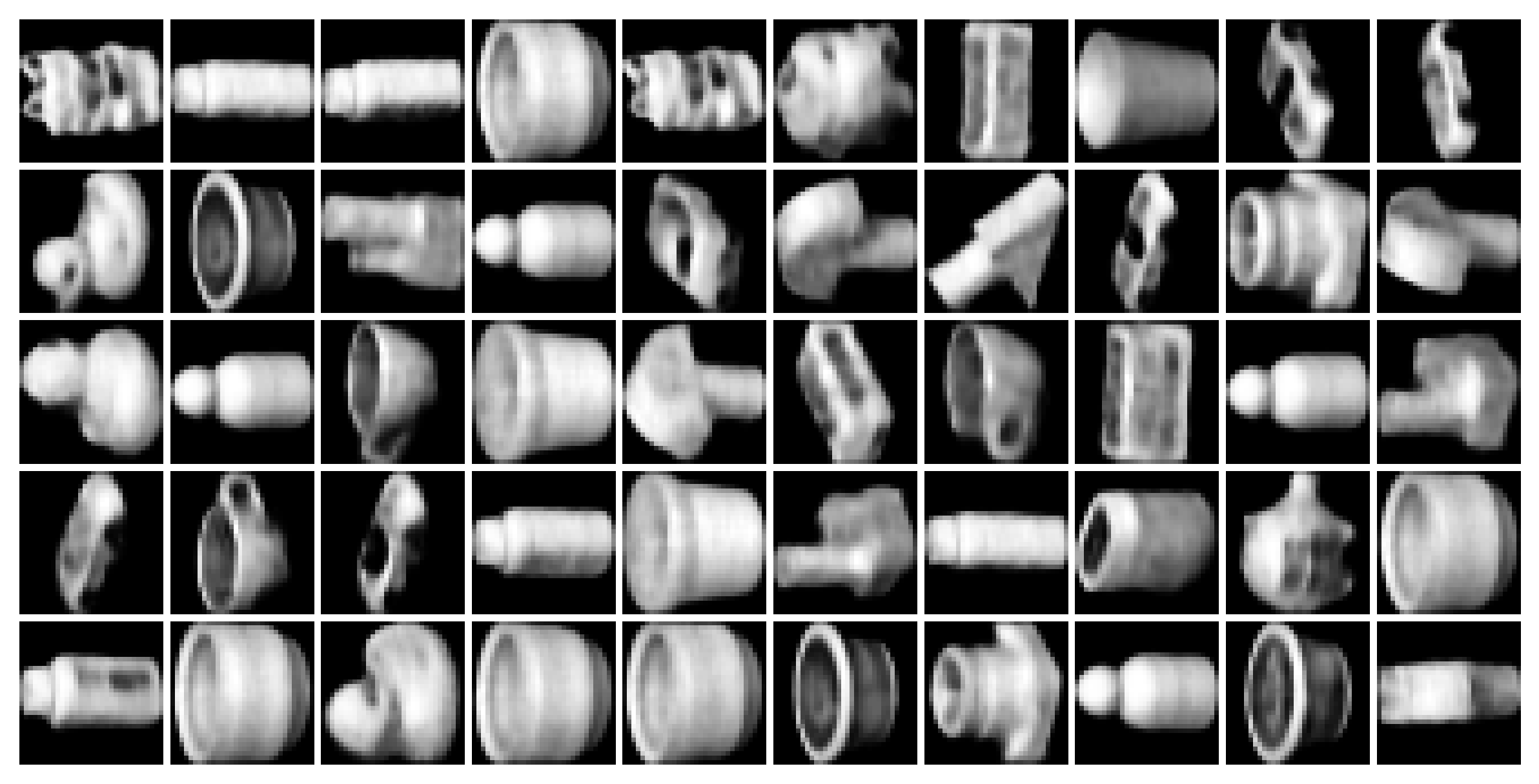}}
	\subfigure[VAE]
	{\includegraphics[width=0.24\textwidth]{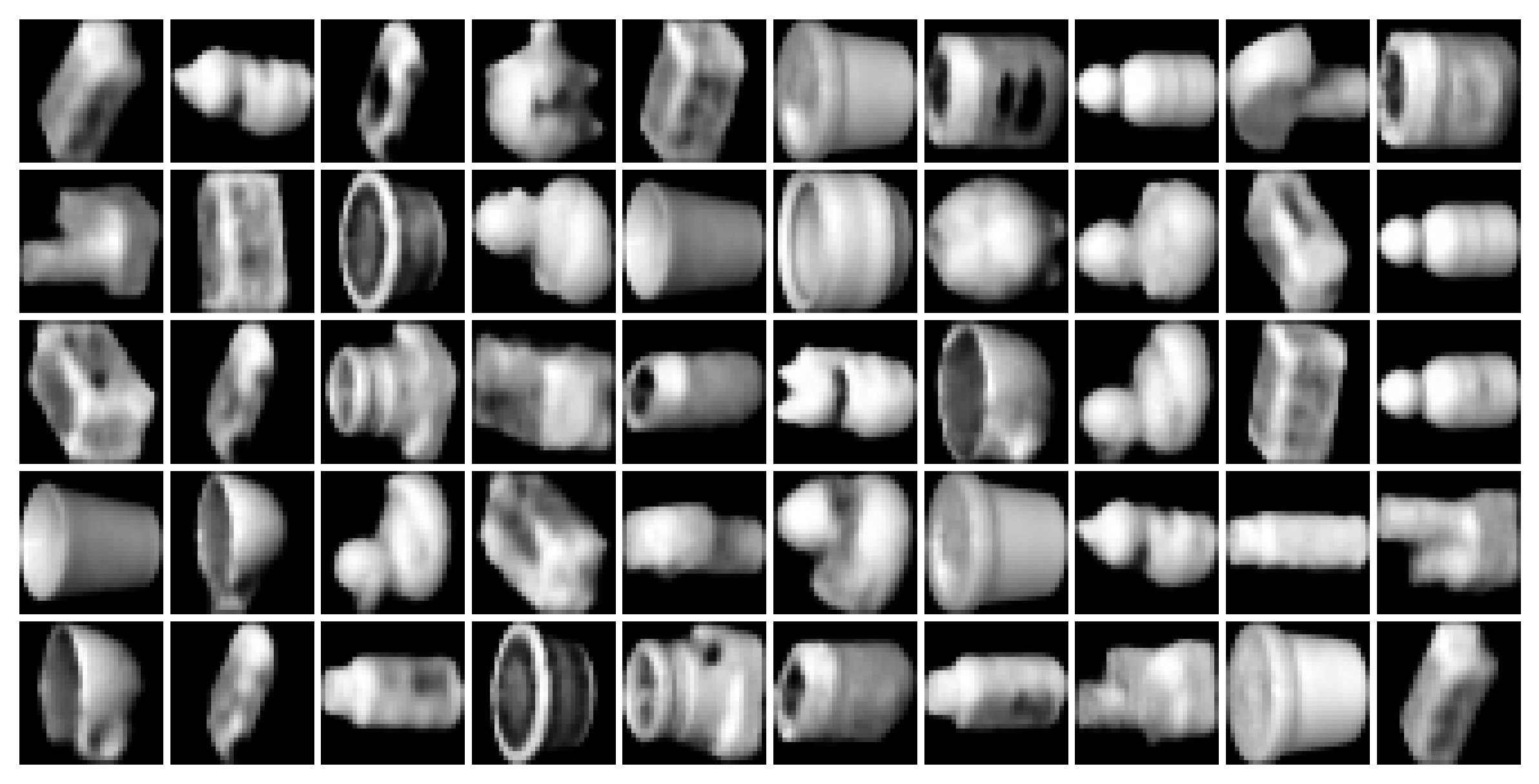}}
	\subfigure[WAE-GAN]
	{\includegraphics[width=0.24\textwidth]{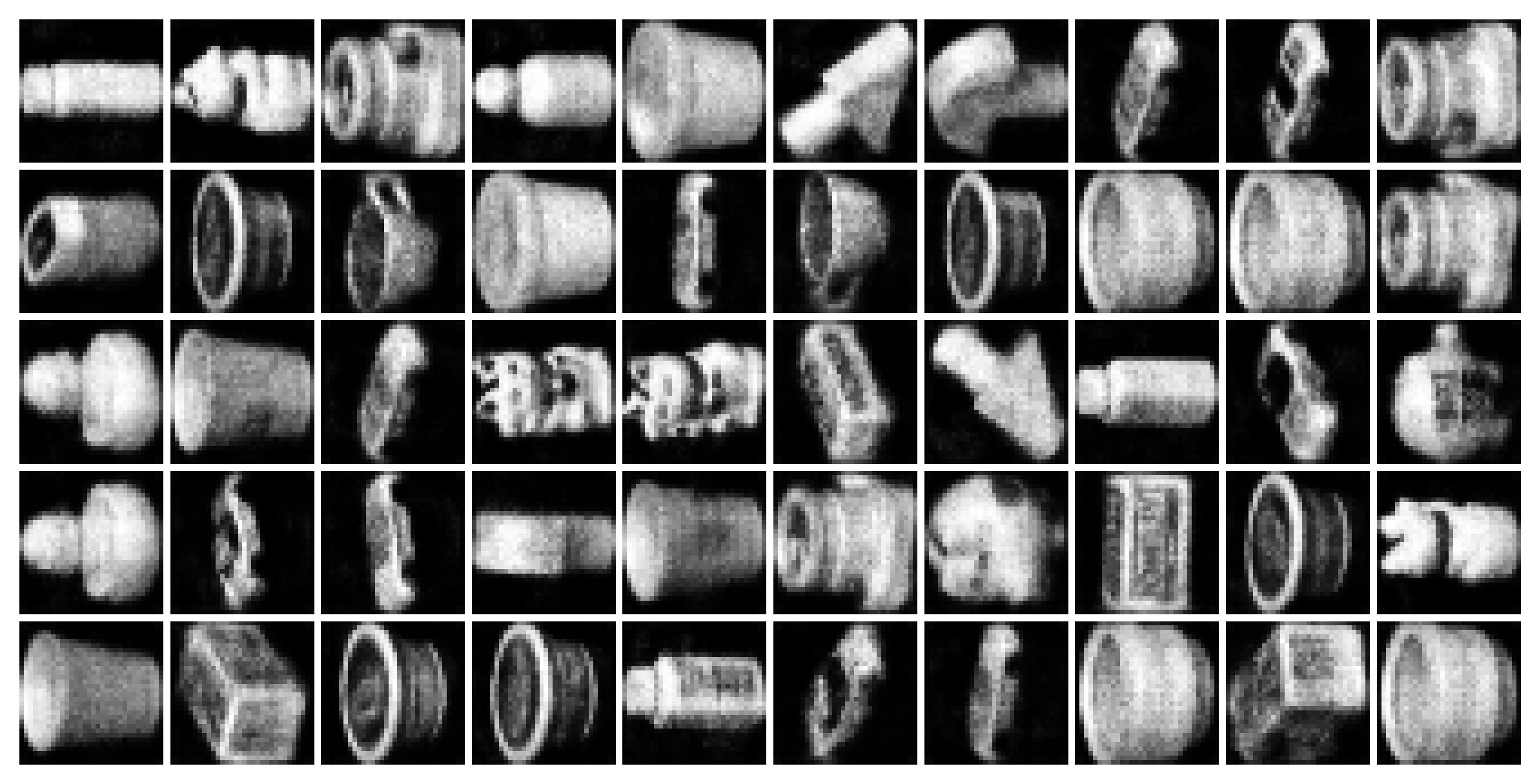}}
	\subfigure[WAE-MMD]
	{\includegraphics[width=0.24\textwidth]{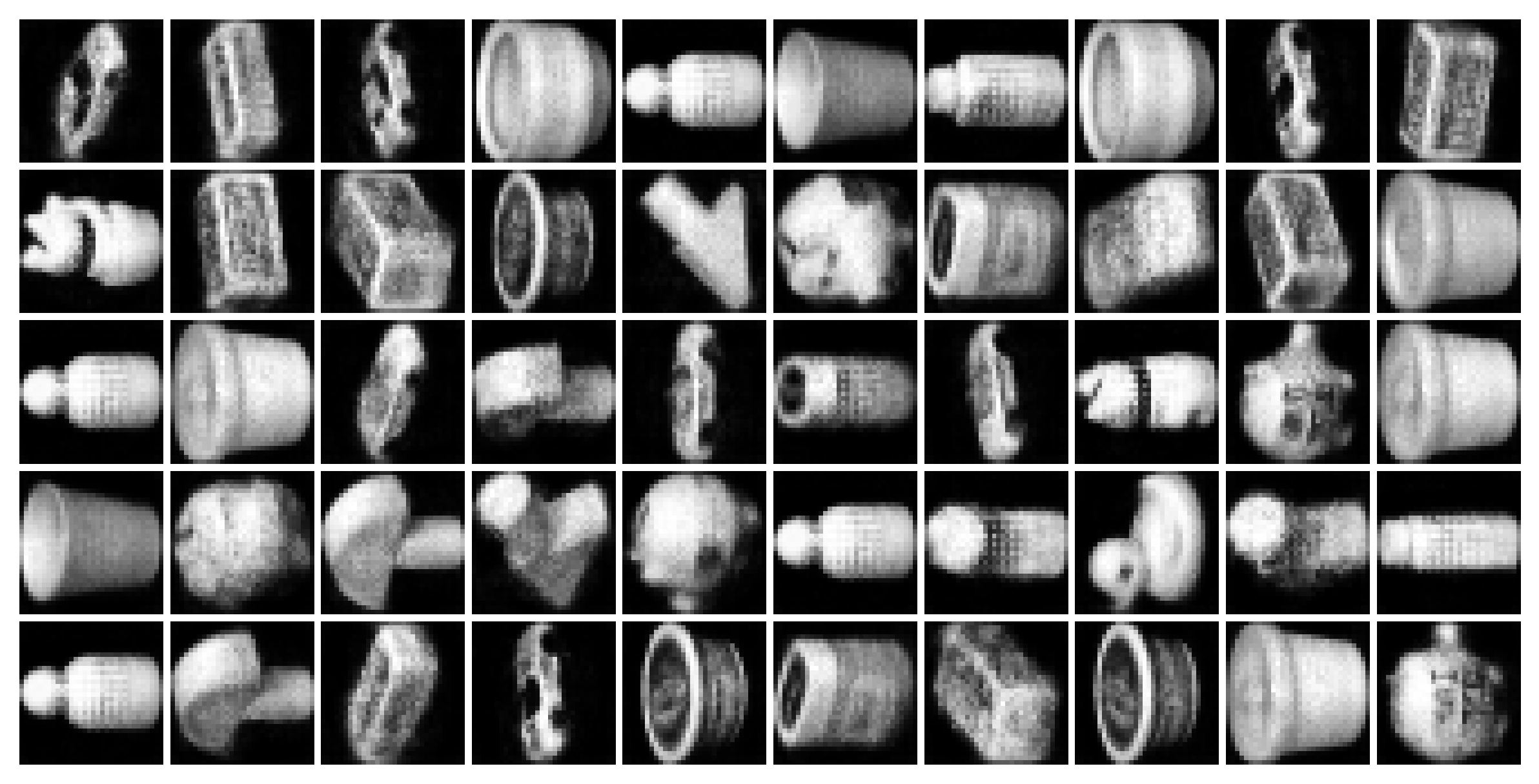}}
	\subfigure[VampPrior]
	{\includegraphics[width=0.24\textwidth]{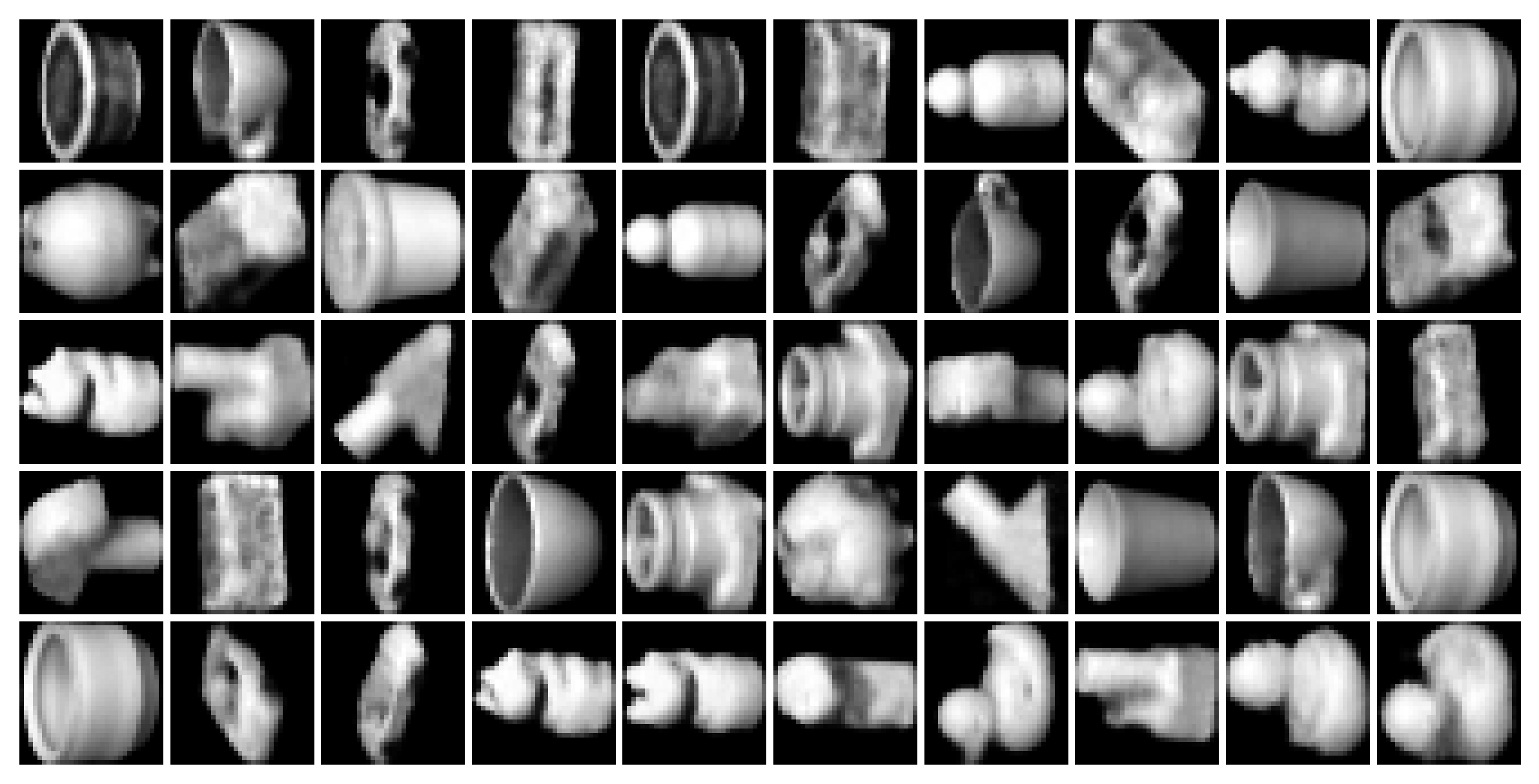}}
	\subfigure[MIM]
	{\includegraphics[width=0.24\textwidth]{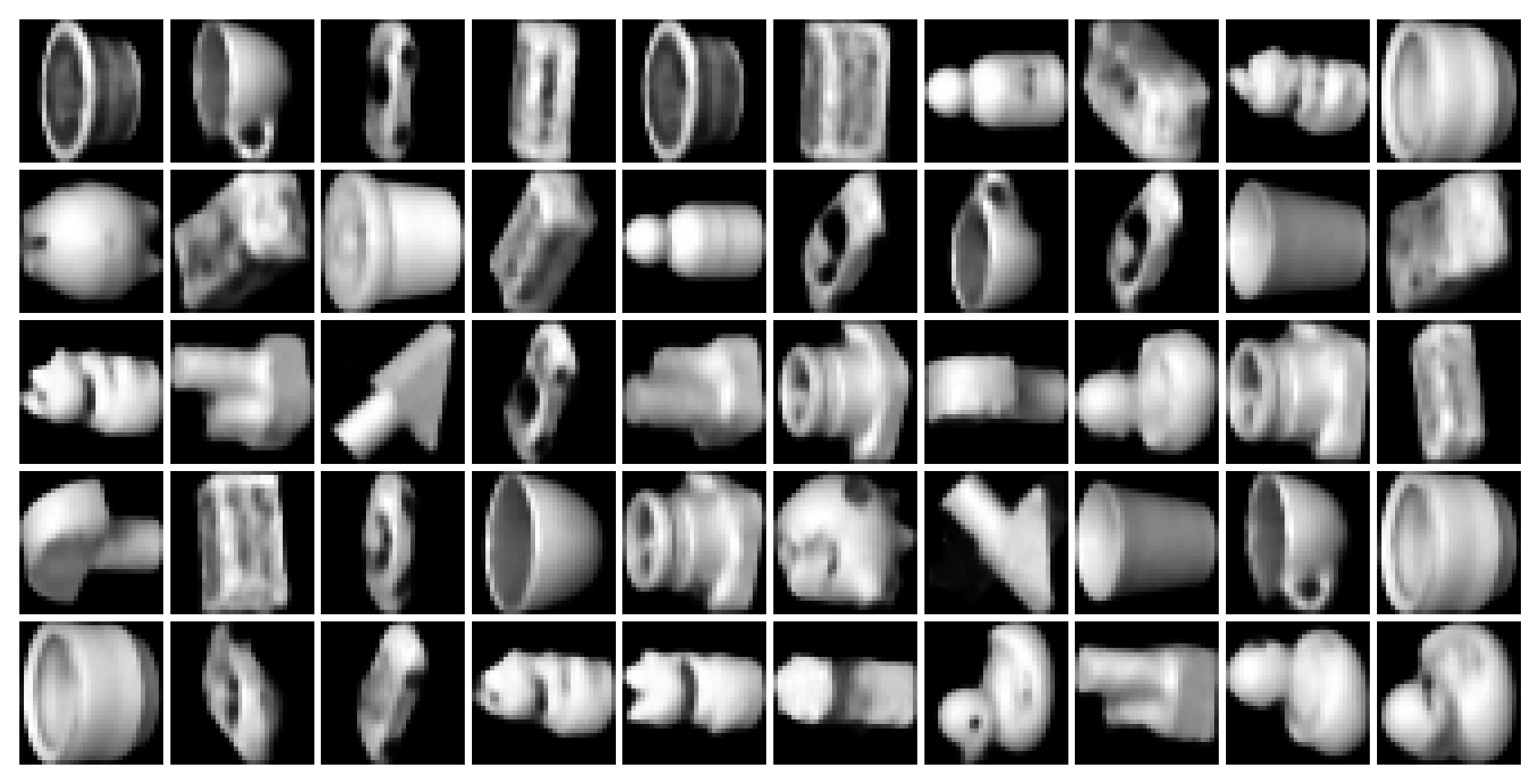}}
	\caption{Reconstructed images on Coil20. dim-$\bz = 80$ for all methods. For SWAEs, the difference of the reconstruction error for different values of $\beta$ is insignificant, and the reconstructed images look visually the same.}
\end{figure*}

\begin{figure*}[h]
	\centering
	\subfigure[Noisy real images]
	{\includegraphics[width=0.24\textwidth]{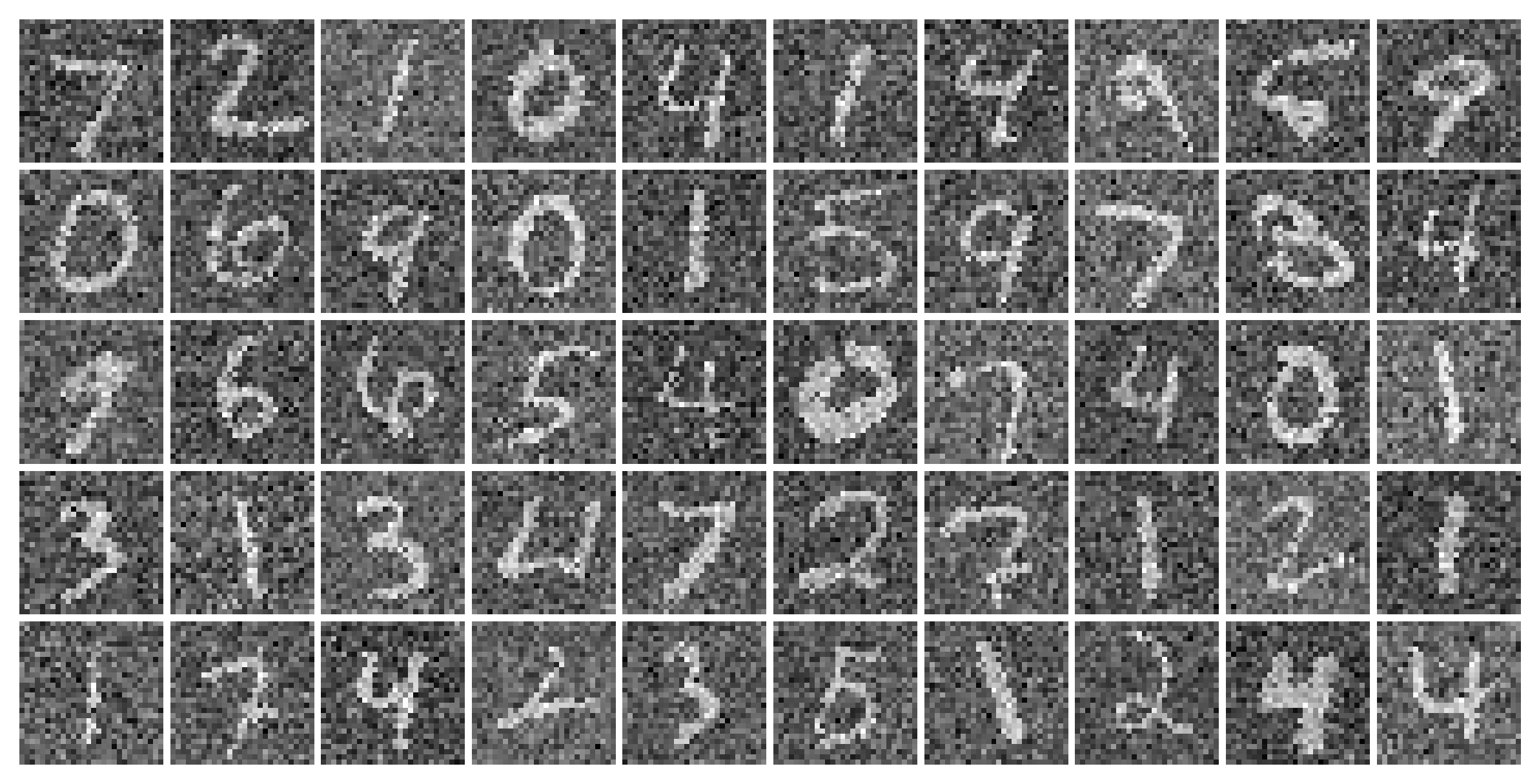}}	
	\subfigure[SWAE ($\beta = 1$)]
	{\includegraphics[width=0.24\textwidth]{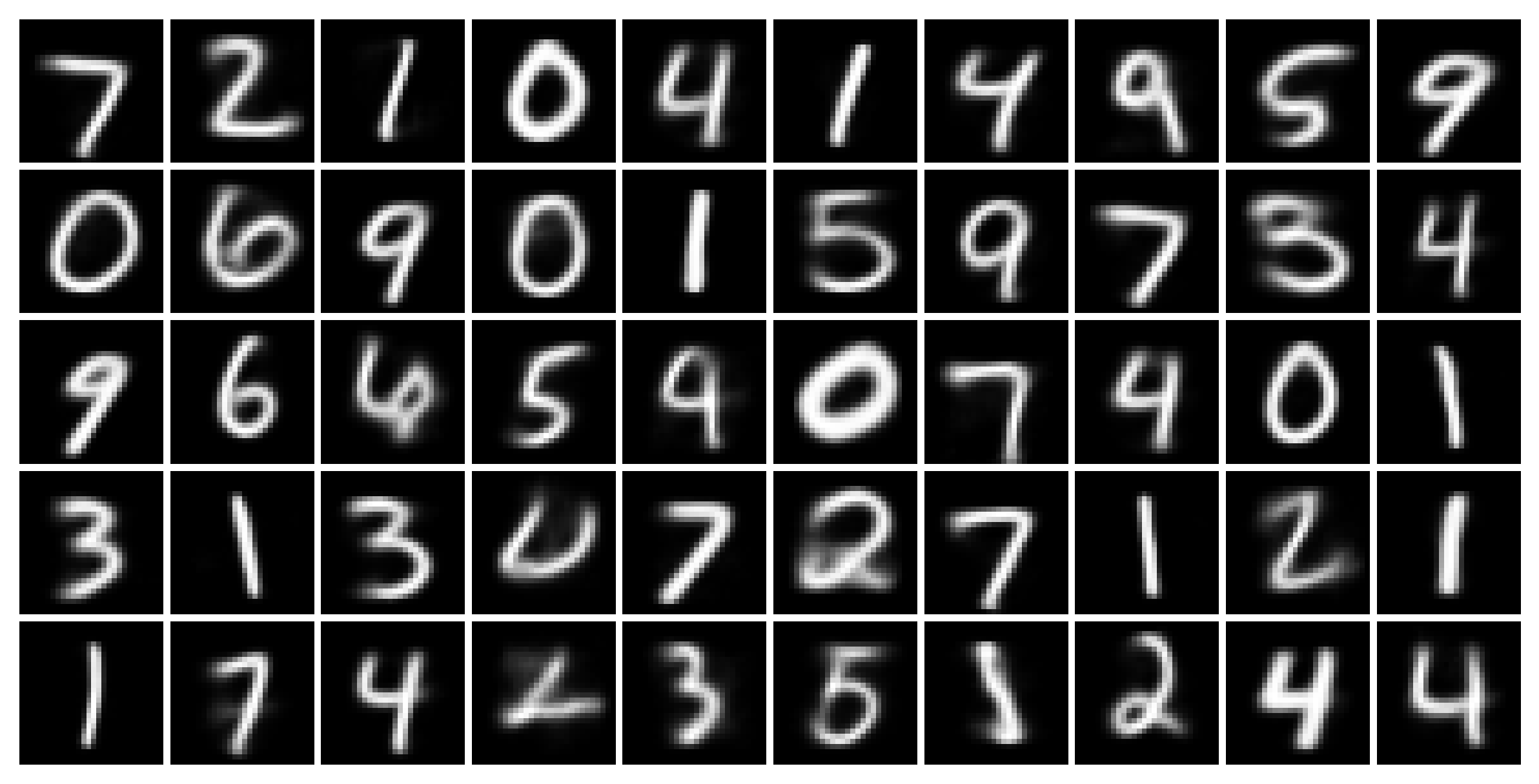}}	
	\subfigure[SWAE ($\beta = 0.5$)]
	{\includegraphics[width=0.24\textwidth]{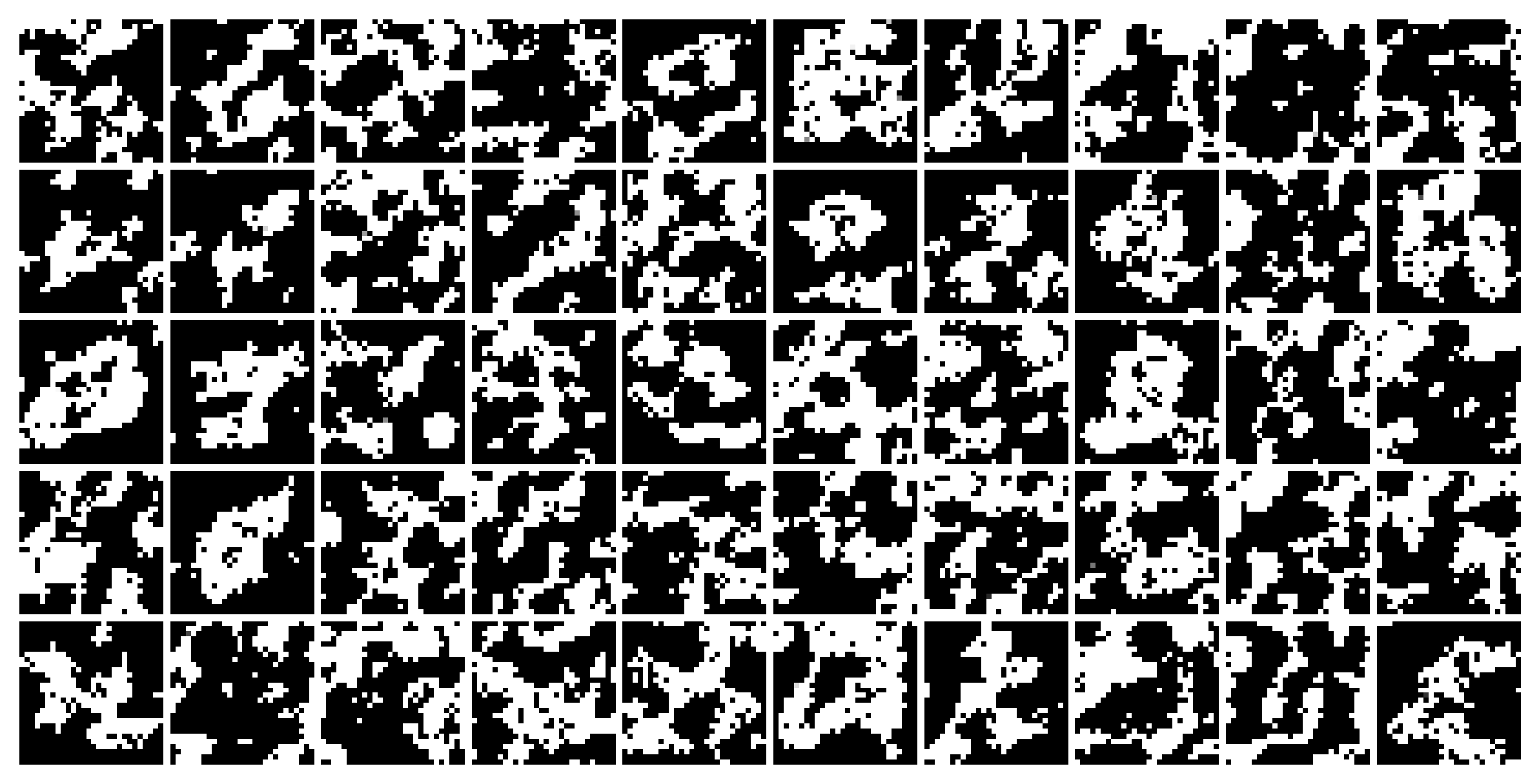}}	
	\subfigure[VAE]
	{\includegraphics[width=0.24\textwidth]{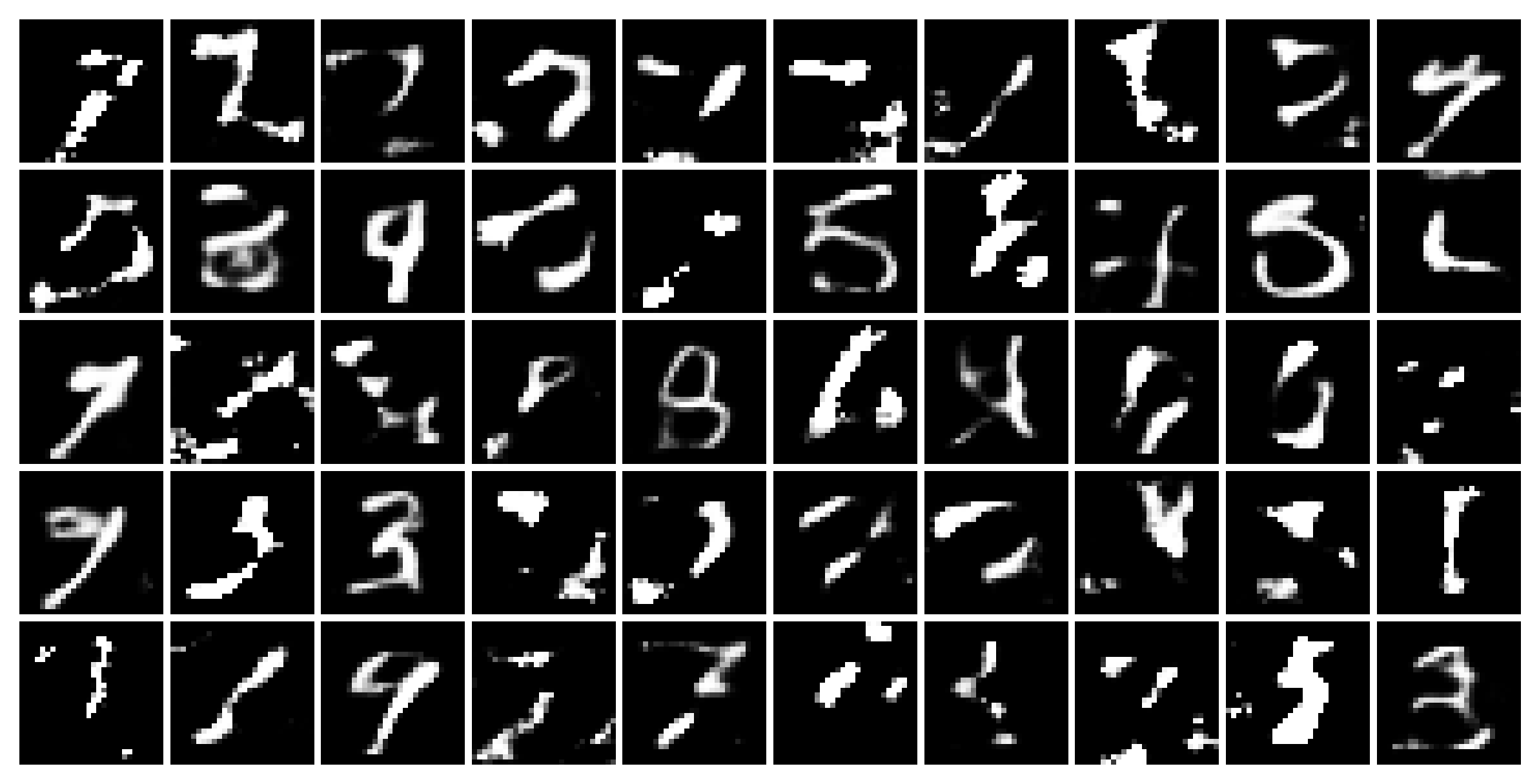}}	
	\subfigure[WAE-GAN]
	{\includegraphics[width=0.24\textwidth]{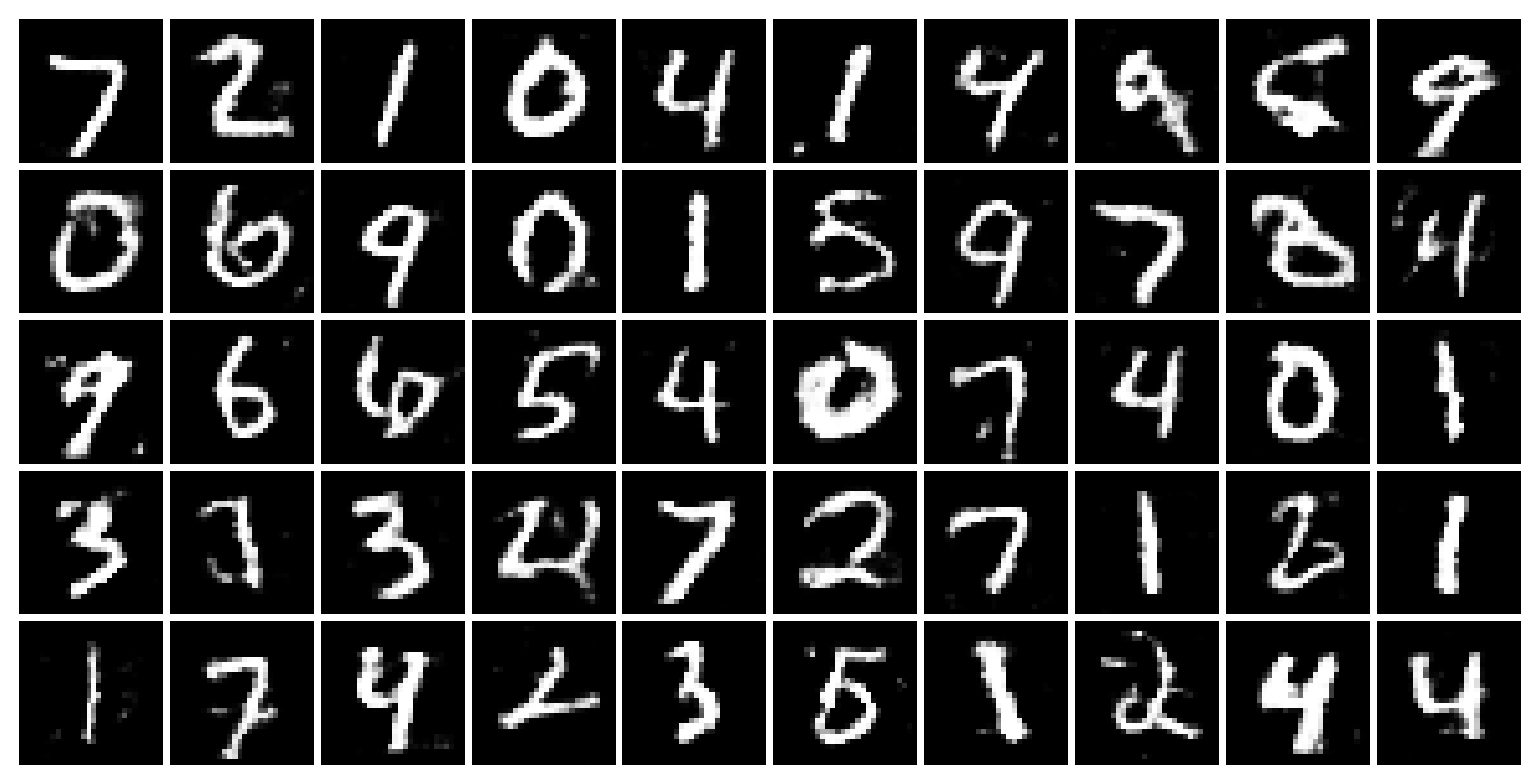}}	
	\subfigure[WAE-MMD]
	{\includegraphics[width=0.24\textwidth]{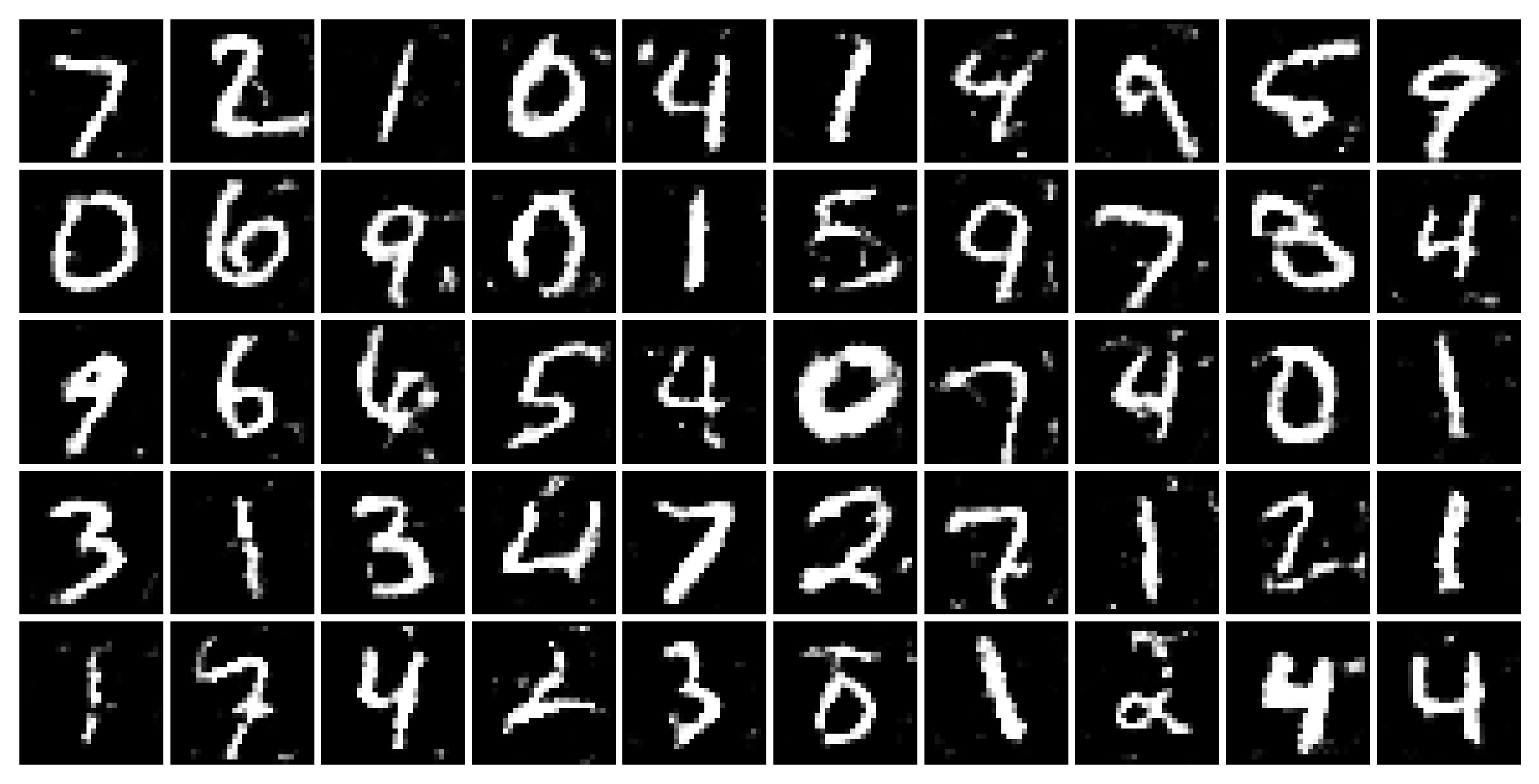}}	
	\subfigure[VampPrior]
	{\includegraphics[width=0.24\textwidth]{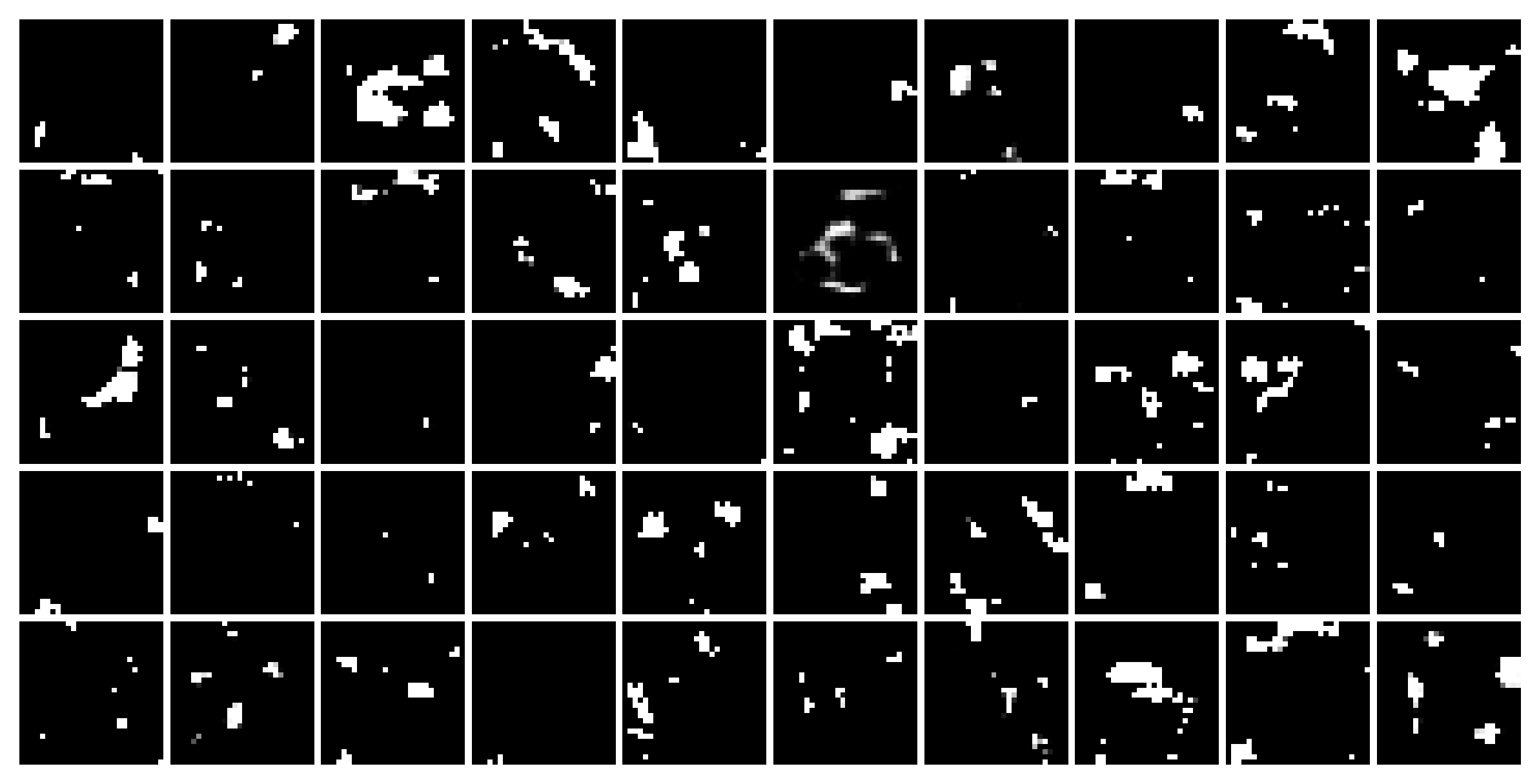}}
	\subfigure[MIM]
	{\includegraphics[width=0.24\textwidth]{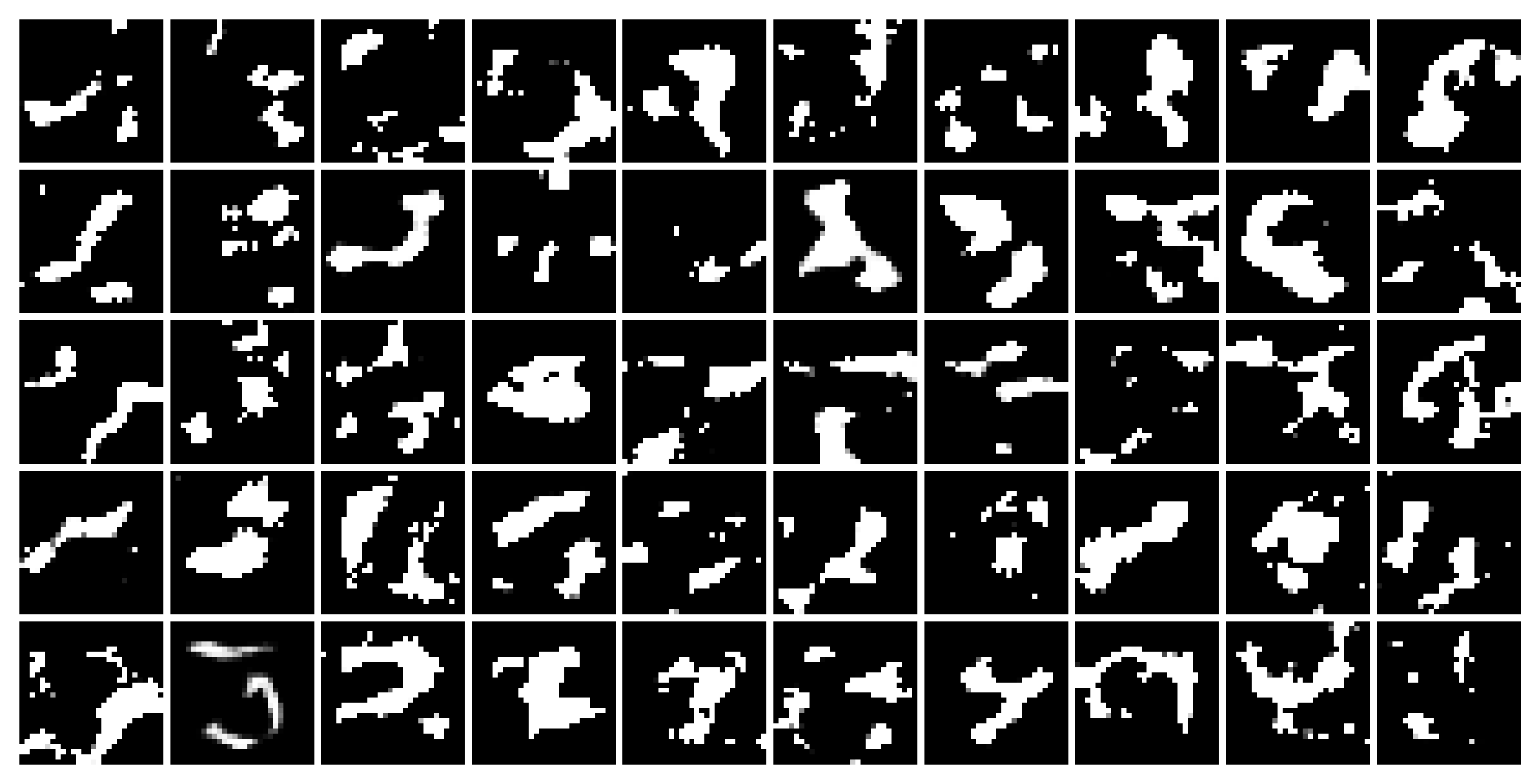}}
	\caption{Denoising effect: reconstructed images on MNIST. dim-$\bz = 80$ for all methods. SWAE ($\beta = 1$), WAE-GAN, and WAE-MMD can recover clean images. However, for WAE-GAN and WAE-MMD, we can still see some noisy dots around the digits.}
\end{figure*}

\end{document}